\pdfoutput=1
\documentclass[11pt, letterpaper]{amsart}

\usepackage[margin=1in]{geometry}
\usepackage{graphicx}

\usepackage{lipsum}
\usepackage{amsfonts}
\usepackage{epstopdf}
\ifpdf
  \DeclareGraphicsExtensions{.eps,.pdf,.png,.jpg}
\else
  \DeclareGraphicsExtensions{.eps}
\fi
\usepackage{enumitem}
\setlist[enumerate]{leftmargin=.5in}
\setlist[itemize]{leftmargin=.5in}

\usepackage{amsmath, amssymb, mathrsfs, mathtools}
\usepackage{amsthm}
\usepackage{hyperref}
\usepackage{cleveref} 
\usepackage{subcaption}
\usepackage{booktabs}
\usepackage[noend]{algpseudocode}
\usepackage{xcolor, soul}
\usepackage{verbatim}
\usepackage{framed}
\usepackage{tikz}
\usepackage[normalem]{ulem}
\usepackage{cancel}
\usepackage{algorithm}

\usepackage{aliascnt}
\theoremstyle{plain}
\newtheorem{theorem}{Theorem}[section]
\crefname{theorem}{Theorem}{Theorems}

\newcommand{\sharedthm}[3]{%
  \newaliascnt{#1}{theorem}%
  \newtheorem{#1}[#1]{#2}%
  \aliascntresetthe{#1}%
  \crefname{#1}{#2}{#3}%
  \Crefname{#1}{#2}{#3}%
}

\theoremstyle{plain}
\sharedthm{lemma}{Lemma}{Lemmas}
\sharedthm{corollary}{Corollary}{Corollaries}
\sharedthm{proposition}{Proposition}{Propositions}
\sharedthm{claim}{Claim}{Claims}
\sharedthm{fact}{Fact}{Facts}

\theoremstyle{definition}
\sharedthm{definition}{Definition}{Definitions}
\sharedthm{example}{Example}{Examples}

\theoremstyle{remark}
\sharedthm{assumption}{Assumption}{Assumptions}
\sharedthm{remark}{Remark}{Remarks}
\sharedthm{hypothesis}{Hypothesis}{Hypotheses}

\allowdisplaybreaks
\everypar{\looseness=-1}

\newcommand{\pr}[1]{\mathbb{P}\left({#1}\right)}
\newcommand{\ep}[1]{\mathbb{E}\left({#1}\right)}

\newcommand{\LPCA}{{LPCA}}
\newcommand{\argmin}{{\mathrm{argmin}}}
\newcommand{\cleandatapoints}{{Y}}
\newcommand{\datapoints}{{X}}

\newcommand{\numpoints}{n}
\newcommand{\ambientdim}{{p}}
\newcommand{\intrinsicdim}{{d}}
\newcommand{\extradim}{{k}}
\newcommand{\ambientdimlong}{{\intrinsicdim{}+\extradim{}}}
\newcommand{\intrinsicdimat}[1]{{d_{#1}}}
\newcommand{\datapoint}[1]{X_{#1}}
\newcommand{\cleandatapoint}[1]{{Y_{#1}}}

\newcommand{\cdatapoint}[1]{\overline{X}_{#1}}
\newcommand{\cdatapointpinv}[1]{{\overline{X}_{#1}^{\dagger}}}
\newcommand{\cdatapointtinv}[1]{{\overline{X}_{#1}^{+}}}
\newcommand{\ccdatapoint}[1]{\overline{Y}_{#1}}

\newcommand{\nbrhd}[1]{{\mathcal{N}_{#1}}}
\newcommand{\Laplacian}{{\mathcal{L}}}
\newcommand{\eigvecs}{{\Phi}}
\newcommand{\eigvecsOB}{{U_\Phi}}

\newcommand{\eigvec}[1]{{\phi_{#1}}}
\newcommand{\eigval}[1]{{\lambda_{#1}}}
\newcommand{\eigFn}{{\phi}}
\newcommand{\eigFnLift}{{\widehat{\phi}}}
\newcommand{\eigvecat}[2]{{\phi_{#1}(\datapoint{#2})}}
\newcommand{\ceigvecat}[2]{{\overline{\phi}_{#1}(\datapoint{#2})}}
\newcommand{\cceigvecat}[2]{{\overline{\psi}_{#1}(\cleandatapoint{#2})}}
\newcommand{\gradeigvec}[1]{{\nabla\phi_{#1}}}
\newcommand{\gradeigcoeff}[1]{C_{#1}}
\newcommand{\estgradeigcoeff}[1]{\widehat{C}_{#1}}

\newcommand{\estgradeigvec}[1]{{\widehat{\nabla}\phi_{#1}}}
\newcommand{\gradeigvecat}[2]{{\nabla\phi_{#1}(\datapoint{#2})}}

\newcommand{\estgradeigvecat}[2]{{\widehat{\nabla}\phi_{#1}(\datapoint{#2})}}
\newcommand{\tkvestgradeigvecat}[2]{{\widehat{\nabla}\phi_{#1}(\datapoint{#2})}}
\newcommand{\tkvestgradceigvecat}[2]{{\widehat{\nabla}\psi_{#1}(\datapoint{#2})}}
\newcommand{\estgradeigvecsat}[2]{{\widehat{\nabla}\phi(\datapoint{#1})^{#2}}}
\newcommand{\numeig}{{m_0}}
\newcommand{\numeigforgrad}{{m}}
\newcommand{\knn}{k_{\mathrm{nn}}}

\newcommand{\fvarexp}{\mathrm{f_{var}}}

\newcommand{\ouralgofullform}{{Laplacian Eigenvector Gradient Orthogonalization}}
\newcommand{\ouralgoacronym}{{LEGO}}
\newcommand{\svd}[1]{{\mathrm{svd}}}
\newcommand{\datamanifold}{{B}}
\newcommand{\tubeofwidth}[1]{{\mathcal{T}^{#1}}}
\newcommand{\globalreach}{{\mathrm{r}}}
\newcommand{\tubescale}{{\varepsilon}}

\newcommand{\OBofTSat}[1]{Q_{#1}}
\newcommand{\OBofNSat}[1]{Q^\perp_{#1}}
\newcommand{\metric}[2]{{g_{#1}^{#2}}}
\newcommand{\emetric}[1]{{\delta_{#1}}}
\newcommand{\NBof}[1]{{N{#1}}}
\newcommand{\TBof}[1]{{T{#1}}}
\newcommand{\TSat}[2]{{T_{#1}{#2}}}
\newcommand{\NBofwidth}[2]{{N{#1}^{#2}}}
\newcommand{\NBofwidthat}[3]{{N_{#3}{#1}^{#2}}}
\newcommand{\NBtoRSMap}[1]{{\Psi^{#1}}}
\newcommand{\NBtoRSMapLift}[1]{{\widehat{\Psi}^{#1}}}
\newcommand{\ScalingFn}[1]{{\mathcal{D}_{#1}}}
\newcommand{\PBScalingFn}[1]{{\mathcal{D}_{#1}^*}}
\newcommand{\ScalingFnLift}[2]{{\widehat{\mathcal{D}}_{#1}^{#2}}}
\newcommand{\apoint}{{x}}
\newcommand{\anormal}{{\nu}}
\newcommand{\anormalcoord}{{n}}
\newcommand{\lengthof}[2]{{\left\|{#1}\right\|_{#2}}}
\newcommand{\projection}[2]{{\pi_{#1}^{#2}}}
\newcommand{\hEnergy}[1]{{E_{#1}}}
\newcommand{\vEnergy}[1]{{E_{#1}^{\perp}}}
\newcommand{\nhEnergy}[1]{{\mathcal{E}_{#1}}}
\newcommand{\nvEnergy}[1]{{\mathcal{E}_{#1}^{\perp}}}
\newcommand{\volMeasure}[1]{{dV_{#1}}}
\newcommand{\SFF}{{\Pi}} 
\newcommand{\CofSSFF}[2]{{h_{#1 #2}}} 
\newcommand{\SSFFMat}[1]{{H_{#1}}} 
\newcommand{\SSFFMatPrime}[1]{{H'_{#1}}}
\newcommand{\StdBasisVec}[1]{{e_{#1}}}
\newcommand{\CSofNC}[2]{{\gamma_{#1}^{#2}}} 
\newcommand{\NCMat}[1]{{\Gamma_{#1}}} 
\newcommand{\identity}[1]{{I_{#1}}}
\newcommand{\RMgrad}{{\mathrm{grad}}}
\newcommand{\absPplCurv}[2]{{\kappa_{#1}^{#2}}}
\newcommand{\NCCurv}[2]{{\kappa_{#1}^{\perp #2}}}
\newcommand{\WMap}[1]{{W_{#1}}}
\newcommand{\indicator}[1]{{\delta_{#1}}}
\newcommand{\SmoothVFsOn}[1]{{\mathfrak{X}(#1)}}
\newcommand{\SmoothSectionOn}[1]{{\Gamma(#1)}}
\newcommand{\numiterations}{{T}}
\newcommand{\alength}{l}
\newcommand{\BesselConst}{C_{\extradim{}}}
\newcommand{\bigO}{\mathcal{O}}
\newcommand{\bigOmega}{\Omega}
\newcommand{\localembed}[2]{\theta_{#1,#2}}
\newcommand{\globalembed}[1]{\Theta_{#1}}
\newcommand{\meanOfNbrhd}[1]{\mu_{#1}}
\newcommand{\anOrthMat}[1]{S_{#1}}
\newcommand{\atranslateVec}[1]{t_{#1}}
\newcommand{\orthMatOf}[1]{\mathbb{O}(#1)}
\newcommand{\normalDirn}[1]{\nu_{#1}}
\newcommand{\DSKernel}[2]{W_{#1#2}}
\newcommand{\localCovMat}[1]{C_{#1}}
\newcommand{\pAngle}[1]{\theta_{#1}}
\newcommand{\TBDiscrep}[1]{\mathcal{D}_{#1}}

\newcommand{\aFn}{f}

\newcommand{\printaddendum}[1]{}%

\newcommand{\bandwidth}{\sigma}
\newcommand{\cleanAdjacency}[2]{\overline{A}_{#1#2}}
\newcommand{\adjacency}[2]{A_{#1#2}}
\newcommand{\nCleanAdjacency}[2]{\overline{\mathcal{K}}_{#1#2}}
\newcommand{\nAdjacency}[2]{\mathcal{K}_{#1#2}}
\newcommand{\GaussianKernel}[1]{k_{#1}}
\newcommand{\noiseRV}[1]{Z_{#1}}
\newcommand{\auxNoiseRV}[1]{z_{#1}}
\newcommand{\cnoiseRV}[1]{\overline{Z}_{#1}}

\newcommand{\varianceProxy}{\tubescale{}}

\newcommand{\constantA}{c}
\newcommand{\constantB}{C_1}
\newcommand{\constantC}{C_2}
\newcommand{\constantD}{C_3}
\newcommand{\constantE}{C_4}
\newcommand{\boundParam}{r}
\newcommand{\nLargeA}{N_1}

\newcommand{\cleanDegree}[1]{\overline{d}_{#1}}
\newcommand{\degree}[1]{d_{#1}}

\newcommand{\nCleanDegree}[1]{\overline{\delta}_{#1}}
\newcommand{\nDegree}[1]{\delta_{#1}}

\newcommand{\cleanDegreeMat}{\overline{D}}
\newcommand{\degreeMat}{D}

\newcommand{\nCleanDegreeMat}{\overline{\mathcal{D}}}
\newcommand{\nDegreeMat}{\mathcal{D}}

\newcommand{\cleanSymGraphLaplacian}{\overline{L}}
\newcommand{\symGraphLaplacian}{L}

\newcommand{\cleanRWGraphLaplacian}{\overline{\mathcal{L}}}
\newcommand{\RWGraphLaplacian}{\mathcal{L}}

\newcommand{\radius}{R}

\newcommand{\ceigval}[1]{\overline{\lambda}_{#1}}
\newcommand{\ceigvec}[1]{\psi_{#1}}

\newcommand{\aVector}{v}
\newcommand{\aCleanVector}{\overline{v}}

\newcommand{\knnradius}[1]{r_{\numpoints{}}^{#1}}

\newcommand{\horizproxy}{horizontal}
\newcommand{\Horizproxy}{Horizontal}

\newcommand{\noisePowerConst}{\mathrm{t}} 

\hypersetup{
  pdftitle={Robust Tangent Space Estimation via Laplacian Eigenvector Gradient Orthogonalization},
  pdfauthor={Dhruv Kohli, Sawyer J. Robertson, Gal Mishne, Alexander Cloninger},
  colorlinks=true,
  linkcolor=blue,
  citecolor=blue,
  urlcolor=blue
}

\begin{document}

\title[Robust Tangent Space Estimation via LEGO]{Robust Tangent Space Estimation via Laplacian Eigenvector Gradient Orthogonalization}


\author{Dhruv Kohli*}
\address{Program in Applied and Computational Mathematics, Princeton University, NJ}
\email{dhruv.kohli@princeton.edu}
\thanks{* Equal contribution.}

\author{Sawyer J. Robertson*}
\address{Department of Mathematics, UC San Diego, CA}
\email{s5robert@ucsd.edu}

\author{Gal Mishne}
\address{Halicio\u{g}lu Data Science Institute, UC San Diego, CA}
\email{gmishne@ucsd.edu}

\author{Alexander Cloninger}
\address{Department of Mathematics and Halicio\u{g}lu Data Science Institute, UC San Diego, CA}
\email{acloninger@ucsd.edu}

\begin{abstract}
Estimating the tangent spaces of a data manifold is a fundamental problem in geometric data analysis. The standard approach, Local Principal Component Analysis (LPCA), struggles in high-noise settings due to a critical trade-off in choosing the neighborhood size. Selecting an optimal size requires prior knowledge of the geometric and noise characteristics of the data that are often unavailable. In this paper, we propose a spectral method, Laplacian Eigenvector Gradient Orthogonalization (LEGO), that utilizes the global structure of the data to guide local tangent space estimation. Instead of relying solely on local neighborhoods, LEGO estimates the tangent space at each data point by orthogonalizing the gradients of low-frequency eigenvectors of the graph Laplacian. We provide two theoretical justifications of our method. First, a differential geometric analysis on the tubular neighborhood of a manifold shows that gradients of the low-frequency Neumann eigenfunctions of the tube align closely with the manifold's tangent bundle, while an eigenfunction with high gradient in directions orthogonal to the manifold lies deeper in the spectrum. Second, a random matrix theoretic analysis also demonstrates that low-frequency eigenvectors are robust to sub-Gaussian noise. These results allow us to derive the asymptotic scaling and stability of the estimated eigenvector gradients. Numerical experiments demonstrate that LEGO yields tangent space estimates that are significantly more robust to noise than those from LPCA, resulting in marked improvements in downstream tasks such as manifold learning, boundary detection, and local intrinsic dimension estimation.
\end{abstract}


\maketitle

\section{Introduction}
Tangent space estimation is a fundamental geometric task with broad applications across numerous domains, including manifold learning~\cite{ltsa,lle,hessianeigenmaps,ldle1v2,ratsv2,meilua2024manifold,de2025low}, data denoising~\cite{gong2010locally}, multi-manifold structure learning~\cite{trillos2023large,wang2011spectral,gong2012robust,arias2011spectral}, local intrinsic dimension estimation~\cite{levina2004maximum}, connection Laplacian approximation~\cite{singer2017spectral,singer2012vector,connectionwassersteinv2}, and regression on manifolds~\cite{cheng2013local}, among others. The standard procedure for estimating the tangent space at a given point $x$ of a data manifold comprises two steps: (i) determining the local intrinsic dimension $\intrinsicdim{}$ if not known \emph{a priori}, and (ii) identifying $\intrinsicdim{}$ orthogonal directions in the ambient space that estimate a basis for the tangent space at $x$.

The most commonly adopted approach for tangent space estimation is Local Principal Component Analysis (\LPCA{}) \cite{cleveland1988locally,ltsa,ratsv2,gong2010locally,singer2012vector,cheng2013local}, which constructs the local covariance matrix using the $\knn$-nearest neighbors of a data point and extracts the leading $\intrinsicdim{}$ eigenvectors as an estimate of the tangent basis at that point. When the local intrinsic dimension $\intrinsicdim{}$ is unknown, it is often inferred by counting the smallest number of top eigenvalues whose normalized cumulative sum (explained variance ratio) exceeds a user-defined threshold.

Due to its local formulation, \LPCA{} offers a straightforward and computationally efficient approach for tangent space estimation. However, a limitation of this local nature is that the presence of noise can significantly degrade the quality of the estimated tangent spaces as demonstrated in Figures~\ref{fig:wave_on_circle:a} and~\ref{fig:wave_on_circle:b}. Specifically, there is a well-known trade-off in the choice of neighborhood size: small neighborhoods are prone to noise corruption, while larger neighborhoods introduce bias due to the underlying curvature and reach of the manifold~\cite{aamari2019nonasymptotic,kaslovsky2014non,tyagi2013tangent}. One potential strategy to address this limitation involves selecting an adaptive neighborhood size~\cite{kaslovsky2014non} that balances these competing effects. Nonetheless, the practical implementation of such adaptive schemes is hindered by the fact that the geometric quantities---curvature, reach, and the noise level---are typically unknown. As a result, selecting an appropriate neighborhood size becomes a challenging and often ill-posed problem.

In contrast, taking cues from the global structure of the data may offer an alternative route to robust tangent space estimation, avoiding the complexities of adaptive neighborhood sizes while allowing them to remain small. This perspective is widely adopted in the literature~\cite{laplacianeigenmaps,diffusionmaps,chen2019selecting,kokotnoisy} where the eigenmodes of the graph Laplacian are frequently used to encode the global geometry of data. This naturally raises the question of whether such global eigenmodes can also be used to inform local geometric structure, and improve tangent space estimation.

\begin{figure*}[t!]
    \centering
    \refstepcounter{figure}
    \refstepcounter{subfigure}\label{fig:wave_on_circle:a}
    \refstepcounter{subfigure}\label{fig:wave_on_circle:b}
    \refstepcounter{subfigure}\label{fig:wave_on_circle:c}
    \addtocounter{figure}{-1}
    \includegraphics[width=\textwidth]{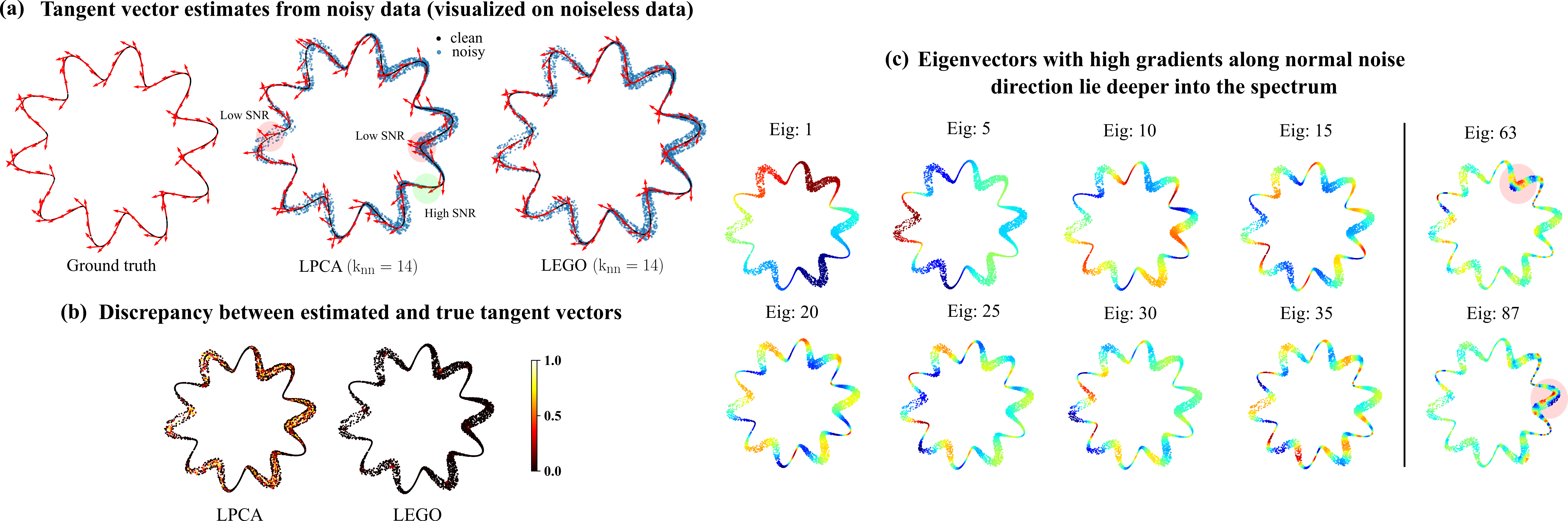}
    \caption{
    Illustration of tangent space estimation using \LPCA{} and \ouralgoacronym{} on a noisy point cloud generated by non-uniform sampling of a closed curve---wave on a circle---with heteroskedastic noise added in the normal direction.
    (a) Clean data points with ground-truth tangent vectors, along with tangent vectors estimated from the noisy data using \LPCA{} ($\knn{} = 14$ and $\intrinsicdim{} = 1$) and \ouralgoacronym{} ($\knn{} = 14$, $\numeigforgrad{} = 20$, $\numeig{} = 100$, $\beta=1/2$ and $\intrinsicdim{} = 1$).
    (b) Cosine dissimilarity between the true and the estimated tangent vectors.
    (c) Eigenvectors of the graph Laplacian constructed from the noisy data~\cite{diffusionmaps}, highlighting that those exhibiting high gradient in the noise direction lie deeper in the spectrum.}
    \label{fig:wave_on_circle}
\end{figure*}

In the continuous setting, such a connection is well established by Jones et al.\ in~\cite{jones2008manifold}, showing that under suitable regularity assumptions, for a given point $\apoint{}$ on a $\intrinsicdim{}$-dimensional Riemannian manifold~\cite{lee2018introduction}, there exist $\intrinsicdim{}$ eigenfunctions of the manifold Laplacian which yield a bilipschitz local parameterization of a sufficiently small neighborhood of $\apoint{}$ into $\mathbb{R}^{\intrinsicdim{}}$. Building on this, in~\cite{ldle1v2}, we introduced an algorithm, Low Distortion Local Eigenmaps (LDLE), which realizes their result in practice by constructing local parameterizations using global eigenvectors of the graph Laplacian. 
Unlike traditional approaches that rely on the first $\intrinsicdim{}$ non-trivial eigenvectors~\cite{laplacianeigenmaps,diffusionmaps}, LDLE selects customized subsets of $\intrinsicdim{}$ eigenvectors for each neighborhood to construct their parameterizations into $\mathbb{R}^{\intrinsicdim{}}$. These parameterizations typically have low distortion, ensuring their Jacobians have full rank and span the $\intrinsicdim{}$-dimensional tangent spaces. 

This provides empirical and theoretical support for using gradients of graph Laplacian eigenvectors to estimate local tangent spaces on data manifolds.
However, in the presence of noise, these eigenvectors may still exhibit non-zero gradients in directions orthogonal to the manifold, causing them to acquire components in the noise directions~\cite{meyer2014perturbation,ding2022impact,shuman2013emerging,hammond2011wavelets,cheng2020spectral} and consequently distorting the tangent space estimates. 

Fortunately, a principle analogous to classical Fourier analysis applies: just as the low-frequency Fourier modes capture the underlying signal while high-frequency modes tend to encode noise~\cite{stein2011fourier,katznelson2004introduction,mallat1999wavelet}, it is commonly observed that the eigenvectors corresponding to small eigenvalues of the graph Laplacian are robust to noise while the ones lying deeper in the spectrum may have nontrivial gradients in the noise directions~\cite{meyer2014perturbation,ding2022impact,shuman2013emerging,hammond2011wavelets,cheng2020spectral}, as demonstrated in Figure~\ref{fig:wave_on_circle:c}. Building upon this insight, in this work, we propose an algorithm that estimates the tangent spaces at data points using the gradients of \emph{low-frequency} global eigenvectors of the graph Laplacian. Moreover, we provide differential geometric and random matrix theoretic arguments to support our approach.

\subsection{Our contributions}
We present a spectral algorithm, \ouralgoacronym{} (\ouralgofullform{}), for estimating tangent spaces at each data point by orthogonalizing the gradients of low-frequency \emph{global} eigenvectors of the graph Laplacian derived from a noisy point cloud. Through comprehensive experiments, we show that \ouralgoacronym{} yields tangent space estimates that are significantly more robust to noise than those obtained via \LPCA{}. We also demonstrate that this increased robustness results in significant improvements across multiple downstream tasks, including manifold learning~\cite{ltsa,lle,mlle,hessianeigenmaps,ratsv2,chaudhury2015global}, boundary detection~\cite{robustboundaryv2,berry2017density,VAUGHN2024101593}, and local intrinsic dimension estimation~\cite{levina2004maximum}.
Theoretically, we justify \ouralgoacronym{} through two complementary frameworks:

$\bullet$\ Differential geometric perspective: We adopt a noise model where clean data lies on a $\intrinsicdim{}$-dimensional smooth submanifold $\datamanifold{}$ embedded in $\mathbb{R}^{\ambientdimlong{}}$, while the noisy observations lie within a tubular neighborhood $\tubeofwidth{\tubescale{}\globalreach{}}$ around $\datamanifold{}$ equipped with Euclidean metric~\cite{gray2003tubes}. Here $\globalreach{}$ is bounded by the global reach of $\datamanifold{}$~\cite{federer1959curvature,niyogi2008finding} and $\tubescale{} \in (0,1)$ controls the tube width. In practice, $\globalreach{}$ represents the maximum noise and $\tubescale{}$ is a parameter that controls the noise level. Assuming that the noise perturbs the data isotropically in directions normal to $\datamanifold{}$, the ``\horizproxy{} space'' at a noisy point---orthogonal to the noise directions---approximates the tangent space of the corresponding clean point. Thus, estimating the \horizproxy{} space provides a principled approach to robust tangent space estimation. To formalize this approach, we study the Neumann eigenfunctions of the Laplacian on the tubular neighborhood $\tubeofwidth{\tubescale{}\globalreach{}}$ of $\datamanifold{}$.
    
We establish bounds on the eigenvalue $\eigval{}$ associated with an eigenfunction $\eigFn{}$ in terms of its \horizproxy{} and vertical energies, $\nhEnergy{\datamanifold{}}(\eigFn{})$ and $\nvEnergy{\datamanifold{}}(\eigFn{})$, which quantify the net gradient of $\eigFn{}$ across the \horizproxy{} spaces and the noise directions, respectively. 
These bounds indicate that for small $\tubescale{}$, $\lambda$ approximately scales as $\bigOmega(\tubescale{}^{-2}\nvEnergy{\datamanifold{}}(\eigFn{}))$ with respect to the vertical energy, and as $\bigO(\nhEnergy{\datamanifold{}}(\eigFn{}))$ with respect to the \horizproxy{} energy. Consequently, eigenfunctions with large gradients across tube cross-sections must have large eigenvalues, while low-frequency eigenfunctions have gradients concentrated in the \horizproxy{} spaces. Motivated by these eigenvalue bounds, under a tubular noise model in the discrete setting, we show that the estimated gradients of low-frequency graph Laplacian eigenvectors have asymptotically small components along noise directions, making their orthogonalization a robust estimator of tangent spaces. These results also guide the selection of the Tikhonov regularization parameter $\beta$ used to estimate the gradients.

$\bullet$\ Random matrix perspective: We adopt a broader information plus noise-type model in which the clean data is injected with sub-Gaussian noise with variance proxy $\varianceProxy{}$. Our data model is reminiscent of that presented in~\cite{el2010information}, with a key distinction being that our data dimension is held constant while $\varianceProxy{} = o_{\numpoints{}}(1)$ in the limit as the number of data points $\numpoints{}$ tends to infinity. This model generalizes the case where the clean data lies on an embedded submanifold and the noise is confined to directions orthogonal to the tangent spaces. 
Using a Gaussian kernel with fixed bandwidth $\bandwidth{}$, we 
derive the random-walk graph Laplacians, $\cleanRWGraphLaplacian{}$ for the clean data and $\RWGraphLaplacian{}$ for the noisy data~\cite{hein2007graph,diffusionmaps}. 

By adapting techniques from the random graph literature (see, e.g., \cite{dengStrongConsistencyGraph}), assuming that the variance proxy $\varianceProxy{}$ scales as $\bigO(1/\sqrt{\numpoints{}\log{\numpoints}})$, we prove that the noisy Laplacian $\RWGraphLaplacian{}$ converges to its clean counterpart $\cleanRWGraphLaplacian{}$ in operator norm at a rate of $\numpoints{}^{-1/2}$ i.e.,
$\|\RWGraphLaplacian{}-\cleanRWGraphLaplacian{}\|_2 = \bigO(\numpoints{}^{-1/2})$
with high probability. By the Davis--Kahan theorem~\cite{yuUsefulVariantDavis2015}, it follows that the eigenvectors of $\RWGraphLaplacian{}$ remain close to those of $\cleanRWGraphLaplacian{}$, provided the eigengaps of the clean Laplacian do not decay too rapidly with $\numpoints{}$.
In the submanifold setting, Weyl's law (e.g., see \cite{chavel1984eigenvalues}) implies that eigengaps tend to shrink deeper in the spectrum. Finally, by assuming a strengthened version of our eigenvector stability result under the tubular noise model in the low-noise regime, we show that the \horizproxy{} component of the gradients estimated from the noisy data converges to that of the clean data.

\subsection{Organization}
In Section~\ref{sec:lego_algo}, we introduce our proposed algorithm, \ouralgoacronym{}, for tangent space estimation. Theoretical justifications for our method are provided in two parts: a differential geometric perspective in Section~\ref{sec:diff_geom_just}, and a random matrix theoretic analysis in Section~\ref{sec:random-laplacians}. In Section~\ref{sec:experiments}, we demonstrate the effectiveness of \ouralgoacronym{} on multiple datasets, highlighting its improved accuracy over \LPCA{} and its benefits for downstream tasks\footnote{Code repository: \url{https://github.com/chiggum/LEGO}}.

\section{Tangent space estimation via gradients of low-frequency global eigenvectors of graph Laplacian}
\label{sec:lego_algo}
Here, we introduce our algorithm, \ouralgoacronym{}, for estimating tangent spaces at noise-perturbed data points that are sampled from a tubular neighborhood of a smooth embedded submanifold. Specifically, we assume that the clean data points lie on the submanifold while the noise is constrained to the subspaces orthogonal to their tangent spaces. Our approach estimates orthonormal bases of the tangent spaces at the clean points by orthogonalizing the gradients of low-frequency global eigenvectors of the graph Laplacian constructed from the noisy data. 

Let $\cleandatapoints{} = [\cleandatapoint{1}, \ldots, \cleandatapoint{\numpoints{}}] \in \mathbb{R}^{\ambientdim{} \times \numpoints{}}$ be a point cloud sampled from a smooth compact $\intrinsicdim{}$-dimensional submanifold $\datamanifold{}$ embedded in $\mathbb{R}^{\ambientdim{}}$. Let $\datapoints{} = [\datapoint{1}, \ldots, \datapoint{\numpoints{}}] \in \mathbb{R}^{\ambientdim{} \times \numpoints{}}$ be the noisy point cloud such that $\datapoint{i}$ is obtained by adding isotropic noise to $\cleandatapoint{i}$ in the directions orthogonal to the tangent space $\TSat{\cleandatapoint{i}}{\datamanifold{}}$. Let $\nbrhd{j} = \{j_1,\ldots,j_{\knn{}}\}$ be a set containing the indices of the $\knn{}$-nearest neighbors of $\datapoint{j}$ obtained using the Euclidean metric in $\mathbb{R}^{\ambientdim{}}$. Let $\Laplacian{} \in \mathbb{R}^{\numpoints{} \times \numpoints{}}$ be the graph Laplacian constructed from $\datapoints{}$ using one of the following kernel-based methods: the random-walk kernel~\cite{diffusionmaps,hein2007graph}, the self-tuned kernel~\cite{zelnik2005self,cheng2022convergence}, or the doubly stochastic kernel~\cite{marshall2019manifold,landa2021doubly}. These construction strategies ensure that, under appropriate scaling of the kernel bandwidth and sampling density, the discrete operator $\Laplacian{}$ converges with high probability to the Laplace--Beltrami operator $\Delta_{\emetric{\ambientdim{}}}$ on a tubular neighborhood of the submanifold $\datamanifold{}$~\cite{belkin2008towards,hein2007graph,singer2006graph,garcia2020error,cheng2022convergence,cheng2024bi}. Recent results~\cite{cheng2022convergence,garcia2020error,calder2022improved,calder2022lipschitz,trillos2025minimax}, notably the approximate $\mathcal{C}^{0,1}$- and $\mathcal{H}^1$-convergence results~\cite{calder2022lipschitz,trillos2025minimax}, also establish the convergence of the spectrum of $\Laplacian{}$ to that of $\Delta_{\emetric{\ambientdim{}}}$ under technical conditions on sampling density, manifold geometry and kernel bandwidth.

We now describe our approach for estimating the gradient of an eigenvector at each data point. Let $\eigvec{i} \in \mathbb{R}^{\numpoints{}}$ denote the eigenvector of $\Laplacian{}$ associated with its $i$th smallest eigenvalue, and let $\gradeigvec{i} \in \mathbb{R}^{\ambientdim{} \times \numpoints{}}$ denote the matrix whose $j$th column, $\gradeigvecat{i}{j} \in \mathbb{R}^{\ambientdim{}}$, is the gradient of $\eigvec{i}$ at $\datapoint{j}$. We treat each of the $\ambientdim{}$ components of the gradient (i.e., each row of $\gradeigvec{i}$) as a smooth function on $\datapoints{}$ and therefore model it as a vector in the span of the eigenvectors of $\Laplacian{}$. Because eigenvectors associated with larger eigenvalues are more susceptible to noise (see, e.g., \cite{kokotnoisy,cheng2020spectral}), we make two restrictions. First, we estimate gradients only for the first $\numeigforgrad{} \ll \numpoints{}$ eigenvectors, yielding $\{\gradeigvec{1}, \ldots, \gradeigvec{\numeigforgrad{}}\}$. Second, we model the components of these gradients within the span of the first $\numeig{} \ll \numpoints{}$ eigenvectors $\{\eigvec{1}, \ldots, \eigvec{\numeig{}}\}$, where $\numeig{} \geq \numeigforgrad{}$. To ensure local fidelity, we require the estimated gradient to reproduce $\eigvec{i}$ to first order on the neighborhood $\{\datapoint{j_s} : j_s \in \nbrhd{j}\}$ of each point $\datapoint{j}$. To this end, we define the centered data points and eigenvectors as
\begin{equation}\label{eq:cdatapoint}
    \cdatapoint{j} = \begin{bmatrix}
        \datapoint{j_1}^T-\datapoint{j}^T\\
        \vdots\\
        \datapoint{j_{\knn{}}}^T-\datapoint{j}^T
    \end{bmatrix} \text{ and } \ceigvecat{i}{j} = \begin{bmatrix}
        \eigvecat{i}{j_1}-\eigvecat{i}{j}\\
        \vdots\\
        \eigvecat{i}{j_{\knn{}}}-\eigvecat{i}{j}
    \end{bmatrix},
\end{equation}
respectively, where $\{\datapoint{j_s}\}_{j_s \in \nbrhd{j}}$ are the $\knn{}$-nearest neighbors of $\datapoint{j}$. Then, for each $i \in [1,\numeigforgrad{}]$, the estimate $\estgradeigvec{i} \in \mathbb{R}^{\ambientdim{} \times \numpoints{}}$ of the gradients $\gradeigvec{i}$ is given by
\begin{equation}\label{eq:estgradeigveci}
   \estgradeigvec{i} = \estgradeigcoeff{i}\eigvecsOB{}^T
\end{equation}
where $\eigvecsOB{} \in \mathbb{R}^{\numpoints{} \times \numeig{}}$ constitutes an orthonormal basis of the range of $\eigvecs{} = \begin{bmatrix} \eigvec{1} & \eigvec{2} & \ldots & \eigvec{\numeig{}} \end{bmatrix} \in \mathbb{R}^{\numpoints{} \times \numeig{}}$ and $\estgradeigcoeff{i}$ is the solution of the following optimization problem,
\begin{equation}\label{eq:optimize}
    \estgradeigcoeff{i} = \underset{\gradeigcoeff{i}\in\mathbb{R}^{\ambientdim{} \times \numeig{}}}{\argmin}\ \frac{1}{\numpoints{}}\sum_{j=1}^{\numpoints{}}\Big(\big\|\cdatapoint{j}\gradeigvecat{i}{j} - \ceigvecat{i}{j}\big\|_2^2 + \eta_{j}\big\|\gradeigvecat{i}{j}\big\|_2^2\Big) \ \text{ s.t. }\ \gradeigvec{i} = \gradeigcoeff{i}\eigvecsOB{}^T.
\end{equation}
Although the gradients of an eigenvector across the data points are coupled via $\eigvecsOB{}$, we can still derive a closed-form optimal solution using vectorization. However, this exact solution requires inverting a matrix of size $\numeig{}\ambientdim{} \times \numeig{}\ambientdim{}$, which is often computationally expensive. Therefore, we follow a two-step procedure to approximate the solution: we first solve the unconstrained problem and then project the result onto the feasible region. This projection step uses the fact that $\eigvecsOB{}$ has orthonormal columns (i.e. $\eigvecsOB{}^T\eigvecsOB{} = \identity{\numeig{}}$). Our two-step solution, also denoted by $\estgradeigcoeff{i}$, is given by:
\begin{equation}\label{eq:estgradeigcoeff}
    \estgradeigcoeff{i} = \begin{bmatrix}
        \cdatapointtinv{1}\ceigvecat{i}{1},\ldots , \cdatapointtinv{\numpoints{}}\ceigvecat{i}{\numpoints{}}
    \end{bmatrix} \eigvecsOB{}, \text{ where } \cdatapointtinv{j} = \begin{cases}
        \cdatapointpinv{j}, & \eta_j = 0\\
        (\cdatapoint{j}^T\cdatapoint{j} + \eta_{j}I_{\ambientdim{}})^{-1}\cdatapoint{j}^T, & \eta_j > 0,
    \end{cases}
\end{equation}
and $\cdatapointpinv{j}$ is the Moore--Penrose pseudoinverse of $\cdatapoint{j}$. 
As demonstrated in~\Cref{subsec:hv_asymptotics}, relying on a naive pseudoinverse (i.e., setting $\eta_j = 0$) often causes gradient estimates to blow up in \emph{low-noise} setting where $\cdatapoint{j}$ is nearly rank-deficient. To mitigate this numerical instability, we apply Tikhonov regularization~\cite{mukherjee2010learning}, with the regularization weight given by
\begin{equation}\label{eq:eta_j}
    \eta_{j} = \sum_{s=1}^{\knn{}} \big((\cdatapoint{j}\cdatapoint{j}^T)_{ss}\big)^{1+\beta} = \sum_{s=1}^{\knn{}} \left\|\datapoint{j_s}-\datapoint{j}\right\|_2^{2(1+\beta)},
\end{equation}
where $\beta \in (0,1)$ is a hyperparameter.
This formulation assumes that the data points are scaled such that the diameter of $\datapoints{}$ is strictly less than $1$, where we define the diameter to be $\mathrm{diam}(\datapoints{}) = 2\max_{i=1}^{\numpoints{}} \|\datapoint{i} - \numpoints{}^{-1}\sum_{j=1}^{n}\datapoint{j}\|_2$.

Having obtained the gradient estimates of the eigenvectors $\{\eigvec{1}, \ldots, \eigvec{\numeigforgrad{}}\}$ at $\datapoint{j}$ given by,
\begin{equation}\label{eq:estgradeigvec}
    \estgradeigvecsat{j}{} = \begin{bmatrix}
        \estgradeigvecat{1}{j} & \ldots & \estgradeigvecat{\numeigforgrad{}}{j}
    \end{bmatrix} \in \mathbb{R}^{\ambientdim{} \times \numeigforgrad{}},
\end{equation}
we obtain an estimate of the orthonormal basis $\OBofTSat{j} \in \mathbb{R}^{\ambientdim{} \times \intrinsicdim{}}$ of the $\intrinsicdim{}$-dimensional tangent space at the $j$th point by using the top $\intrinsicdim{}$ left singular vectors of the matrix $\estgradeigvecsat{j}{}$.
If the intrinsic dimension $\intrinsicdim{}$ is not known, then it can be estimated by selecting the smallest number of top eigenvalues (squared singular values) of the matrix $\estgradeigvecsat{j}{}$ whose normalized cumulative sum exceeds a user-defined threshold $\fvarexp{}$, as described in the pseudocode below.

\begin{algorithm}[h]
\caption{Tangent space estimation via \ouralgoacronym{}.}
\label{algo:LEGO}
\begin{algorithmic}[1]
\Require $\datapoints{}$, $\Laplacian{}$, $\knn{}$, $\numeigforgrad{}$ and $\numeig{}$ where $\numeigforgrad{} \leq \numeig{}$, $\beta \in (0,1)$, either $\intrinsicdim{}$ or $\fvarexp{} \in (0,1)$.
\State $\eigvec{1},\ldots,\eigvec{\numeig{}} \leftarrow $ eigenvectors of $\Laplacian{}$ corresponding to $\numeig{}$ smallest eigenvalues \label{algo:lego:eigcomp}
\State For $i \in [1,\numeigforgrad{}]$, estimate $\estgradeigvec{i}$ using Eq.~\ref{eq:estgradeigveci}, \ref{eq:estgradeigcoeff} and~\ref{eq:eta_j} \label{algo:lego:gradest}
\For{$j \in [1,\numpoints{}]$}
\State Set $\estgradeigvecsat{j}{}$ as in Eq.~\ref{eq:estgradeigvec}.
\State $U \in \mathbb{R}^{\ambientdim{} \times \ambientdim{}}, \sigma_1 \geq \ldots \geq \sigma_{\ambientdim{}} \leftarrow$ left singular vectors and values of $\estgradeigvecsat{j}{}$
\If{$\intrinsicdim{}$ is provided}
\State $\OBofTSat{j} \leftarrow $ first $\intrinsicdim{}$ columns of $U$
\Else
\State $\intrinsicdimat{j} \leftarrow \min \{\intrinsicdim{}' \in [1,\ambientdim{}]:\sum_{i=1}^{\intrinsicdim{}'}\sigma_i^2/\sum_{i=1}^{\ambientdim{}} \sigma_i^2 \geq \fvarexp{}\}$
\State $\OBofTSat{j} \leftarrow $ first $\intrinsicdimat{j}$ columns of $U$
\EndIf
\EndFor
\State \Return $\{\OBofTSat{j}\}_1^{\numpoints{}}$
\end{algorithmic}
\end{algorithm}

\subsection{Hyperparameter selection for \ouralgoacronym{}}
\label{subsec:hyperparams}
We describe the selection strategies for four key hyperparameters: $\knn{}$, $\numeigforgrad{}$, $\numeig{}$, and $\beta$.

$\bullet$\ \textbf{$\knn{}$}: Our method requires that the graph Laplacian eigenvectors and their gradients accurately approximate the continuous eigenfunctions on the tubular neighborhood from which the noisy samples are drawn. Existing results~\cite{belkin2008towards,hein2007graph,singer2006graph,garcia2020error,cheng2022convergence,calder2022lipschitz,cheng2024bi,trillos2025minimax} on discrete-to-continuous spectrum convergence advocate for the choice $\knn{} \sim \log\numpoints{}$ or $\knn{} \sim \numpoints{}^{\alpha}$ where $\alpha \in (1/\ambientdim{},1)$. In practice, $\knn{}$ is kept small to avoid
spurious edges or ``shortcuts'' that distort the local geometry. 

$\bullet$\ \textbf{$\numeigforgrad{}$}: Under the tubular noise model considered in this work, tangent space estimation corresponds to finding the \horizproxy{} spaces within a tubular neighborhood $\tubeofwidth{\tubescale{}\globalreach{}}$ of $\datamanifold{}$. Here, $\globalreach{}$ is any number bounded by the global reach of $\datamanifold{}$, and $\tubescale{} \in (0,1)$ dictates the tube width. Accurate estimation requires $\numeigforgrad{}$ eigenfunctions whose gradients (a) exhibit minimal vertical leakage and (b) collectively span the $\intrinsicdim{}$-dimensional \horizproxy{} space everywhere in the tube.

Regarding (a), \Cref{sec:diff_geom_just} shows that the vertical energy of an eigenfunction with eigenvalue $\eigval{}$ is at most of order $\frac{1}{\BesselConst{}}(\tubescale{}\globalreach{})^{2}\eigval{}$, where $\BesselConst{}$ depends solely on the codimension $\extradim{}=\ambientdim{}-\intrinsicdim{}$ and the boundary conditions. Restricting to eigenvalues well below $\BesselConst{}(\tubescale{}\globalreach{})^{-2}$ ensures gradients remain nearly \horizproxy{}. By Weyl's law, the number of such eigenfunctions scales as $\mathrm{Vol}(\tubeofwidth{\tubescale{}\globalreach{}})(\tubescale{}\globalreach{})^{-\ambientdim{}} \sim (\tubescale{}\globalreach{})^{-\intrinsicdim{}}$, which diverges as $\tubescale{} \to 0$. This is consistent with the expectation that under low noise, many eigenfunctions have their gradients aligned with the \horizproxy{} spaces. 

Requirement (b) is more subtle. Although low eigenvalues ensure minimal vertical leakage, it is unclear whether these gradients collectively span the \horizproxy{} spaces throughout the tube. Existing literature~\cite{berard1994embedding,portegies2016embeddings}, particularly~\cite{jones2008manifold}, suggests that for a local radius $\iota(x)$ capturing the distance of $x$ from the boundary, the eigenfunctions with eigenvalues below $\iota(x)^{-2}$ (up to a constant depending on intrinsic dimension and the $\mathcal{C}^{2}$-metric on the manifold) have gradients spanning the full tangent space at $x$. However, (i) these results do not separate \horizproxy{} and vertical directions, and (ii) the radius $\iota(x)$ approaches zero as we move closer to the boundary. This motivates the following open problem:

\begin{framed}
\noindent  Fix $\tubescale{} \in (0,1)$ and for $\apoint{} \in \tubeofwidth{\tubescale{}\globalreach{}}$, let $J(\apoint{}) = [\gradeigvec{1}(\apoint{}), \dots, \gradeigvec{\numeigforgrad{}}(\apoint{})]^{T}$. For a given $\delta > 0$, does there exist a number of eigenfunctions, $\numeigforgrad{} \equiv \numeigforgrad{}(\delta, \tubescale{})$, such that at every $\apoint{} \in \tubeofwidth{\tubescale{}\globalreach{}}$, the principal angles between the span of the top-$\intrinsicdim{}$ right singular vectors of $J(\apoint{})$ and the \horizproxy{} space at $\apoint{}$ are bounded above by $\delta$? Furthermore, what is the minimum achievable $\delta^{*}(\tubescale{}) = \inf\{\delta > 0 : \numeigforgrad{}(\delta, \tubescale{}) \text{ exists}\}$ and what is the corresponding number of eigenfunctions $\numeigforgrad{}^*(\tubescale{}) = \numeigforgrad{}(\delta^{*}(\tubescale{}), \tubescale{})$?
\end{framed}

We expect that a solution to this problem would guide the choice of $\numeigforgrad{}$. In this work, we simply select a small integer value for $\numeigforgrad{}$; our ablation analysis (\Cref{fig:hyperparm_analysis}) demonstrates that tangent space estimates remain stable for a wide range of $\numeigforgrad{}$, preventing the need for excessive fine-tuning.

$\bullet$\ \textbf{$\numeig{}$}: Assuming the eigenvector gradients are smooth on the data manifold, we project them onto the subspace spanned by the first $\numeig{}$ low-frequency eigenvectors, which acts as a low-pass filter. In our experiments, we set a sufficiently large $\numeig{} \in [2\numeigforgrad{}, 5\numeigforgrad{}]$. This ensures each gradient component is well approximated, retaining the dominant signal while suppressing high-frequency noise from anomalies or non-smooth deviations.

$\bullet$\ \textbf{$\beta$}: As established in \Cref{prop:vertical_comp_reg_grad}, $\beta = 1/2$ is the optimal choice when no prior information about the noise level is available. We adopt this as our default setting across all experiments.

\subsection{Time complexity}
\label{subsec:time}
Given the local intrinsic dimension $\intrinsicdim{}$, the cost of estimating the tangent space at each point, i.e., computing the top $\intrinsicdim{}$ principal directions from the local neighborhood using \LPCA{} is $\bigO(\knn{} \ambientdim{} \intrinsicdim{})$. The time complexity of applying \LPCA{} to all $\numpoints{}$ points is therefore $\bigO(\numpoints{}\knn{} \ambientdim{} \intrinsicdim{})$.
In contrast, the time complexity of each stage of \ouralgoacronym{} is as follows:
\begin{enumerate}[leftmargin=*,label=(\roman*)]
    \item $\bigO(\numpoints{}\knn{}\numeig{} \numiterations{})$ to compute the eigenvectors $\eigvec{1}, \ldots, \eigvec{\numeig{}}$ of the graph Laplacian $\Laplacian{}$ using an iterative eigensolver where, $\numiterations{}$ is the number of iterations required for convergence~\cite{lehoucq1998arpack}.
    \item $\bigO(\numpoints{}\numeig{}^2)$ to compute an orthonormal basis $\eigvecsOB{}$ of the eigenvectors $\eigvecs{}$.
    \item $\bigO(\numpoints{}\knn{} \ambientdim{}(\min\{\knn{},\ambientdim{}\} + \numeigforgrad{})) +\bigO(\numpoints{}\numeig{}\numeigforgrad{}\ambientdim{})$ to estimate the gradients $\estgradeigvec{i}$ for all $i \in [1,\numeigforgrad{}]$, where the first term corresponds to the estimation of $\estgradeigcoeff{i}$ in Eq.~\ref{eq:estgradeigcoeff} and the second term corresponds to multiplication by $\eigvecsOB{}$ in Eq.~\ref{eq:estgradeigcoeff} and by $\eigvecsOB{}^T$ in Eq.~\ref{eq:estgradeigveci}.
    \item $\bigO(\numpoints{}\numeigforgrad{} \ambientdim{} \intrinsicdim{})$ to compute the orthonormal basis $\OBofTSat{j}$, $j \in [1, \numpoints{}]$, using the eigenvector gradients.
\end{enumerate}
Therefore, the total time complexity of \ouralgoacronym{} is 
$$\bigO(\numpoints{}(\knn{}\numeig{}\numiterations{} + \numeig{}^2 + \knn{}\ambientdim{}\min\{\knn{},\ambientdim{}\} + \knn{}\ambientdim{}\numeigforgrad{} +  \numeig{}\numeigforgrad{}\ambientdim{} + \numeigforgrad{}\ambientdim{}\intrinsicdim{})).$$
Since the number of eigenvectors $\numeig{}$ depends on the geometry of the manifold $\datamanifold{}$ and the noise characteristics rather than the number of points, we can simplify the time complexity by assuming $\numeig{} = \bigO(1)$. Since $\numeigforgrad{} \leq \numeig{}$, we also have $\numeigforgrad{} = \bigO(1)$. It is also standard practice to assume a constant number of iterations, $\numiterations{} = \bigO(1)$. Under these assumptions, the time complexity of \ouralgoacronym{} simplifies to:
$$\bigO(\numpoints{}(\knn{}\ambientdim{}\min\{\knn{},\ambientdim{}\} + \ambientdim{}\intrinsicdim{})).$$
Overall, the time complexity of \ouralgoacronym{} is linear in $\numpoints{}$, $\intrinsicdim{}$ and $\max\{\knn{},\ambientdim{}\}$, and quadratic in $\min\{\knn{},\ambientdim{}\}$. In contrast, the complexity of \LPCA{} is strictly linear across all parameters.

\section{Eigenfunctions with high gradient along the cross sections of a tube lie deeper into the spectrum}
\label{sec:diff_geom_just}
Noisy data is often modeled as a sample drawn from a tubular neighborhood surrounding an underlying smooth submanifold~\cite{gray2003tubes,niyogi2008finding,aizenbud2021non,aizenbud2025estimation,little2011estimating,little2017multiscale,genovese2012minimax}. Under this noise model, the graph Laplacian constructed from such data~\cite{diffusionmaps,cheng2022convergence,cheng2024bi} converges to the continuous Laplacian of the tubular neighborhood. Recent results~\cite{calder2022lipschitz,trillos2025minimax} have also shown $\mathcal{C}^{0,1}$ and $\mathcal{H}^1$ convergence of the spectral properties of the graph Laplacian to those of the continuous Laplacian. This motivates the study of the gradients of a Laplacian eigenfunction on the tube to better understand the behavior of the graph Laplacian eigenvector gradients derived from noisy data. Here, building on~\cite{haag2015generalised}, we show that eigenfunctions exhibiting high gradient across the cross-sections of the tubular neighborhood necessarily correspond to higher eigenvalues. Consequently, eigenfunctions associated with low eigenvalues exhibit minimal gradient in directions normal to the submanifold. 
The practical implication of our result is that the gradients of the low-frequency eigenvectors of the graph Laplacian tend to have small components in the noise directions, making them suitable for tangent space estimation.

\subsection{Preliminaries}
In the following we describe the constructs from~\cite{haag2015generalised,gray2003tubes} that are needed for our results. Let $\datamanifold{} \subset \mathbb{R}^{\ambientdimlong{}}$ be a smooth embedded compact connected $\intrinsicdim{}$-dimensional submanifold with or without boundary, equipped with the metric $\metric{\datamanifold{}}{}$ induced by the Euclidean metric $\emetric{\ambientdimlong{}}$. Let $\NBof{\datamanifold{}}$ be the normal bundle of $\datamanifold{}$ equipped with the metric $\metric{\datamanifold{}}{\perp} = \emetric{\ambientdimlong{}}|_{\NBof{\datamanifold{}}}$. Assume that there exists a tubular neighborhood $\tubeofwidth{\globalreach{}}$ of $\datamanifold{}$ such that $\datamanifold{} \subset \tubeofwidth{\globalreach{}} \subset \mathbb{R}^{\ambientdimlong{}}$, where $\globalreach{}$ is any finite number bounded by the global reach, meaning, normal segments of length less than $\globalreach{}$ issuing from distinct points of $\datamanifold{}$ do not intersect~\cite{niyogi2008finding}. Define a map 

$$\NBtoRSMap{}: \NBof{\datamanifold{}} \rightarrow \mathbb{R}^{\ambientdimlong{}}, (\apoint{},\anormal{}) \mapsto \apoint{} + \anormal{}$$
which, when restricted to 

$$\NBofwidth{\datamanifold{}}{\globalreach{}} = \{(\apoint{},\anormal) \in \NBof{\datamanifold{}}: \lengthof{\anormal}{\emetric{\ambientdimlong{}}} < \globalreach{}\} \subset \NBof{\datamanifold{}},$$
is diffeomorphic to its image $\tubeofwidth{\globalreach{}}$. Let $\projection{}{}:\NBofwidth{\datamanifold{}}{\globalreach{}} \rightarrow \datamanifold{}$ be the canonical projection $\projection{}{}(\apoint{},\anormal) = \apoint{}$ onto $\datamanifold{}$. By equipping $\NBofwidth{\datamanifold{}}{\globalreach{}}$ with the pullback metric $\metric{}{} = \NBtoRSMap{*}\emetric{\ambientdimlong{}}$, the tubular neighborhood $\tubeofwidth{\globalreach{}}$ is isometric to $\NBofwidth{\datamanifold{}}{\globalreach{}}$. This also holds for $\tubescale{}$-tubular neighborhood $\tubeofwidth{\tubescale{} \globalreach{}}$ of $\datamanifold{}$ and the normal bundle $\NBofwidth{\datamanifold{}}{\tubescale{} \globalreach{}}$ for $\tubescale{} < 1$. To keep the dependence on $\tubescale{}$ explicit, it is convenient to work with $\NBofwidth{\datamanifold{}}{\globalreach{}}$ with the pullback metric $\metric{}{\tubescale{}} = \PBScalingFn{\tubescale{}}\metric{}{}$ where the map $\ScalingFn{\tubescale{}}:\NBofwidth{\datamanifold{}}{\globalreach{}} \rightarrow \NBofwidth{\datamanifold{}}{\tubescale{} \globalreach{}}$ is given by $\ScalingFn{\tubescale{}}(\apoint{},\anormal) = (\apoint{},\tubescale{} \anormal)$. In fact, $\NBofwidth{\datamanifold{}}{\globalreach{}}$ equipped with $\metric{}{\tubescale{}}$ is isometric to $\tubeofwidth{\tubescale{} \globalreach{}}$ equipped with Euclidean metric $\emetric{\ambientdimlong{}}$. Due to this construction, the Laplacian $-\Delta_{\metric{}{\tubescale{}}}$ on  $\NBofwidth{\datamanifold{}}{\globalreach{}}$ is unitarily equivalent to the Euclidean Laplacian $-\Delta_{\emetric{\ambientdimlong{}}}$ on $\tubeofwidth{\tubescale{} \globalreach{}}$ i.e. for functions $\eigFn{} \in  C^{\infty}(\NBofwidth{\datamanifold{}}{\globalreach{}})$ it holds that

$$-\Delta_{\metric{}{\tubescale{}}}\eigFn{} = -\ScalingFnLift{\tubescale{}}{-1}\NBtoRSMapLift{}\Delta_{\emetric{\ambientdimlong{}}}\NBtoRSMapLift{-1}\ScalingFnLift{\tubescale{}}{}\eigFn{}$$
where $\ScalingFnLift{\tubescale{}}{}$ and $\NBtoRSMapLift{}$ are the unitary lifts associated with $\ScalingFn{\tubescale{}}$ and $\NBtoRSMap{}$, respectively. Specifically, $\ScalingFnLift{\tubescale{}}{}: L^2(\NBofwidth{\datamanifold{}}{\globalreach{}}, \volMeasure{\metric{}{\tubescale{}}}) \rightarrow L^2(\NBofwidth{\datamanifold{}}{\tubescale{} \globalreach{}}, \volMeasure{\metric{}{}})$ and $\NBtoRSMapLift{}: L^2(\tubeofwidth{\tubescale{} \globalreach{}}, \volMeasure{\emetric{\ambientdimlong{}}}) \rightarrow L^2(\NBofwidth{\datamanifold{}}{\tubescale{} \globalreach{}}, \volMeasure{\metric{}{}})$ are given by,

$$(\ScalingFnLift{\tubescale{}}{}\eigFn{})(\apoint{},\anormal) = \eigFn{}(\apoint{},\anormal/\tubescale{})$$ and

$$\NBtoRSMapLift{}\eigFn{} = \eigFn{} \circ \NBtoRSMap{}.$$
It follows that if $\eigFn{}$ is an eigenfunction of the Laplacian $-\Delta_{\emetric{\ambientdimlong{}}}$ on $\tubeofwidth{\tubescale{} \globalreach{}}$ with eigenvalue $\eigval{}$ then $\ScalingFnLift{\tubescale{}}{-1}\NBtoRSMapLift{}\eigFn{}$ is an eigenfunction of the Laplacian $-\Delta_{\metric{}{\tubescale{}}}$ on $\NBofwidth{\datamanifold{}}{\globalreach{}}$ with the same eigenvalue.

Using a local coordinate system, we define the \horizproxy{} energy of an eigenfunction $\eigFn{}$ on the tube $\tubeofwidth{\tubescale{} \globalreach{}}$ that captures the net gradient of $\eigFn{}$ along the submanifold $\datamanifold{}$, and the vertical energy of $\eigFn{}$ that measures its net gradient normal to $\datamanifold{}$ i.e. across the cross-sections of the tubular neighborhood $\tubeofwidth{\globalreach{}}$. To this end, let $\apoint{}^1, \ldots, \apoint{}^\intrinsicdim$ be the local coordinates on $\datamanifold{}$ and $\{\StdBasisVec{\alpha}\}_1^{\extradim{}}$ be a local orthonormal frame of $\NBofwidth{\datamanifold{}}{\globalreach{}}$ with respect to $\metric{\datamanifold{}}{\perp}$ such that every normal vector $\anormal{}(\apoint{}) \in \NBofwidthat{\datamanifold{}}{}{\apoint{}}$ can be written as $\anormal{}(\apoint{}) = \anormalcoord{}^{\alpha}\StdBasisVec{\alpha}(\apoint{})$. Consequently, $(\apoint{}^1,\ldots,\apoint{}^{\intrinsicdim{}},\anormalcoord{}^1,\ldots,\anormalcoord{}^{\extradim{}})$ form local coordinates of $\NBofwidth{\datamanifold{}}{\globalreach{}}$. These bundle coordinates induce the local coordinate vector fields,
\begin{equation}\label{eq:coord_vec_fields}
    \partial_i|_{(\apoint{},\anormalcoord{})} = \partial_{\apoint{}^i}, \ \partial_{\intrinsicdim{}+\alpha}|_{(\apoint{},\anormalcoord{})} = \partial_{\anormalcoord{}^\alpha},\ \ i \in [1,\intrinsicdim{}], \alpha \in [1,\extradim{}].
\end{equation}
For $\aFn{} \in C^{\infty}(\NBofwidth{\datamanifold{}}{\globalreach{}})$, define its canonical gradients as 
\begin{equation}
    \nabla_{\apoint{}}\aFn{} = [\partial_{\apoint{}^1}\aFn{},\ldots,\partial_{\apoint{}^{\intrinsicdim{}}}\aFn{}]^T \in \mathbb{R}^{\intrinsicdim{}},\ \nabla_{\anormalcoord{}}\aFn{} = [\partial_{\anormalcoord{}^1}\aFn{},\ldots,\partial_{\anormalcoord{}^{\extradim{}}}\aFn{}]^T \in \mathbb{R}^{\extradim{}}
\end{equation}
and $\nabla \aFn{} \in \mathbb{R}^{\ambientdimlong{}}$ is the concatenation of the two vectors. Then, for  $\eigFn{} \in C^{\infty}(\tubeofwidth{\tubescale{} \globalreach{}})$, the \horizproxy{} energy $\hEnergy{\datamanifold{}}(\eigFn{})$ and the vertical energy $\vEnergy{\datamanifold{}}(\eigFn{})$ of $\eigFn{}$ are given by (here $\eigFnLift{} = \ScalingFnLift{\tubescale{}}{-1}\NBtoRSMapLift{}\eigFn{}$ is the unitary lift of $\eigFn{}$ onto $\NBofwidth{\datamanifold{}}{\globalreach{}}$)
\begin{align}
    \hEnergy{\datamanifold{}}(\eigFn{}) &= \frac{\int_{\NBofwidth{\datamanifold{}}{\globalreach{}}} \nabla_{\apoint{}}\eigFnLift{}^T\metric{\datamanifold{}}{-1}\nabla_{\apoint{}}\eigFnLift{} \volMeasure{\metric{}{\tubescale{}}}}{\int_{\NBofwidth{\datamanifold{}}{\globalreach{}}}\eigFnLift{}^2\volMeasure{\metric{}{\tubescale{}}}}\\
    \vEnergy{\datamanifold{}}(\eigFn{}) &= \frac{\int_{\NBofwidth{\datamanifold{}}{\globalreach{}}} \nabla_{\anormalcoord{}}\eigFnLift{}^T\nabla_{\anormalcoord{}}\eigFnLift{} \volMeasure{\metric{}{\tubescale{}}}}{\int_{\NBofwidth{\datamanifold{}}{\globalreach{}}}\eigFnLift{}^2\volMeasure{\metric{}{\tubescale{}}}}.
\end{align}
\begin{remark}
\label{rmk:hv_energy}
We deliberately define the \horizproxy{} energy using the coordinate derivatives $\nabla_{\apoint{}}\eigFnLift{}$ and the metric $\metric{\datamanifold{}}{-1}$ of the base manifold, rather than the horizontal component of the Riemannian gradient of $\eigFnLift{}$ with respect to $\metric{}{\tubescale{}}$ (the quantity used to define the standard horizontal energy, i.e., the quadratic form of the horizontal Laplacian). As we show in~\Cref{lem:pullbackmetric}, the horizontal directions $\partial_i^H$ underlying this component differ from the coordinate vector fields $\partial_{\apoint{}^i}$ by a rotation of the fiber, whose size is governed by the normal connection. Consequently, the horizontal component of the gradient contains a term proportional to the vertical
gradient $\nabla_{\anormalcoord{}}\eigFnLift{}$, and a \horizproxy{} energy built from it would mix the variation of $\eigFn{}$ across the fibers with its variation along $\datamanifold{}$. In particular, an eigenfunction that varies only across the fibers could then register a non-zero \horizproxy{} energy, solely because the normal bundle twists. For tangent space estimation, we need the variation of $\eigFn{}$ along $\datamanifold{}$, separated from its variation in the normal directions along which the noisy points are displaced. Our definition of \horizproxy{} energy isolates this quantity. In contrast, no such issue arises for the vertical energy: the vertical directions
$\partial_{\anormalcoord{}^\alpha}$ are canonical and, by~\Cref{lem:pullbackmetric},
orthogonal to the horizontal ones, so $\vEnergy{\datamanifold{}}(\eigFn{})$ coincides with
the standard vertical energy, i.e., the quadratic form of the vertical Laplacian
associated with the Euclidean metric on the fibers.
\end{remark}

We further define the normalized \horizproxy{} and vertical energies of $\eigFn{} \in C^{\infty}(\tubeofwidth{\tubescale{} \globalreach{}})$ as
\begin{equation}\label{eq:normalized_energies}
    \nhEnergy{\datamanifold{}}(\eigFn{}) = \frac{1}{\eigval{\datamanifold{}_2}}\hEnergy{\datamanifold{}}(\eigFn{})\ \ \text{ and }\ \ \nvEnergy{\datamanifold{}}(\eigFn{}) = \frac{\globalreach{}^2}{\BesselConst{}}\vEnergy{\datamanifold{}}(\eigFn{}),
\end{equation}
respectively, where $\eigval{\datamanifold{}_2}$ and $\BesselConst{}/\globalreach{}^2$ are the first non-zero eigenvalues of the Laplacian $-\Delta_{\metric{\datamanifold{}}{}}$ on $\datamanifold{}$ and $-\Delta_{\emetric{\extradim{}}}$ on a ball of radius $\globalreach{}$ in $\mathbb{R}^{\extradim{}}$, respectively. Here, $\BesselConst{}$ is a constant that depends on the dimension $\extradim{}$ and the choice of the boundary conditions---either Neumann or Dirichlet.
The above normalizations enable relative comparison of the \horizproxy{} and vertical energies.

To further motivate the above definitions, consider the example where $\datamanifold{} = [0,\alength{}] \times \{0\} \subset \mathbb{R}^2$ is an interval of length $\alength{}$, and its tubular neighborhood of radius $\globalreach{}$ is given by a rectangular domain $\tubeofwidth{\globalreach{}} = \NBofwidth{\datamanifold{}}{\globalreach{}} = [0,\alength{}] \times (-\globalreach{},\globalreach{})$. The Neumann eigenfunctions of the Laplacian $-\Delta_{\emetric{2}} = -\Delta_{\metric{}{}}$ on the scaled tube $\tubeofwidth{\tubescale{} \globalreach{}} = \NBofwidth{\datamanifold{}}{\tubescale{} \globalreach{}}$ are of the form $\eigFn{}_{i,j}(\apoint{},\anormalcoord{}) = \cos(i\pi \apoint{}/\alength{})\cos(j \pi (\anormalcoord{}+\tubescale{}\globalreach{})/2\tubescale{}\globalreach{})$
with corresponding eigenvalues $\eigval{i,j} = (i\pi/\alength{})^2 + (j\pi/2\tubescale{}\globalreach{})^2$ where $i,j \in \mathbb{Z}_{\geq 0}$. Lifting $\eigFn{}_{i,j}$ back to the unscaled tube $\NBofwidth{\datamanifold{}}{\globalreach{}}$ gives $\eigFnLift{}_{i,j}(\apoint{},\anormalcoord{}) = \cos(i\pi \apoint{}/\alength{})\cos(j \pi (\anormalcoord{} + \globalreach{})/2\globalreach{})$ from which we compute the unnormalized \horizproxy{} energy as $\hEnergy{\datamanifold{}}(\eigFn{}_{i,j}) = (i\pi/\alength{})^2$ and the vertical energy as $\vEnergy{\datamanifold{}}(\eigFn{}_{i,j}) = (j\pi/2\globalreach{})^2$. Consequently, the normalized \horizproxy{} and vertical energies are given by $\nhEnergy{\datamanifold{}}(\eigFn{}_{i,j}) = i^2$ and $\nvEnergy{\datamanifold{}}(\eigFn{}_{i,j}) = j^2$, respectively.
In particular, the eigenvalue can be expressed as $\eigval{}_{i,j} = \eigval{\datamanifold{}_2}\nhEnergy{\datamanifold{}}(\eigFn{}_{i,j}) + \BesselConst{}(\tubescale{}\globalreach{})^{-2}\nvEnergy{\datamanifold{}}(\eigFn{}_{i,j})$ highlighting that, if $\globalreach{}$ is small enough such that $\BesselConst{}\globalreach{}^{-2} \geq \eigval{\datamanifold{}_2}$ then, the contribution of the vertical energy to $\eigval{i,j}$ scales as $\tubescale{}^{-2}$ relative to the \horizproxy{} energy. This means that a unit increase in the vertical energy of an eigenfunction results in a much larger increase in the eigenvalue for small $\tubescale{}$.

We end this subsection by defining a few constructs that capture the geometry of $\datamanifold{}$ and are utilized in our results.
Let $\SFF{}$ be the second fundamental form on $\datamanifold{}$ then the coefficients of the scalar second fundamental form on $\datamanifold{}$ are given by,
\begin{equation}
    \CofSSFF{\alpha i}{j} = \metric{\datamanifold{}}{\perp}(\StdBasisVec{\alpha}, \SFF{}(\partial_{\apoint{}^i},\partial_{\apoint{}^j})) = \CofSSFF{\alpha j}{i}. \label{eq:h_alpha}
\end{equation}
Let $\nabla^{\perp}$ be the normal connection with respect to $\{\StdBasisVec{\alpha}\}_1^{\extradim{}}$ then the Christoffel symbols of the normal connection are given by,
\begin{equation}\label{eq:gamma}
    \CSofNC{i\alpha}{\beta} = \metric{\datamanifold{}}{\perp}(\nabla^{\perp}_{\partial_{\apoint{}^i}}\StdBasisVec{\alpha}, \StdBasisVec{\beta}).
\end{equation}
Define a symmetric matrix $\SSFFMat{\alpha}(\apoint{}) \in \mathbb{R}^{\intrinsicdim{} \times \intrinsicdim{}}$ such that for $i,j \in [1,\intrinsicdim{}]$,
\begin{equation}\label{eq:H_alpha}
    (\SSFFMat{\alpha})_{i,j} = \CofSSFF{\alpha i}{j}.
\end{equation}
Also define a rectangular matrix $\NCMat{\beta}(\apoint{}) \in \mathbb{R}^{\intrinsicdim{}\times \extradim{}}$ such that for $j \in [1,\intrinsicdim{}]$ and $\alpha,\beta \in [1,\extradim{}]$,
\begin{equation}\label{eq:Gamma_beta}
    (\NCMat{\beta})_{j,\alpha} = \CSofNC{j \beta}{\alpha}
\end{equation}

\subsection{Our results}
Because the span of $\{\partial_{i}|_{(\apoint{},\anormalcoord{})}\}_1^\intrinsicdim{}$ may not be orthogonal to  $\{\partial_{\intrinsicdim{}+\alpha}|_{(\apoint{},\anormalcoord{})}\}_1^\extradim{}$, we apply Gram--Schmidt orthogonalization to define a new basis $\{\partial_i^H|_{(\apoint{},\anormalcoord{})}\}_1^\intrinsicdim{}$ by projecting each $\partial_i|_{(\apoint{},\anormalcoord{})}$ orthogonal to the span of $\{\partial_{\intrinsicdim{}+\alpha}|_{(\apoint{},\anormalcoord{})}\}_1^\extradim{}$. This orthogonalization yields the following representation of the pullback metric.
\begin{lemma}
\label{lem:pullbackmetric}
The pullback metric $\metric{}{\tubescale{}} = \PBScalingFn{\tubescale{}}\metric{}{} = \PBScalingFn{\tubescale{}}\NBtoRSMap{*}\emetric{\ambientdimlong{}}$ with respect to the new basis  on $\NBofwidth{\datamanifold{}}{\globalreach{}}$, $\{\partial^H_{1}|_{(\apoint{},\anormalcoord{})},\ldots,\partial^H_{\intrinsicdim{}}|_{(\apoint{},\anormalcoord{})},\partial_{\intrinsicdim{}+1}|_{(\apoint{},\anormalcoord{})},\ldots,\partial_{\intrinsicdim{}+\extradim{}}|_{(\apoint{},\anormalcoord{})}\}$, is given by 
\begin{equation}
    \metric{}{\tubescale{}}(\apoint{},\anormalcoord{}) = \begin{bmatrix}
    \metric{\datamanifold{}}{1/2}(\identity{\intrinsicdim{}} - \tubescale{} \anormalcoord{}^{\alpha}\metric{\datamanifold{}}{-1/2}\SSFFMat{\alpha}\metric{\datamanifold{}}{-1/2})^2 \metric{\datamanifold{}}{1/2}& \\
     & \tubescale{}^2 \identity{\extradim{}}
\end{bmatrix}.
\end{equation}
Consequently, the Riemannian gradient of $\eigFnLift{} \in C^{\infty}(\NBofwidth{\datamanifold{}}{\globalreach{}})$ with respect to $\metric{}{\tubescale{}}$ is given by,
\begin{equation}
    \RMgrad{}\eigFnLift{}(\apoint{},\anormalcoord{}) = \begin{bmatrix}
    \SSFFMatPrime{}(\apoint{}, \anormalcoord{})\left(\nabla_{\apoint{}} \eigFnLift{}(\apoint{},\anormalcoord{}) - \anormalcoord{}^{\beta}\NCMat{\beta}\nabla_{\anormalcoord{}} \eigFnLift{}(\apoint{},\anormalcoord{})\right)\\
     \tubescale{}^{-2}\nabla_{\anormalcoord{}} \eigFnLift{}(\apoint{},\anormalcoord{})
\end{bmatrix} \label{eq:RMgradeigFnLift}
\end{equation}
where $\SSFFMatPrime{}(\apoint{}, \anormalcoord{}) \coloneqq \metric{\datamanifold{}}{}(\apoint{})^{-1/2}(\identity{\intrinsicdim{}} - \tubescale{} \anormalcoord{}^{\alpha}\metric{\datamanifold{}}{}(\apoint{})^{-1/2}\SSFFMat{\alpha}(\apoint{}) \metric{\datamanifold{}}{}(\apoint{})^{-1/2})^{-2}\metric{\datamanifold{}}{}(\apoint{})^{-1/2}.$
\end{lemma}
Note that $\metric{}{\tubescale{}}$ is guaranteed to be positive semidefinite. However, for large $\tubescale{}$ it can become singular for certain values of $(\anormalcoord{}^1,\ldots,\anormalcoord{}^{\extradim{}})$. The following lemma provides a necessary and sufficient condition on $\tubescale{}$ that ensures the positivity of $\metric{}{\tubescale{}}$ throughout $\NBofwidth{\datamanifold{}}{\globalreach{}}$.
\begin{lemma}
\label{lem:positivity}
    Let $\absPplCurv{}{}(\apoint{}) \in \mathbb{R}_{\geq 0}$ and $\absPplCurv{}{*} \in \mathbb{R}_{\geq 0}$ be the absolute maximum principal curvature at $\apoint{} \in \datamanifold{}$ and across $\datamanifold{}$, respectively, given by $\absPplCurv{}{*} = \max_{\apoint{} \in \datamanifold{}}\absPplCurv{}{}(\apoint{})$ where
    $$\absPplCurv{}{}(\apoint{}) = \max_{\left\|v\right\|_2=1}\left(\sum_{\alpha=1}^{\extradim{}}\left(v^T\metric{\datamanifold{}}{}(\apoint{})^{-1/2}\SSFFMat{\alpha}(\apoint{})\metric{\datamanifold{}}{}(\apoint{})^{-1/2}v\right)^2\right)^{1/2}.$$
    Then, $\metric{}{\tubescale{}}$ is positive definite on $\NBofwidth{\datamanifold{}}{\globalreach{}}$ if and only if $\tubescale{}\globalreach{}\absPplCurv{}{*} < 1$. Moreover,
    \begin{equation}\label{eq:det_g_bound}
            \tubescale{}^{2\extradim{}}\det(\metric{\datamanifold{}}{})\left(1-\tubescale{} \globalreach{} \absPplCurv{}{*}\right)^{2\intrinsicdim{}} \leq \det(\metric{}{\tubescale{}}) \leq \tubescale{}^{2\extradim{}}\det(\metric{\datamanifold{}}{})\left(1+\tubescale{} \globalreach{} \absPplCurv{}{*}\right)^{2\intrinsicdim{}}.
    \end{equation}
\end{lemma}
By definition, we have $\globalreach{}\absPplCurv{}{*} < 1$ and $\tubescale{} < 1$. Consequently, the condition $\tubescale{}\globalreach{}\absPplCurv{}{*} < 1$ is satisfied in our setting, which ensures that $\metric{}{\tubescale{}}$ is positive definite on $\NBofwidth{\datamanifold{}}{\globalreach{}}$.

Now we state our main result which shows that the eigenvalue $\eigval{}$ corresponding to an eigenfunction $\eigFn{}$ scales as $\bigOmega\left(\frac{\BesselConst{}}{(\tubescale{}\globalreach{})^{2}}\right)$ with respect to a unit increase in $\nvEnergy{\datamanifold{}}(\eigFn{})$ and as $\mathcal{O}\left(\frac{\eigval{\datamanifold{}_2}}{(1-\tubescale{} \globalreach{}\absPplCurv{}{*})^{2}}\right)$ with respect to a unit increase in $\nhEnergy{\datamanifold{}}(\eigFn{})$.
\begin{theorem}
\label{thm:eval_bounds}
    If $\eigFn{}$ is a Neumann or Dirichlet eigenfunction of the Laplacian $-\Delta_{\emetric{\ambientdimlong{}}}$ on $\tubeofwidth{\tubescale{} \globalreach{}}$
    then the corresponding eigenvalue $\eigval{}$ satisfies
    \begin{equation}\label{eq:lambda_lb}
        \eigval{} \geq \frac{\BesselConst{}}{(\tubescale{}\globalreach{})^{2}}\nvEnergy{\datamanifold{}}(\eigFn{}),
    \end{equation}
    \begin{equation} 
        \eigval{} \geq \frac{\eigval{\datamanifold{}_2}\nhEnergy{\datamanifold{}}(\eigFn{})}{(1+ \tubescale{} \globalreach{}\absPplCurv{}{*})^2} + \frac{\BesselConst{}\nvEnergy{\datamanifold{}}(\eigFn{})}{(\tubescale{}\globalreach{})^{2}} -\frac{2\NCCurv{}{*}\sqrt{\eigval{\datamanifold{}_2}\BesselConst{}\nhEnergy{\datamanifold{}}(\eigFn{})\nvEnergy{\datamanifold{}}(\eigFn{})}}{(1+ \tubescale{} \globalreach{}\absPplCurv{}{*})^2}
    \end{equation}
    and
    \begin{equation}
        \eigval{} \leq \frac{\eigval{\datamanifold{}_2}\nhEnergy{\datamanifold{}}(\eigFn{})}{(1-\tubescale{} \globalreach{}\absPplCurv{}{*})^{2}} +\left(\left(\frac{\NCCurv{}{*}}{1-\tubescale{} \globalreach{}\absPplCurv{}{*}}\right)^2 + \frac{1}{(\tubescale{}\globalreach{})^{2}}\right)\BesselConst{}\nvEnergy{\datamanifold{}}(\eigFn{}) + \frac{2\NCCurv{}{*}\sqrt{\eigval{\datamanifold{}_2}\BesselConst{}\nhEnergy{\datamanifold{}}(\eigFn{})\nvEnergy{\datamanifold{}}(\eigFn{})}}{(1 - \tubescale{} \globalreach{}\absPplCurv{}{*})^2}
    \end{equation}
    where $\NCCurv{}{}(\apoint{}) \in \mathbb{R}^{\extradim{}}_{\geq 0}$ quantifies the maximum rate of twisting of $\StdBasisVec{\beta}$, $\beta \in [1,\extradim{}]$, in any direction in the tangent space $\TSat{\apoint{}}{\datamanifold{}}$, and $\NCCurv{}{*} \in \mathbb{R}_{\geq 0}$ quantifies maximum twisting among all normal directions. Specifically, $\NCCurv{\beta}{}(\apoint{}) = \left\|\metric{\datamanifold{}}{}(\apoint{})^{-1/2}\NCMat{\beta}(\apoint{})\right\|_2$, $\beta \in [1,\extradim{}]$, and $\NCCurv{}{*} = \max_{\apoint{} \in \datamanifold{}}\left\|\NCCurv{}{}(\apoint{})\right\|_2$.
\end{theorem}
The following corollary utilizes the above result to obtain the scaling of the \horizproxy{} and vertical energies in terms of the eigenvalues.
\begin{corollary}
\label{cor:eval_bounds}
If $\globalreach{} \leq \sqrt{\BesselConst{}/\eigval{\datamanifold{}_2}}$ then
$$\nvEnergy{\datamanifold{}}(\eigFn{}) \leq \tubescale{}^{2} \frac{\eigval{}}{\eigval{\datamanifold{}_2}}.$$
If $\frac{ \tubescale{} \globalreach{}\NCCurv{}{*}}{1+ \tubescale{} \globalreach{}\absPplCurv{}{*}} < 1$ then
$$\nhEnergy{\datamanifold{}}(\eigFn{}) \leq \frac{(1+ \tubescale{} \globalreach{}\absPplCurv{}{*})^2}{1-\left(\frac{ \tubescale{} \globalreach{}\NCCurv{}{*}}{1+ \tubescale{} \globalreach{}\absPplCurv{}{*}}\right)^2} \frac{\eigval{}}{\eigval{\datamanifold{}_2}}.$$
If $\frac{\BesselConst{}}{(\tubescale{}\globalreach{})^{2}}\nvEnergy{\datamanifold{}}(\eigFn{}) \leq (1-\eta) \eigval{}$ such that $\frac{( \tubescale{} \globalreach{}\NCCurv{}{*})^2}{( \tubescale{} \globalreach{}\NCCurv{}{*})^2 + (1- \tubescale{} \globalreach{}\absPplCurv{}{*})^2} < \eta  < 1$ then
\begin{align*}
\nhEnergy{\datamanifold{}}(\eigFn{}) \geq \left((1- \tubescale{} \globalreach{}\absPplCurv{}{*})\sqrt{\eta} - \tubescale{} \globalreach{}\NCCurv{}{*} \sqrt{1-\eta}\right)^2 \frac{\eigval{}}{\eigval{\datamanifold{}_2}}.
\end{align*}
\end{corollary}

An immediate consequence of the above result is that when $\tubescale{} \ll 1$, for any eigenvalue $\eigval{}$ of $-\Delta_{\emetric{\ambientdimlong{}}}$ that is not too large---say of order $\bigO(\tubescale{}^{-2(1-t)})$ for some $t \in (0,1)$---the vertical energy $\nvEnergy{\datamanifold{}}(\eigFn{})$ of the corresponding eigenfunction $\eigFn{}$ is small, of order $\bigO(\tubescale{}^{2t})$. Consequently, the gradient of such an eigenfunction has a small component in the normal directions to $\datamanifold{}$, making it a suitable candidate for tangent space estimation. Moreover, when $\tubescale{}$ is sufficiently small or alternatively under structural constraints on $\datamanifold{}$, for example $\NCCurv{}{*} = 0$, so that $\frac{ \tubescale{} \globalreach{}\NCCurv{}{*}}{1+ \tubescale{} \globalreach{}\absPplCurv{}{*}} < 1$, the \horizproxy{} energy scales as $\bigO(\eigval{})$. Additionally, if the vertical energy contribution of an eigenfunction to its total energy i.e. $\eigval{}$, is fractional, then the \horizproxy{} energy scales as $\bigOmega(\eigval{})$.

Under Dirichlet boundary conditions, the eigenvalues of $-\Delta_{\emetric{\ambientdimlong{}}}$ on $\tubeofwidth{\tubescale{} \globalreach{}}$ scale as $\bigOmega\left(\frac{1}{(\tubescale{} \globalreach{})^2}\right)$ \cite{duclos1995curvature, grieser2008thin, post2012spectral}, rendering Dirichlet eigenfunctions unsuitable for tangent space estimation. Intuitively, since a Dirichlet eigenfunction must vanish on the boundary, it must have a non-zero gradient across the cross-sections of the tube. In fact, a smaller tube width will result in high gradient across the cross-sections, making them unsuitable for estimating tangent spaces. In contrast, we demonstrate that under Neumann boundary conditions, eigenfunctions corresponding to small eigenvalues do exist, provided the Laplacian $-\Delta_{\metric{\datamanifold{}}{}}$ on $\datamanifold{}$ possesses sufficiently small Neumann eigenvalues. To establish this, we utilize a Neumann eigenfunction of $-\Delta_{\metric{\datamanifold{}}{}}$ with eigenvalue $\eigval{}_{\datamanifold{}}$ to construct a function on $\tubeofwidth{\tubescale{} \globalreach{}}$ whose Rayleigh quotient is bounded by $\frac{(1+\tubescale{} \globalreach{}\absPplCurv{}{*})^{\intrinsicdim{}}}{(1-\tubescale{} \globalreach{}\absPplCurv{}{*})^{\intrinsicdim{}+2}}\eigval{}_{\datamanifold{}}$.

\begin{theorem}
\label{thm:low_eval_exist}
    Let $\eigFn{}_{\datamanifold{}} \in C^{\infty}(\datamanifold{})$ be nonzero, and let
    \[
        \mathcal{R}_{\datamanifold{}}(\eigFn{}_{\datamanifold{}}) \coloneqq \frac{\int_{\datamanifold{}} \langle \RMgrad{}\eigFn{}_{\datamanifold{}}, \RMgrad{}\eigFn{}_{\datamanifold{}} \rangle_{\metric{\datamanifold{}}{}}\volMeasure{\metric{\datamanifold{}}{}}}{\int_{\datamanifold{}} \eigFn{}_{\datamanifold{}}^2 \volMeasure{\metric{\datamanifold{}}{}}}
    \]
    denote its Rayleigh quotient. Define the extension $\eigFnLift{}:\NBofwidth{\datamanifold{}}{\globalreach{}} \rightarrow \mathbb{R}$ of $\eigFn{}_{\datamanifold{}}$, constant on each fiber, by
    \[
        \eigFnLift{}(\apoint{},\anormalcoord{}) = \frac{\eigFn{}_{\datamanifold{}}(\apoint{})}{\left(\int_{\NBofwidth{\datamanifold{}}{\globalreach{}}} (\eigFn{}_{\datamanifold{}} \circ \projection{}{})^2\volMeasure{\metric{}{\tubescale{}}}\right)^{1/2}}.
    \]
    Then $\eigFn{} = \NBtoRSMapLift{-1}\ScalingFnLift{\tubescale{}}{}\eigFnLift{}$, defined on $\tubeofwidth{\tubescale{}\globalreach{}}$, satisfies
    \begin{equation}\label{eq:lambda_B_bound}
        \frac{(1-\tubescale{}\globalreach{}\absPplCurv{}{*})^{\intrinsicdim{}}}{(1+\tubescale{}\globalreach{}\absPplCurv{}{*})^{\intrinsicdim{}+2}}\,\mathcal{R}_{\datamanifold{}}(\eigFn{}_{\datamanifold{}}) \leq \frac{\int_{\tubeofwidth{\tubescale{}\globalreach{}}}|\nabla\eigFn{}|^2\volMeasure{\emetric{\ambientdimlong{}}}}{\int_{\tubeofwidth{\tubescale{}\globalreach{}}}\eigFn{}^2\volMeasure{\emetric{\ambientdimlong{}}}} \leq \frac{(1+\tubescale{}\globalreach{}\absPplCurv{}{*})^{\intrinsicdim{}}}{(1-\tubescale{}\globalreach{}\absPplCurv{}{*})^{\intrinsicdim{}+2}}\,\mathcal{R}_{\datamanifold{}}(\eigFn{}_{\datamanifold{}}).
    \end{equation}
    If $\eigFn{}_{\datamanifold{}}$ is a Neumann eigenfunction of $-\Delta_{\metric{\datamanifold{}}{}}$ with eigenvalue $\eigval{}_{\datamanifold{}}$, then $\mathcal{R}_{\datamanifold{}}(\eigFn{}_{\datamanifold{}}) = \eigval{}_{\datamanifold{}}$.
\end{theorem}

The above result, \Cref{thm:eval_bounds} and the min-max principle yield the following corollary: if $\eigval{\datamanifold{}_\ell} = \bigO\big((\tubescale{}^{1-t}\globalreach{})^{-2}\big)$ for some $t \in (0,1)$, then the first $\ell$ Neumann eigenfunctions of the tube all have vertical energy of order $\bigO(\tubescale{}^{2t})$.
\begin{corollary}
\label{cor:lowfreqegnfn}
Let $0 = \eigval{\datamanifold{}_1} < \eigval{\datamanifold{}_2} \leq \cdots$ be the Neumann eigenvalues of $-\Delta_{\metric{\datamanifold{}}{}}$ on $\datamanifold{}$. Let $0 = \eigval{1} < \eigval{2} \leq \cdots$ be the Neumann eigenvalues of $-\Delta_{\emetric{\ambientdimlong{}}}$ on $\tubeofwidth{\tubescale{}\globalreach{}}$, with corresponding orthonormal eigenfunctions $\eigFn{}_1, \eigFn{}_2, \ldots$. Both sequences are counted with multiplicity. Then, for every $\ell \geq 1$ and $j \leq \ell$,
\begin{equation}\label{eq:vert_energy_bound}
    \nvEnergy{\datamanifold{}}(\eigFn{}_j) \leq \tubescale{}^{2}\,\frac{\globalreach{}^{2}}{\BesselConst{}}\,\frac{(1+\tubescale{}\globalreach{}\absPplCurv{}{*})^{\intrinsicdim{}}}{(1-\tubescale{}\globalreach{}\absPplCurv{}{*})^{\intrinsicdim{}+2}}\,\eigval{\datamanifold{}_\ell}.
\end{equation}
In particular, for $\ell \geq 2$, the $\ell-1$ mutually orthogonal, non-constant eigenfunctions $\eigFn{}_2, \ldots, \eigFn{}_\ell$ all satisfy the above equation.
\end{corollary}

\subsection{\Horizproxy{} and vertical components of the estimated eigenvector gradient}
\label{subsec:hv_asymptotics}
We now analyze the asymptotic scaling of the estimated eigenvector gradients in \ouralgoacronym{}. We begin by introducing the tubular noise model, which we also utilize in~\Cref{subsec:eig_grad_convergence} to prove eigenvector gradient stability under small data perturbations. Suppose the clean data $\cleandatapoints{}$ is sampled uniformly from $\datamanifold{}$, and the noisy observations $\datapoints{}$ are drawn from the tubular neighborhood $\tubeofwidth{\tubescale{}\globalreach{}}$. We model the noisy points as $\datapoint{j} = \cleandatapoint{j} + \noiseRV{j}$ where $\noiseRV{j} = \OBofNSat{j}\auxNoiseRV{j}$, the columns of $\OBofNSat{j} \in \mathbb{R}^{\ambientdim{} \times \extradim{}}$ form an orthonormal basis for the normal space at $\cleandatapoint{j}$, and $\auxNoiseRV{j} \in \mathbb{R}^{\extradim{}}$ is drawn uniformly from a zero-centered ball of radius $\tubescale{}\globalreach{}$  where $\tubescale{} \in (0,1)$. As defined in~\Cref{sec:lego_algo}, let $\OBofTSat{j} \in \mathbb{R}^{\ambientdim{} \times \intrinsicdim{}}$ denote the orthonormal basis for the tangent space at $\cleandatapoint{j}$.

We now derive the limiting behavior of the mean squared \horizproxy{} and vertical components of the estimated gradients of an eigenvector $\eigFn{}_{i}$ in \ouralgoacronym{}. This asymptotic analysis is motivated by~\Cref{cor:eval_bounds}, and is governed by three key parameters: the relative tube width $\tubescale{}$ (controlling the noise level), the neighborhood radius $\knnradius{}$ (where local distances satisfy $\left\|\datapoint{j_s}-\datapoint{j}\right\|_2 = \Theta(\knnradius{})$, $s \in [1,\knn{}]$ and $j \in [1,\numpoints{}]$), and the Tikhonov regularization parameter $\beta$. Throughout this subsection, the eigenvector index $i \in [1,\numeigforgrad{}]$ is fixed.

\begin{assumption}
\label{assump:eval_bounds}
Suppose that for sufficiently large $\numpoints{}$, there exist vectors $\mathfrak{g}_{ij} \in \mathbb{R}^{\intrinsicdim{}}$ and $\mathfrak{g}_{ij}^{\perp} \in \mathbb{R}^{\extradim{}}$ for each $j \in [1,\numpoints{}]$ such that
\begin{equation*}
    \ceigvecat{i}{j} =  \cdatapoint{j}(\OBofTSat{j}\mathfrak{g}_{ij} + \OBofNSat{j}\mathfrak{g}_{ij}^{\perp}) + \xi_j^{\circ 2},
\end{equation*}
where $\cdatapoint{j}$ and $\ceigvecat{i}{j}$ are defined in Eq.~\ref{eq:cdatapoint}, and $\xi_j^{\circ 2}$ is the Hadamard (elementwise) square of the vector $\xi_j \in \mathbb{R}^{\knn{}}$ with entries $\xi_{j_s} = \bigO(\left\|\datapoint{j_s}-\datapoint{j}\right\|_2)$. 
Furthermore, we assume that the eigenvector $\eigFn{}_{i}$ is such that the mean squared norms of these discrete components scale as: 
\begin{align*}
    \frac{1}{\numpoints{}}\sum_{j=1}^{\numpoints{}}\left\|\mathfrak{g}_{ij}\right\|_2^2 &= \Theta(\eigval{i}),\\
    \frac{1}{\numpoints{}} \sum_{j=1}^{\numpoints{}}\left\|\mathfrak{g}_{ij}^{\perp}\right\|_2^2 &= \bigO(\eigval{i} \tubescale{}^{2\noisePowerConst{}}) \ \text{for some $\noisePowerConst{} \in (0,1]$}.
\end{align*}
\end{assumption}

The above assumption posits that the mean squared vertical component of the discrete gradients scales as $\tubescale{}^{2\noisePowerConst{}}$ relative to the horizontal component. Although stated here as an assumption, this relationship can possibly be established by combining~\Cref{cor:eval_bounds}, discrete-to-continuous spectrum convergence results~\cite{garcia2020error,calder2022improved,calder2022lipschitz,trillos2025minimax} and the convergence guarantees for local linear regression estimators~\cite{ruppert1994multivariate,cheng2013local}. We defer this analysis to future work.

With the above assumption and for general configuration of the noisy points within each neighborhood, we derive the asymptotic scaling of the \horizproxy{} and vertical gradient components of $\eigFn{}_i$ at $\datapoint{j}$, estimated \textit{without} regularization.

\begin{proposition}
\label{prop:vertical_comp_estim_grad}
Suppose Assumption~\ref{assump:eval_bounds} holds. In addition, assume that for $\numpoints{}$ sufficiently large, $\mathrm{range}(\OBofTSat{j})$ is contained in $\mathrm{range}(\cdatapoint{j}^T)$ and $\left\|\cdatapoint{j}^{\dagger}\right\|_2 = \bigO\left((\sqrt{\knn{}}\min\{\tubescale{}, \knnradius{}\})^{-1}\right)$ for each $j \in [1,\numpoints{}]$. 
Then the \horizproxy{} and vertical components of $\estgradeigvecat{i}{j} = \cdatapoint{j}^{\dagger}\ceigvecat{i}{j}$ satisfy,
\begin{align*}
    \frac{1}{\numpoints{}} \sum_{j=1}^{\numpoints{}}\left\|\OBofTSat{j}^T\estgradeigvecat{i}{j}\right\|_2^2 &= \Theta(\eigval{i}) + \bigO\left(\frac{\knnradius{2}}{\min\{\tubescale{},\knnradius{}\}}\left(\sqrt{\eigval{i}} + \frac{\knnradius{2}}{\min\{\tubescale{},\knnradius{}\}}\right)\right),\\
    \frac{1}{\numpoints{}} \sum_{j=1}^{\numpoints{}}\left\|(\OBofNSat{j})^T\estgradeigvecat{i}{j}\right\|_2^2 &= \bigO\left(\left(\sqrt{\eigval{i}}\tubescale{}^{\noisePowerConst{}} + \frac{\knnradius{2}}{\min\{\tubescale{},\knnradius{}\}}\right)^2\right).
\end{align*}
\end{proposition}

The second assumption naturally holds for sufficiently large neighborhoods ($\knn{} \gg \ambientdim{}+1$) of affinely independent points. The third is satisfied by generic neighborhoods where $\eigval{\min}(\cdatapoint{j}^T\cdatapoint{j})$ scales as $\bigOmega(\knn\tubescale{}^2)$ under low noise, and as $\bigOmega(\knn{} \knnradius{2})$ under high noise. Under these conditions, the mean squared \horizproxy{} component of the estimated gradients scales as $\Theta(\eigval{i})$, while the vertical component scales as $\bigO(\eigval{i}\tubescale{}^{2\noisePowerConst{}})$, both up to a residual term. Notably, in the high-noise setting, this residual scales as $\bigO(\knnradius{})$, which vanishes as $\numpoints{} \rightarrow \infty$. However, under low noise ($\tubescale{} \ll \knnradius{2}$), the residual scales as $(\tubescale{}^{-1}\knnradius{2})^2$, leading to a numerical blow-up.

To overcome this instability, \ouralgoacronym{} employs Tikhonov regularization for gradient estimation (see Eq.~\ref{eq:estgradeigcoeff} and Eq.~\ref{eq:eta_j}). Retaining Assumption~\ref{assump:eval_bounds} and assuming vanishing correlation between the \horizproxy{} and vertical components of the centered noisy neighborhoods, the following result derives the asymptotic behavior of these regularized gradient estimates.

\begin{proposition}
\label{prop:vertical_comp_reg_grad}
Suppose Assumption~\ref{assump:eval_bounds} holds.
In addition, assume that for $\numpoints{}$ sufficiently large, $\left\|(\cdatapoint{j}\OBofTSat{j})^T(\cdatapoint{j}\OBofNSat{j})\right\|_2 = o_{\numpoints{}}(\knn{} \knnradius{} \max\{\knnradius{2}, \min\{\tubescale{}, \knnradius{}\}\})$ for each $j \in [1,\numpoints{}]$. Then the \horizproxy{} and vertical components of $\tkvestgradeigvecat{i}{j} = \cdatapoint{j}^{+}\ceigvecat{i}{j}$ satisfy,
\begin{align*}
    \frac{1}{\numpoints{}} \sum_{j=1}^{\numpoints{}}\left\|\OBofTSat{j}^T\tkvestgradeigvecat{i}{j}\right\|_2^2 &= \Theta(\eigval{i}) + \bigO(\eigval{i}(\knnradius{2\beta} + \tubescale{}^{\noisePowerConst{}}\widetilde{r}_{\numpoints{}}) + \sqrt{\eigval{i}}\knnradius{1-\beta}),\\
    \frac{1}{\numpoints{}} \sum_{j=1}^{\numpoints{}}\left\|(\OBofNSat{j})^T\tkvestgradeigvecat{i}{j}\right\|_2^2 &= \bigO((\sqrt{\eigval{i}} (\tubescale{}^{\noisePowerConst{}} + \widetilde{r}_{\numpoints{}}) + \knnradius{1-\beta})^2),
\end{align*}
where $\widetilde{r}_{\numpoints{}}$ depends on the regime of the noise level $\tubescale{}$ as follows:
\begin{align*}
    \widetilde{r}_{\numpoints{}} = \begin{cases}
        \bigO(\knnradius{2\beta}), & \knnradius{} < \tubescale{} < 1\\
        \bigO( \tubescale{}^{-1} \knnradius{1+2\beta}), & \knnradius{1+\beta} < \tubescale{} \leq \knnradius{}\\
        \bigO(\tubescale{}\knnradius{-1}), & \knnradius{2} <  \tubescale{} \leq \knnradius{1+\beta} \\
        \bigO( \knnradius{}), & 0 < \tubescale{} \leq \knnradius{2}.
    \end{cases}
\end{align*}
\end{proposition}

The additional assumption ensures that for sufficiently large $\numpoints{}$, the \horizproxy{} and vertical components of the centered noisy neighborhoods are asymptotically uncorrelated. This condition naturally holds under isotropic noise (including the uniform noise assumed here) and prevents pathological cases where the noise heavily correlates with the clean data points. Consequently, the mean squared \horizproxy{} component of the regularized gradient estimates scales as $\Theta(\eigval{i})$, while the vertical component scales as $\bigO(\eigval{i}\tubescale{}^{2\noisePowerConst{}})$. Crucially, the residual terms now converge to zero as $\numpoints{} \rightarrow \infty$ across all noise levels, ensuring stable estimates.

This result also guides the optimal choice of the Tikhonov regularization hyperparameter $\beta$ defined in Eq.~\ref{eq:eta_j}. If the noise scale $\tubescale{}$ is unknown, balancing the worst-case rate of $\widetilde{r}_{\numpoints{}}$, namely $\bigO(\knnradius{\beta})$, with $\bigO(\knnradius{1-\beta})$ yields an optimal $\beta = 1/2$. Because noise characteristics are typically unknown in practice, we default to $\beta = 1/2$ across all our experiments.

\section{Robustness of Laplacian eigenvectors under noise}
\label{sec:random-laplacians}
Low-frequency eigenvectors of the Laplacian often exhibit high stability under noise perturbations of the data, as illustrated in Figure~\ref{fig:noise_stability_eig_1}. In this section, we analyze the stability of the Laplacian eigenvectors from the angle of robustness of random kernel Laplacians to sub-Gaussian noise. In turn, by the Davis--Kahan theorem~\cite{yuUsefulVariantDavis2015}, if the low-frequency eigengaps of the random kernel Laplacians do not vanish too quickly, the corresponding Laplacian eigenvectors will be robust to noise perturbations as well.

\begin{figure}[h!]
    \centering
    \includegraphics[width=0.65\textwidth]{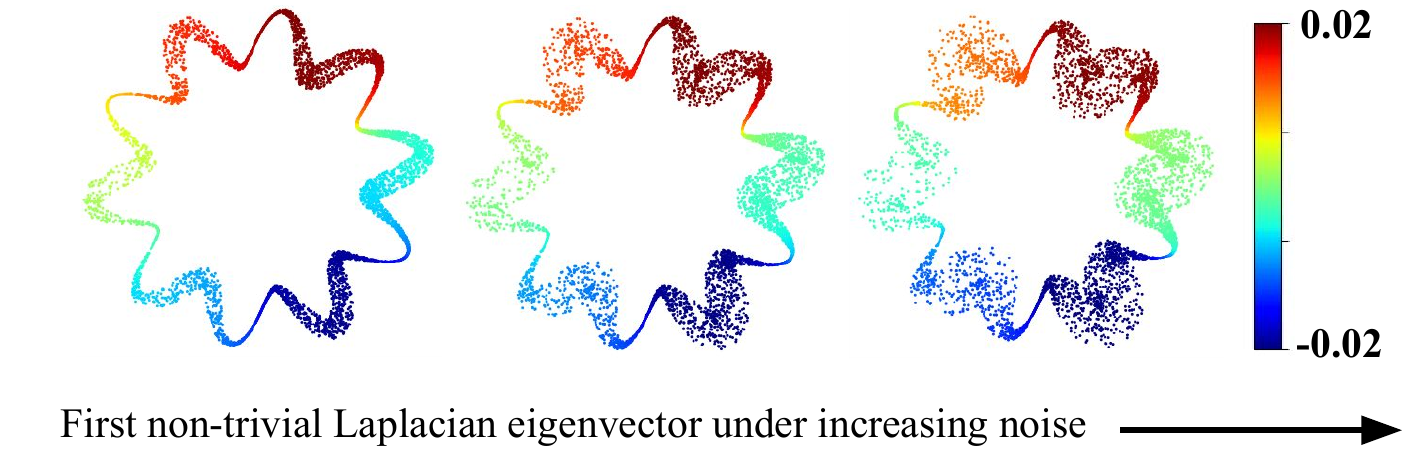}
    \caption{The first non-trivial eigenvector of the Laplacian $\RWGraphLaplacian{}$ is plotted against increasing noise level $\varianceProxy{}$. Here, the noise is independent but non-identically distributed as in Figure~\ref{fig:wave_on_circle:a} (also see Remark~\ref{rmk:noise}). The common colorbar represents the range of values, from minimum to maximum, across the eigenvectors.}
    \label{fig:noise_stability_eig_1}
\end{figure}

As before, let  $\cleandatapoints{} = \{\cleandatapoint{1}, \ldots, \cleandatapoint{\numpoints{}}\}$ be the clean data points sampled from a bounded region in $\mathbb{R}^\ambientdim{}$. Fixing a bandwidth $\bandwidth{}>0$, we define the clean or ground-truth kernel adjacency matrix $\cleanAdjacency{}{} \in\mathbb{R}^{\numpoints{}\times \numpoints{}}$ entrywise by the formula
\begin{equation}\label{eq:kernel-adj-mean-new}
    \cleanAdjacency{i}{j} = \GaussianKernel{\bandwidth{}}(\cleandatapoint{i} - \cleandatapoint{j}),
\end{equation}
where $\GaussianKernel{\bandwidth{}}:\mathbb{R}^{\ambientdim{}}\rightarrow(0, 1]$ is the Gaussian kernel defined as,
\begin{equation}\label{def:sigma}
    \GaussianKernel{\bandwidth{}}(z) = e^{-\|z\|_2^2/\bandwidth{}^2}.
\end{equation}
Later, we will make use of the following estimate on the Lipschitz constant of $\GaussianKernel{\bandwidth{}}$.

\begin{lemma}\label{lem:lipschitz-gaussian}
    For any $\bandwidth{}>0$ and $z_1, z_2 \in\mathbb{R}^{\ambientdim{}}$, it follows that
    \begin{equation}
        |\GaussianKernel{\bandwidth{}}(z_1) - \GaussianKernel{\bandwidth{}}(z_2)| \leq \frac{\sqrt{2/e}}{\bandwidth{}}\|z_1-z_2\|_2.
    \end{equation}
\end{lemma}

Noise is injected into each data point through a random variable $\noiseRV{i}$, leading to a corrupted dataset $\{\datapoint{1},\ldots, \datapoint{\numpoints{}}\}$ given by
\begin{equation}\label{eq:noise-model}
    \datapoint{i} = \cleandatapoint{i} + \noiseRV{i},\quad 1 \leq i \leq \numpoints{}.
\end{equation}
We assume the $\noiseRV{i}$ are mean-zero, independent and identically distributed, and are sub-Gaussian with variance proxy $\varianceProxy{}\geq 0$, precisely defined after the following remarks.

\begin{remark}\label{rmk:noise}
\begin{enumerate}[leftmargin=*]
    \item The above setup generalizes the case where the clean data points lie on a $\intrinsicdim{}$-dimensional submanifold embedded in $\ambientdim{} = \ambientdimlong{}$-dimensional Euclidean space while the noisy data points are a sample from a tubular neighborhood of the manifold i.e. $\noiseRV{i} = \OBofNSat{i}\auxNoiseRV{i}$ where $\OBofNSat{i}$ is an orthonormal basis of the normal space at $\cleandatapoint{i}$ and $\auxNoiseRV{i} \in \mathbb{R}^{\extradim{}}$ is uniformly distributed in a ball centered at the origin and of radius $\varianceProxy{}$.
    \item If $\noiseRV{i}$ are non-identically distributed sub-Gaussian random variables with variance proxy $\varianceProxy{}_i$ then the following results hold by replacing $\varianceProxy{}$ with $\max_1^\numpoints{} \varianceProxy{}_i$.
\end{enumerate}
\end{remark}

Let $\noiseRV{}\in\mathbb{R}^\ambientdim{}$ be a random vector. We say that $\noiseRV{}$ is a {\normalfont sub-Gaussian random vector} with variance proxy $\varianceProxy{}\geq 0$ and center $\mu\in\mathbb{R}^\ambientdim{}$ if, for all $\alpha\in\mathbb{R}^\ambientdim{}$, it holds
    \begin{equation}\label{eq:sub-Gaussian}
        \ep{\exp(\alpha^\top (\noiseRV{}-\mu))} \leq \exp(\|\alpha\|_2^2\varianceProxy{}^2/2).
    \end{equation}
We say that $\noiseRV{}$ is centered if Eq.~\ref{eq:sub-Gaussian} holds when $\mu=0_\ambientdim{}$.

Eq.~\ref{eq:noise-model} leads to a noisy kernel adjacency matrix, given entrywise by the expression
    \begin{equation}\label{eq:kernel-adj-random-new}
        \adjacency{i}{j} = \GaussianKernel{\bandwidth{}}(\datapoint{i} - \datapoint{j}) = \exp(-\|\datapoint{i}-\datapoint{j}\|_2^2 / \bandwidth{}^2).
    \end{equation}
We investigate the concentration of $\|\adjacency{}{}-\cleanAdjacency{}{}\|$ in the regime where (i) the variance proxy $\varianceProxy{}$ of the noise shrinks as $\numpoints{}$ gets large, and (ii) the feature dimension $\ambientdim{}$ of the data remains fixed.
This regime differs from that of, e.g.,~\cite{ding2022impact}, where the variance proxy remains bounded from below while the feature dimension grows. 
Instead, we make assumptions on the decay of the variance proxy $\varianceProxy{}$ to obtain guarantees on the concentration of $\|\adjacency{}{}-\cleanAdjacency{}{}\|$ at a distance of roughly $\bigO(\numpoints{}^{1/2})$ w.h.p. as $\numpoints{}$ gets large.
Note that we do not need to assume the clean data are uniformly bounded to ensure concentration of $\|\adjacency{}{}-\cleanAdjacency{}{}\|$; however, this will be required later when we need to ensure that degrees are bounded from below.

\begin{theorem}\label{th:concentration-of-a-sub-Gaussian}
    Assume that there exists $\constantA{}>0$ for which $\frac{\varianceProxy{}}{\bandwidth{}} < \frac{\sqrt{\constantA{}}}{\sqrt{\numpoints{}\log{\numpoints{}}}}$, and let $\boundParam{}>2$ be fixed. Then there exists a constant $\constantB{} \equiv \constantB{}(\boundParam{}, \constantA{}) > 0$ such that for $\numpoints{} \geq \nLargeA{}(\ambientdim{}, \boundParam{})$ sufficiently large,
    \begin{align}
        \|\adjacency{}{} - \cleanAdjacency{}{}\|_F\leq \constantB{}\numpoints{}^{1/2} \text{ and } \|\adjacency{}{} - \cleanAdjacency{}{}\|_\infty \leq \constantB{}\numpoints{}^{1/2},
    \end{align}
    each with probability at least $1 - \numpoints{}^{-\boundParam{}+2}$.
\end{theorem}
The proof follows from the Lipschitz continuity of $\GaussianKernel{\bandwidth{}}$ and an application of the well-known tail bound for quadratic forms of sub-Gaussian random vectors (see \cite{hsu2012tail}).

Next, we define the degree of a node $i$ with clean and noisy adjacency matrices as,
\begin{equation}\label{eq:degrees}
    \degree{i} = \sum_{j=1}^\numpoints{} \adjacency{i}{j},\quad \cleanDegree{i} = \sum_{j=1}^\numpoints{} \cleanAdjacency{i}{j}.
\end{equation}
Let $\degreeMat{}, \cleanDegreeMat{}$ to be the diagonal matrices consisting of $\degree{i}$ and $\cleanDegree{i}$, respectively. The following lemma utilizes the uniform boundedness of the clean data to bound the degrees from below. We will use this result to de-randomize bounds in the subsequent results on the stability of the normalized adjacency matrix and the resulting graph Laplacian.
\begin{lemma}\label{lem:degree-bounded-from-zero}
    Assume that there exists $\constantA{}>0$ for which $\frac{\varianceProxy{}}{\bandwidth{}} < \frac{\sqrt{\constantA{}}}{\sqrt{\numpoints{}\log{\numpoints{}}}}$, and let $\boundParam{}>2$ be fixed. Also assume that for some $\radius{}>0$, $\|\cleandatapoint{i}\|_{2}\leq \radius{}$ for all $i \in [1,\numpoints{}]$. Then there exists a positive constant $\constantC{} \equiv \constantC{}(\ambientdim{}, \boundParam{}, \constantA{})$ such that for $\numpoints{}$ sufficiently large,
    \begin{align}
        \min\left\{\degree{\min}, \cleanDegree{\min}\right\} \geq \constantC{} \exp\left(-\frac{8\radius{}^2}{\bandwidth{}^2}\right)\numpoints{}
    \end{align}
    with probability at least $1-\numpoints{}^{-\boundParam{}+2}$, where $\degree{\min} = \min_{i=1}^{\numpoints{}}\degree{i}$ and $\cleanDegree{\min} = \min_{i=1}^{\numpoints{}}\cleanDegree{i}$.
\end{lemma}
We define the clean and noisy normalized adjacency matrices to be
\begin{equation}\label{eq:normalized_adjacency}
    \nCleanAdjacency{}{} = \cleanDegreeMat{}^{-1}\cleanAdjacency{}{}\cleanDegreeMat{}^{-1}, \quad \nAdjacency{}{} = \degreeMat{}^{-1}\adjacency{}{}\degreeMat{}^{-1}.
\end{equation}
Using~\Cref{th:concentration-of-a-sub-Gaussian} and~\Cref{lem:degree-bounded-from-zero}, we obtain concentration of $\|\nAdjacency{}{}-\nCleanAdjacency{}{}\|$ as follows.
\begin{theorem}\label{th:concentration-of-k-sub-Gaussian}
Assume that there exists $\constantA{}>0$ for which $\frac{\varianceProxy{}}{\bandwidth{}} < \frac{\sqrt{\constantA{}}}{\sqrt{\numpoints{}\log{\numpoints{}}}}$, and let $\boundParam{}>2$ be fixed. Also assume that for some $\radius{}>0$, $\|\cleandatapoint{i}\|_{2}\leq \radius{}$ for all $i \in [1,\numpoints{}]$. Then there exists a positive constant $\constantD{} \equiv \constantD{}(\ambientdim{}, \boundParam{}, \constantA{})$ such that for $\numpoints{}$ sufficiently large,
\begin{align}
    \|\nAdjacency{}{} - \nCleanAdjacency{}{}\|_F\leq \constantD{} \exp\left(\frac{32\radius{}^2}{\bandwidth{}^2}\right)\numpoints{}^{-3/2} \text{ and } \|\nAdjacency{}{} - \nCleanAdjacency{}{}\|_\infty\leq \constantD{} \exp\left(\frac{32\radius{}^2}{\bandwidth{}^2}\right)\numpoints{}^{-3/2},
\end{align}
each with probability at least $1 - 2\numpoints{}^{-\boundParam{}+2}$.
\end{theorem}
The proof is an adaptation of the proof of Theorem~4 of Deng, Ling, and Strohmer~\cite{dengStrongConsistencyGraph}. We first show that $\|\nAdjacency{}{} - \nCleanAdjacency{}{}\| \leq  \constantB{} \numpoints{}^{5/2}/\min\left\{  \degree{\min}, \cleanDegree{\min} \right\}^4$ with high probability. This is a randomized bound that depends on $\degree{\min}$, which is de-randomized with high probability using Lemma~\ref{lem:degree-bounded-from-zero}, to obtain the final result. Next, we define the normalized degrees to be
\begin{equation}\label{eq:ndegrees}
    \nDegree{i} = \sum_{j=1}^\numpoints{} \nAdjacency{i}{j},\quad \nCleanDegree{i} = \sum_{j=1}^\numpoints{} \nCleanAdjacency{i}{j}.
\end{equation}
Let $\nDegreeMat{}, \nCleanDegreeMat{}$ be the diagonal matrices consisting of $\nDegree{i}$ and $\nCleanDegree{i}$, respectively. Finally, we define the random-walk graph Laplacians~\cite{hein2007graph,diffusionmaps} on the clean and the noisy data as
\begin{align}\label{eq:rw-laplacians}
    \cleanRWGraphLaplacian{} = \identity{\numpoints{}} - \nCleanDegreeMat{}^{-1}\nCleanAdjacency{}{} ,\quad \RWGraphLaplacian{} = \identity{\numpoints{}} - \nDegreeMat{}^{-1}\nAdjacency{}{}.
\end{align}
The above construction of the graph Laplacian is used in all of our experiments. Using~\Cref{th:concentration-of-k-sub-Gaussian}, we derive the stability of the Laplacian as follows.
\begin{theorem}\label{th:final-concentration-of-l-sub-Gaussian}
Assume that there exists $\constantA{}>0$ for which $\frac{\varianceProxy{}}{\bandwidth{}} < \frac{\sqrt{\constantA{}}}{\sqrt{\numpoints{}\log{\numpoints{}}}}$, and let $\boundParam{}>2$ be fixed. Also assume that for some $\radius{}>0$, $\|\cleandatapoint{i}\|_{2}\leq \radius{}$ for all $i \in [1,\numpoints{}]$. Then there exists a positive constant $\constantE{} \equiv \constantE{}(\ambientdim{}, \boundParam{}, \constantA{})$ such that for $\numpoints{}$ sufficiently large,
\begin{align}
    \|\RWGraphLaplacian{} - \cleanRWGraphLaplacian{}\|_F\leq \constantE{} \exp\left(\frac{48\radius{}^2}{\bandwidth{}^2}\right)\numpoints{}^{-1/2}
\end{align}
with probability at least $1 - 2\numpoints{}^{-\boundParam{}+2}$.
\end{theorem}

It follows from the Davis--Kahan theorem (see \cite{yuUsefulVariantDavis2015}) that as long as the eigengaps of the clean Laplacian $\cleanRWGraphLaplacian{}$ do not vanish too quickly as $\numpoints{}$ gets large, the eigenvectors of $\RWGraphLaplacian{}$ and $\cleanRWGraphLaplacian{}$ will remain close. We state a somewhat simplified version of this below.

\begin{corollary}\label{cor:eigenvector-stability}
Instate the assumptions of~\Cref{th:concentration-of-a-sub-Gaussian} and~\Cref{th:final-concentration-of-l-sub-Gaussian}. Let
\begin{align}
    \symGraphLaplacian{} &= \identity{\numpoints{}} - \nDegreeMat{}^{-1/2}\nAdjacency{}{}\nDegreeMat{}^{-1/2},
    &\cleanSymGraphLaplacian{} &= \identity{\numpoints{}} - \nCleanDegreeMat{}^{-1/2}\nCleanAdjacency{}{}\nCleanDegreeMat{}^{-1/2}
\end{align}
denote the symmetric normalized Laplacians associated with $\RWGraphLaplacian{}$ and $\cleanRWGraphLaplacian{}$.
Enumerate the eigenvalues of $\RWGraphLaplacian{}$ and $\cleanRWGraphLaplacian{}$, respectively, as follows:
    \begin{align}
        \eigval{1}\leq \eigval{2}\leq\dotsc\leq\eigval{\numpoints{}},\\
        \ceigval{1}\leq \ceigval{2}\leq\dotsc\leq\ceigval{\numpoints{}}.
    \end{align}
For each $1\leq i\leq \numpoints{}$, let $\aVector_i$ and $\aCleanVector_i$ be unit-norm eigenvectors of $\symGraphLaplacian{}$ and $\cleanSymGraphLaplacian{}$ corresponding to $\eigval{i}$ and $\ceigval{i}$, respectively, and define
\begin{align}
    \eigvec{i} &= \frac{\nDegreeMat{}^{-1/2}\aVector_i}{\|\nDegreeMat{}^{-1/2}\aVector_i\|_2},
    &\ceigvec{i} &= \frac{\nCleanDegreeMat{}^{-1/2}\aCleanVector_i}{\|\nCleanDegreeMat{}^{-1/2}\aCleanVector_i\|_2}.
\end{align}
Then $\eigvec{i}$ and $\ceigvec{i}$ are unit-norm \emph{right} eigenvectors of $\RWGraphLaplacian{}$ and $\cleanRWGraphLaplacian{}$, respectively.
Let $\numeigforgrad{} \ll \numpoints{}$ be fixed and assume that for each $1\leq i \leq \numeigforgrad{}-1$, it holds that
    \begin{align}
        |\ceigval{i} - \ceigval{i+1}| = \omega(\numpoints{}^{-1/2}).
    \end{align}
Then for each $1\leq i \leq \numeigforgrad{}-1$ fixed, there exists a choice of sign $\tau_i\in\{\pm 1\}$ so that if $\eigvec{i}$ and $\ceigvec{i}$, respectively, denote the eigenvectors of $\RWGraphLaplacian{}$ and $\cleanRWGraphLaplacian{}$ with eigenvalues $\eigval{i}$ and $\ceigval{i}$, then it holds
    \begin{align}
        \|\eigvec{i} - \tau_i\ceigvec{i}\|_2 = \exp\left(\frac{56\radius{}^2}{\bandwidth{}^2}\right)o_{\numpoints{}}(1)
    \end{align}
as $\numpoints{}\to\infty$ with probability at least $1-2\numpoints{}^{-\boundParam{}+2}$, where $o_{\numpoints{}}(1)\to 0$ as $\numpoints{}\to\infty$ (for fixed $\bandwidth{}$).
\end{corollary}

Two remarks are in order. First, there are natural improvements of~\Cref{cor:eigenvector-stability} to settings where as $\numpoints{}$ gets large $\cleanRWGraphLaplacian{}$ picks up eigenvalues with vanishingly small gaps. We would ask instead that the first $\numeigforgrad{}$ eigenvalues of $\cleanRWGraphLaplacian{}$ break into finitely many small groups and the distances between such groups decay at a rate no worse than $\omega(\numpoints{}^{-1/2})$. In this scenario, the distance $\|\eigvec{i} - \tau_i\ceigvec{i}\|_2$ would be replaced with the distance between the corresponding eigenspaces; i.e., convergence of the eigenvectors up to some orthogonal alignment matrix. For simplicity we do not state such improvements here.
Second, the above result is stated for a fixed kernel bandwidth $\bandwidth{}$. If instead $\bandwidth{}$ is allowed to vary with $\numpoints{}$, then maintaining the convergence rate $o_{\numpoints{}}(1)$ requires the bandwidth to decay at a rate of roughly $1/(\log \numpoints{})^{\alpha}$ for some $0<\alpha<1/2$.

\subsection{Convergence of the horizontal component of estimated noisy gradients to clean gradients}
\label{subsec:eig_grad_convergence}

To analyze the stability of eigenvector gradients estimated on noisy data in \ouralgoacronym{}, we adopt the tubular noise model described in~\Cref{subsec:hv_asymptotics} with sufficiently fast shrinking tube width---a specific case of the broader information-plus-noise model considered above. Using 
an assumption based on the eigenvector stability established in~\Cref{cor:eigenvector-stability} and under fast decaying variance proxy, we show that the \horizproxy{} component of the gradient estimated from noisy data converges to its clean counterpart. By employing Tikhonov regularization with $\beta = 1/2$ (which~\Cref{subsec:hv_asymptotics} establishes as the optimal choice for unknown noise level), the following result formalizes this stability. It shows that as the sample size $\numpoints{}$ grows and the neighborhood size $\knnradius{}$ shrinks, a sufficiently small noise perturbation guarantees the convergence of the regularized noisy gradient to the clean gradient.

\begin{proposition}
\label{prop:rmt_eig_grad_low_noise}
Suppose for $\numpoints{}$ sufficiently large, each of the following holds:
\begin{enumerate}
    \item $\varianceProxy{}_{\numpoints{}} = \bigO(\knnradius{3/2+\gamma})$ for a fixed $\gamma \in (0, 1/2)$,
    \item for each $j \in [1,\numpoints{}]$,
    \begin{enumerate}[leftmargin=0.25cm]
        \item $\left\|\ceigvecat{i}{j} - \cceigvecat{i}{j}\right\|_2 < \sqrt{\knn{}}\varianceProxy{}_{\numpoints{}}$, 
        \item there exists $\mathfrak{g}_{ij} \in \mathbb{R}^{\intrinsicdim{}}$ such that $\|\mathfrak{g}_{ij}\|_2 = \bigO(1)$ and
        \begin{equation*}
            \cceigvecat{i}{j} = \ccdatapoint{j}\OBofTSat{j}\mathfrak{g}_{ij} + \xi_j^{\circ 2}
        \end{equation*}
        where $\xi_j^{\circ 2}$ is the elementwise square of $\xi_j \in \mathbb{R}^{\knn{}}$ with entries $\xi_{j_s} = \bigO(\left\|\cleandatapoint{j_s}-\cleandatapoint{j}\right\|_2)$,
        \item $\left\|(\cdatapoint{j}\OBofTSat{j})^T(\cdatapoint{j}\OBofNSat{j})\right\|_2 = o_{\numpoints{}}(\knn{} \knnradius{}\varianceProxy{}_{\numpoints{}})$.
    \end{enumerate}
\end{enumerate}
Then the \horizproxy{} component of the regularized gradient estimate from the noisy data, $\tkvestgradeigvecat{i}{j} = \cdatapoint{j}^{+}\ceigvecat{i}{j}$, converges to that from the clean data, $\tkvestgradceigvecat{i}{j} = \ccdatapoint{j}^{+}\cceigvecat{i}{j}$ as follows,
\begin{align*}
    \left\|\OBofTSat{j}^T\tkvestgradeigvecat{i}{j} - \OBofTSat{j}^T\tkvestgradceigvecat{i}{j}\right\|_2 = \bigO(\knnradius{2\gamma}).
\end{align*}
\end{proposition}

Condition~2(a) strengthens the $o_{\numpoints{}}(1)$ convergence of a noisy eigenvector to its clean counterpart (established in \Cref{cor:eigenvector-stability}) by assuming this rate scales linearly with the variance proxy $\varianceProxy{}_{\numpoints{}}$, which in turn decays sufficiently fast relative to the neighborhood radius $\knnradius{}$, as required by Condition~1. 
At present, it is unclear whether the rate in Condition~2(a) can be derived under the broader information-plus-noise model of~\Cref{cor:eigenvector-stability} without imposing additional constraints on the geometry of the clean data or the distribution of the eigengaps. Establishing such a result is an interesting direction for future work. 

Condition~2(b) relies on the geometric smoothness of the clean eigenvectors, allowing for a first-order Taylor approximation in which the gradient proxy $\mathfrak{g}_{ij}$ remains bounded and independent of the sample size $\numpoints{}$. Finally, Condition~2(c) assumes that the \horizproxy{} and vertical projections of the centered noisy neighborhoods exhibit vanishing empirical correlation. This holds naturally under isotropic noise, and hence under the orthogonal, uniform noise of our tubular noise model. Under these regularity conditions, our result guarantees that the \horizproxy{} component of the noisy gradient converges to that of the clean gradient, where the final convergence rate, $\bigO(\knnradius{2\gamma})$, is driven by how rapidly the variance proxy $\varianceProxy{}_{\numpoints{}}$ decays relative to the local neighborhood radius $\knnradius{}$.

\section{Experiments}
\label{sec:experiments}
In this section, we estimate tangent spaces on several noisy synthetic and real-world datasets using \LPCA{} and \ouralgoacronym{}, compare the estimated tangent spaces against the ground truth, and assess their utility in the following downstream tasks: (a) manifold learning, where we compute an intrinsic-dimensional parameterization of the underlying data manifold; (b) boundary detection, where we identify points that lie on or near the boundary of the data manifold; and (c) local intrinsic dimension estimation, where we determine the dimension of the tangent space at each data point.
For completeness, we briefly describe how the estimated tangent spaces are utilized in these tasks in Appendix~\ref{sec:downstream_tasks}.

To quantify the accuracy of a tangent space estimate $\OBofTSat{j} \in \mathbb{R}^{\ambientdim{} \times \intrinsicdim{}}$ at the $j$-th data point, we compute its deviation from the ground truth tangent space $\OBofTSat{j}^*$, obtained from clean data. Specifically, we calculate the principal angles $\pAngle{j,1}, \ldots, \pAngle{j,\intrinsicdim{}}$ between the subspaces spanned by $\OBofTSat{j}$ and $\OBofTSat{j}^*$~\cite{knyazev2002principal}. The quality of the estimate is measured by the discrepancy score:
\begin{equation}\label{eq:TBDiscrep}
    \TBDiscrep{j} = \sum_{i=1}^{\intrinsicdim{}}(1-\cos(\pAngle{j,i})).
\end{equation}

\subsection{High-aspect ratio Swiss roll and a truncated torus}
We begin with two synthetic datasets: a high-aspect ratio Swiss roll and a truncated torus. For the Swiss roll, we generate $\numpoints{} = 10700$ uniformly distributed points in $\mathbb{R}^3$, forming the clean dataset $\cleandatapoints{}$ (Figure~\ref{fig:swissroll:a}).
Each point is perturbed by adding uniform noise in the direction normal to the underlying tangent space. Specifically, the noisy data points are given by $\datapoint{j} = \cleandatapoint{j} + \eta_j \anormal{}_j$, where $\anormal{}_j$ is the outward normal to the tangent space at $\cleandatapoint{j}$ and the coefficient $\eta_j$ is uniformly distributed in $(-\tubescale{},\tubescale{})$, where $\tubescale{} = 0.0175$. The resulting noisy dataset $\datapoints{}$ is shown in Figure~\ref{fig:swissroll:a}.

For the truncated torus, we sample $\numpoints{} = 3617$ uniformly distributed points on a subset of the torus in $\mathbb{R}^3$ as shown in Figure~\ref{fig:truncted_torus:a}.
Here, each data point $\cleandatapoint{j}$ is parameterized by $(u,v) \in [0,2\pi)^2$ i.e. $\cleandatapoint{j} \equiv \cleandatapoint{j}(u_j, v_j)$. 
We corrupt the clean data with heteroskedastic noise added along the normal direction to the tangent space at each point. 
The noisy data points are given by $\datapoint{j} = \cleandatapoint{j} + \eta_j \anormal{}_j$, where $\anormal{}_j$ is the outward normal direction to the tangent space at $\cleandatapoint{j}$ and the coefficient $\eta_j$ is uniformly distributed in $(-\tubescale(u_j), \tubescale(u_j))$ where $\tubescale(u) = 10^{-2} + 2.5 \times 10^{-2}(1 + \cos(2u))$. The noisy dataset $\datapoints$ is shown in Figure~\ref{fig:truncted_torus:a}.

For both datasets, we estimate an orthonormal basis $\OBofTSat{j}$ of the $2$-dimensional tangent space at each $\datapoint{j}$ using \LPCA{} and \ouralgoacronym{}, and then compute the discrepancy $\TBDiscrep{j}$ (Eq.~\ref{eq:TBDiscrep}) between the estimates $\OBofTSat{j}$ and the ground-truth $\OBofTSat{j}^*$ (Figures~\ref{fig:swissroll:b} and~\ref{fig:truncted_torus:b}).
These results show that \ouralgoacronym{} produces significantly more accurate estimates, while \LPCA{} estimates are highly sensitive to noise. 
Noise ablation (Figure~\ref{fig:noise_ablation}) confirms that \LPCA{} estimates degrade rapidly with noise, whereas \ouralgoacronym{} consistently yields reliable estimates.
Hyperparameter analysis (Figure~\ref{fig:hyperparm_analysis}) also shows that \ouralgoacronym{} estimates remain stable across a broad range of values for $\numeigforgrad{}$ and $\numeig{}$.

\begin{figure}[t!]
    \centering
    \refstepcounter{figure}
    \refstepcounter{subfigure}\label{fig:swissroll:a}
    \refstepcounter{subfigure}\label{fig:swissroll:b}
    \refstepcounter{subfigure}\label{fig:swissroll:c}
    \refstepcounter{subfigure}\label{fig:swissroll:d}
    \refstepcounter{subfigure}\label{fig:swissroll:e}
    \addtocounter{figure}{-1}
    \includegraphics[width=\textwidth]{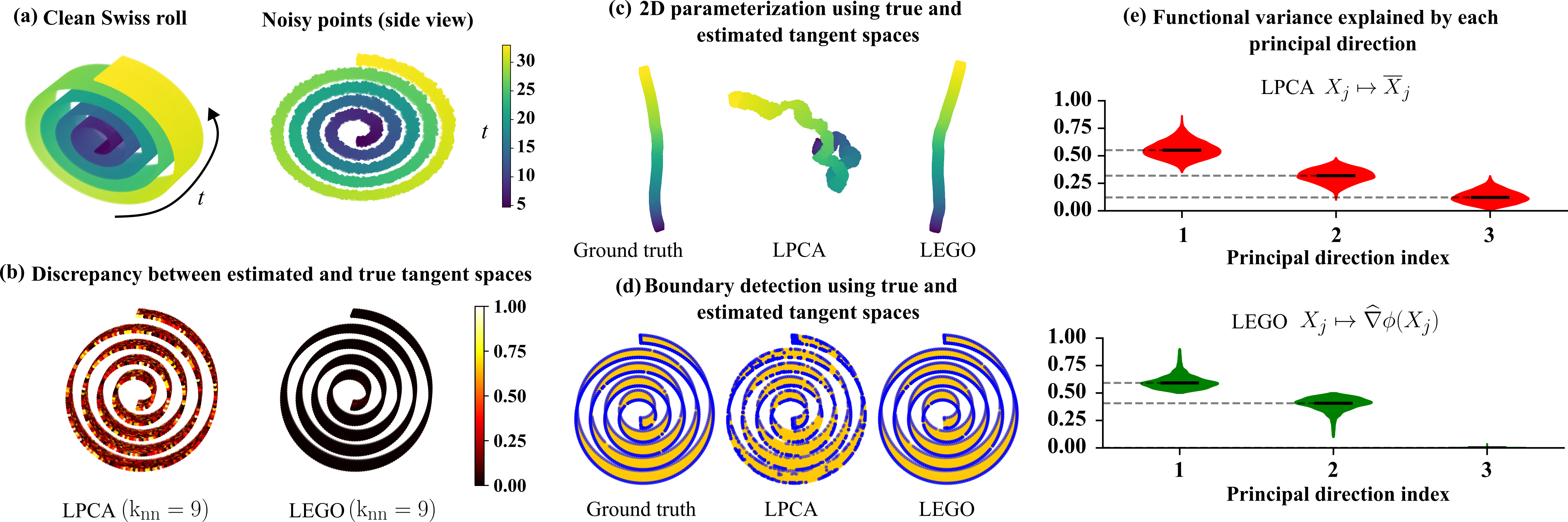}
    \caption{(a) Clean and noisy Swiss roll with high-aspect ratio in $\mathbb{R}^3$ colored by the ``roll'' parameter. (b) Discrepancy between the true and the estimated tangent spaces due to \LPCA{} ($\knn{} = 9$) and \ouralgoacronym{} ($\knn{} = 9$, $\numeig{}=100$, $\numeigforgrad{}=40$, $\beta = 1/2$), as computed using Eq.~\ref{eq:TBDiscrep}. (c, d) $2$-dimensional parameterization of the noisy data, and the boundary points detected from the noisy data using the estimated and the true tangent spaces (see Appendix~\ref{subsec:manifold_learning} and~\ref{subsec:boundary_detection} for details). (e) The functional variance explained by each of the three principal directions in \LPCA{} and \ouralgoacronym{} (see Appendix~\ref{subsec:local_intrinsic_dimension}).}
    \label{fig:swissroll}
\end{figure}

\begin{figure}[t!]
    \centering
    \refstepcounter{figure}
    \refstepcounter{subfigure}\label{fig:truncted_torus:a}
    \refstepcounter{subfigure}\label{fig:truncted_torus:b}
    \refstepcounter{subfigure}\label{fig:truncted_torus:c}
    \refstepcounter{subfigure}\label{fig:truncted_torus:d}
    \refstepcounter{subfigure}\label{fig:truncted_torus:e}
    \addtocounter{figure}{-1}
    \includegraphics[width=\textwidth]{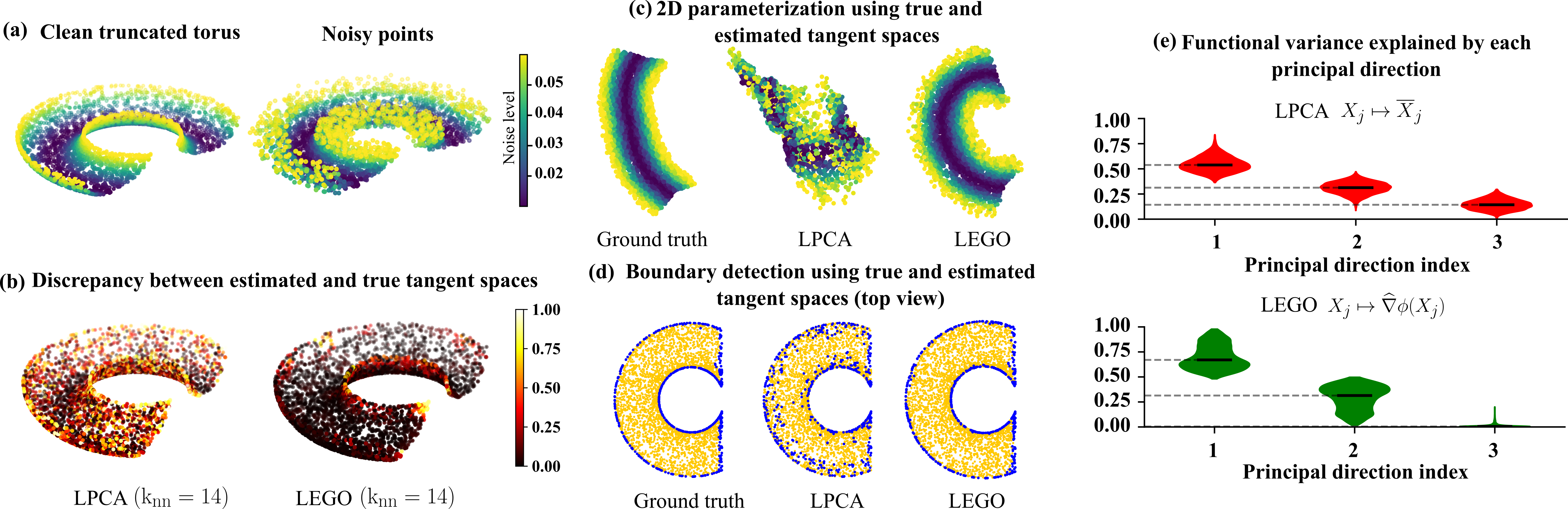}
    \caption{(a) Clean and noisy truncated torus in $\mathbb{R}^3$ colored by the noise level. (b) Discrepancy between the true and the estimated tangent spaces due to \LPCA{} ($\knn{} = 14$) and \ouralgoacronym{} ($\knn{} = 14$, $\numeig{}=100$, $\numeigforgrad{}=20$, $\beta = 1/2$), as computed using Eq.~\ref{eq:TBDiscrep}. (c, d) $2$-dimensional parameterization of the noisy data, and the boundary points detected from the noisy data using the estimated and the true tangent spaces (see Appendix~\ref{subsec:manifold_learning} and~\ref{subsec:boundary_detection} for details) (e) The functional variance explained by each of the three principal directions in \LPCA{} and \ouralgoacronym{} (see Appendix~\ref{subsec:local_intrinsic_dimension}).}
    \label{fig:truncted_torus}
\end{figure}

To assess how these tangent space estimates affect downstream tasks, we use them to compute a $2$-dimensional embedding of the noisy data and to detect boundary points (see Appendix~\ref{sec:downstream_tasks}). As shown in Figure~\ref{fig:swissroll:c} and~\ref{fig:truncted_torus:c}, and Figure~\ref{fig:swissroll:d} and~\ref{fig:truncted_torus:d}, the embeddings and the detected boundary points based on \LPCA{} estimates are severely degraded by noise, while those based on \ouralgoacronym{} closely match the results obtained using the true tangent spaces. This is expected, since the accuracy of the tangent space estimation is critical to the performance of several algorithms~\cite{ltsa,mlle,hessianeigenmaps,ratsv2,berry2017density,robustboundaryv2} designed for these downstream tasks. 

Finally, by setting $\intrinsicdim{} = \ambientdim{} = 3$, we compute the functional variance explained by each principal direction  (Appendix~\ref{subsec:local_intrinsic_dimension}). As shown in Figures~\ref{fig:swissroll:e} and~\ref{fig:truncted_torus:e}, \ouralgoacronym{} concentrates functional variance in the first two directions, consistent with the true intrinsic structure, while \LPCA{} spuriously allocates variance to the third direction, reflecting its sensitivity to noise.

\begin{figure*}[t!]
    \centering
    \refstepcounter{figure}
    \refstepcounter{subfigure}\label{fig:s1puppets:a}
    \refstepcounter{subfigure}\label{fig:s1puppets:b}
    \refstepcounter{subfigure}\label{fig:s1puppets:c}
    \refstepcounter{subfigure}\label{fig:s1puppets:d}
    \refstepcounter{subfigure}\label{fig:s1puppets:e}
    \refstepcounter{subfigure}\label{fig:s1puppets:f}
    \addtocounter{figure}{-1}
    \includegraphics[width=\textwidth]{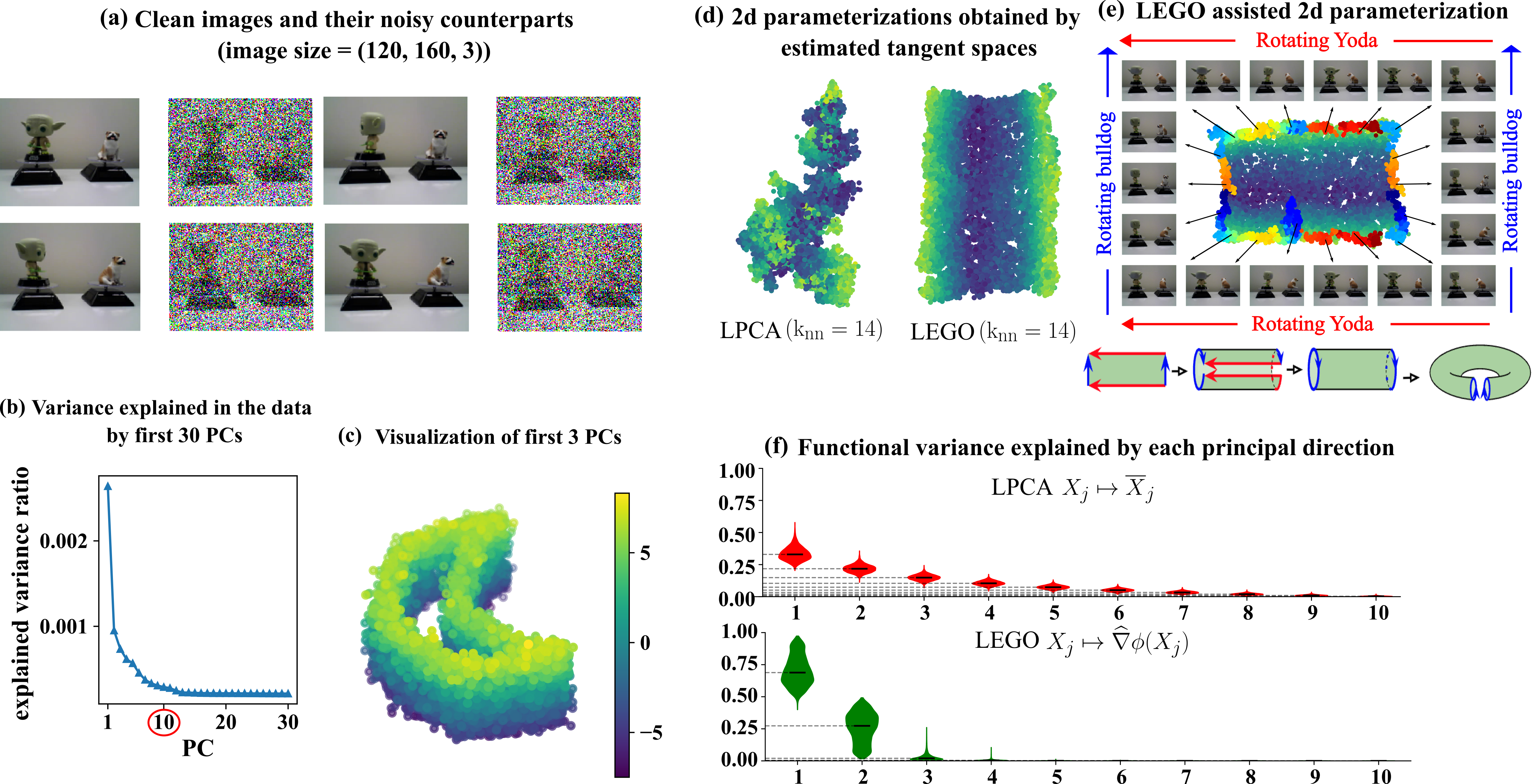}
    \caption{(a) Sample clean images from the Yoda and bulldog dataset~\cite{lederman2018learning} (first and third columns), along with their noise-perturbed versions (second and fourth columns). (b) Explained variance ratio for the first $30$ principal directions obtained via PCA. As the variance saturates after $10$ dimensions, we project the noisy images onto the first $10$ principal components. (c) Visualization of the noisy data using its first three principal components. The colorbar corresponds to the third component. (d) Two-dimensional torn embeddings of the noisy data using the estimated tangent spaces (see Appendix~\ref{subsec:manifold_learning} and~\cite{ratsv2} for details). (e) The torn $2$d embedding obtained using \ouralgoacronym{} estimates, equipped with the gluing instructions that identify the same-colored points along the tear, reveals a toroidal topology. The corresponding clean images along the opposite edges further confirm this structure. (f) Functional variance explained by each of the $10$ principal directions obtained from \LPCA{} and \ouralgoacronym{} (see Appendix~\ref{subsec:local_intrinsic_dimension}).}
    \label{fig:s1puppets}
\end{figure*}

\subsection{Puppets data}
In this real-world experiment, we use an image dataset from~\cite{lederman2018learning}, consisting of $\numpoints{} = 8100$ camera snapshots of a platform with two objects---Yoda and a bulldog---each rotating about its vertical axis at distinct frequencies. As a result, the intrinsic geometry of the dataset corresponds to a $2$-dimensional flat torus. The original images of size $240 \times 320 \times 3$ are first downsampled by a factor of $2$, then normalized to the range $[0,1]$, followed by addition of uniformly distributed noise in $(-1,1)$ to each pixel channel. Examples of both clean and noisy images are shown in Figure~\ref{fig:s1puppets:a} (the pixel values are clipped to $[0,1]$ for visualization). Due to computational constraints, we first reduce the dimensionality of the noisy dataset. Based on the explained variance ratio shown in Figure~\ref{fig:s1puppets:b}, we project the data to $\ambientdim{} = 10$ dimensions, resulting in the final dataset $\datapoints{}$ which is used for tangent space estimation.

We then estimate the $2$-dimensional tangent spaces using both \LPCA{} ($\knn{} = 14$)  and \ouralgoacronym{} ($\knn{} = 14$, $\numeig{}=100$, $\numeigforgrad{}=20$, $\beta = 1/2$). These estimates are used to compute a $2$-dimensional embedding of the noisy data. Because the data lies on a closed manifold, directly aligning the local intrinsic-dimensional embeddings derived from the tangent space estimates using standard methods leads to a collapse---specifically, the resulting intrinsic parameterization is non-injective. To obtain an injective embedding, we adopt the \emph{tear-enabled} alignment framework introduced in~\cite{ratsv2}, which produces a torn $2$d embedding of the data. As shown in Figure~\ref{fig:s1puppets:d}, the embedding based on \LPCA{} estimates is non-interpretable, whereas \ouralgoacronym{} produces a clear rectangular embedding. When visualized with gluing instructions (Figure~\ref{fig:s1puppets:e})---which identify the same-colored points along the tear---it becomes evident that opposite edges of the rectangle should be glued, revealing the underlying toroidal topology. Moreover, examining the clean images corresponding to the points on opposite edges shows that only one of the two puppets undergoes rotation, further supporting the toroidal structure.

Finally, by setting $\intrinsicdim{} = \ambientdim{} = 10$, we compute the functional variance explained by each of the $10$ principal directions obtained by applying \LPCA{} and \ouralgoacronym{} to the noisy data (see  Appendix~\ref{subsec:local_intrinsic_dimension}). As shown in Figure~\ref{fig:s1puppets:f}, \ouralgoacronym{} concentrates the functional variance in the first two directions, faithfully capturing the underlying $2$d structure. In contrast, \LPCA{} distributes the variance across multiple dimensions, highlighting its sensitivity to noise and its inability to accurately recover the local intrinsic geometry in the noisy setting.


\appendix

\section{Overview of downstream tasks involving tangent space estimation}
\label{sec:downstream_tasks}
\subsection{Bottom-up manifold learning}
\label{subsec:manifold_learning}
Given data points $\datapoints{} = [\datapoint{1},\ldots,\datapoint{\numpoints{}}] \in \mathbb{R}^{\ambientdim{} \times \numpoints{}}$, we recover a $\intrinsicdim{}$-dimensional parameterization of the data in three stages~\cite{ratsv2}: First, we estimate an orthonormal basis $\OBofTSat{j} \in \mathbb{R}^{\ambientdim{} \times \intrinsicdim{}}$ of the tangent space at each data point $\datapoint{j}$ (using \LPCA{} or \ouralgoacronym{}) and project neighbors $\{\datapoint{j_s}: j_s \in \nbrhd{j}\}$ of $\datapoint{j}$ to obtain $\intrinsicdim{}$-dimensional local coordinates,

$$\localembed{j_s}{j} = \OBofTSat{j}^T(\datapoint{j_s} - \meanOfNbrhd{j}) \ \text{ where }\ \meanOfNbrhd{j} = \frac{1}{\knn{}}\sum_{s=1}^{\knn{}}\datapoint{j_s}.$$

Second, we align the overlapping local views by estimating an orthogonal matrix $\anOrthMat{j}^* \in \orthMatOf{\intrinsicdim{}}$ and a translation vector $\atranslateVec{j}^* \in \mathbb{R}^{\intrinsicdim{}}$ that minimize the alignment error:

$$\min_{(\anOrthMat{j}, \atranslateVec{j})_{j=1}^{\numpoints{}}} \sum_{i,i'=1}^{\numpoints{}}\sum_{k \in \nbrhd{i} \cap \nbrhd{i'}}\left\|(\anOrthMat{i}^T\localembed{k}{i} + \atranslateVec{i})-(\anOrthMat{i'}^T\localembed{k}{i'} + \atranslateVec{i'})\right\|_2^2.$$

Here, the parameters are initialized using Procrustes analysis and then refined using Riemannian gradient descent (RGD)~\cite{krishnan2005global,ldle2v2,ratsv2}. Finally, the global embedding is obtained by averaging the transformed local coordinates:

$$\globalembed{k} = \frac{\sum_{j: k \in \nbrhd{j}}\anOrthMat{j}^{*T}\localembed{k}{j} + \atranslateVec{j}^*}{|\{j:k \in \nbrhd{j}\}|}$$

Note that when the data lies on a closed manifold---as in the case of the Yoda-Bulldog dataset---the above alignment strategy leads to a collapsed, non-injective embedding into $\mathbb{R}^{\intrinsicdim{}}$~\cite{ratsv2}. To address this, we utilize the tear-enabled rigid alignment framework~\cite{ratsv2}, which tears the manifold to produce a flat $2$d embedding with gluing instructions at the tear (see Figure~\ref{fig:s1puppets:e}).

\begin{figure}[t]
    \centering
    \includegraphics[width=0.65\textwidth]{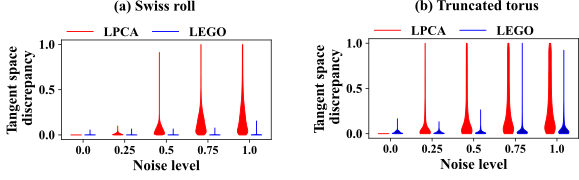}
    \caption{The discrepancy between the true and the estimated tangent spaces from the noisy data $\datapoint{j} = \cleandatapoint{j} + s\eta_j \normalDirn{j}$, $j \in [1,\numpoints{}]$, as the noise level $s$ varies between $0$ and $1$. In our experiments in Section~\ref{sec:experiments}, we used the maximum noise level i.e. $s = 1$.}
    \label{fig:noise_ablation}
\end{figure}
\begin{figure}[t]
    \centering
    \includegraphics[width=0.65\textwidth]{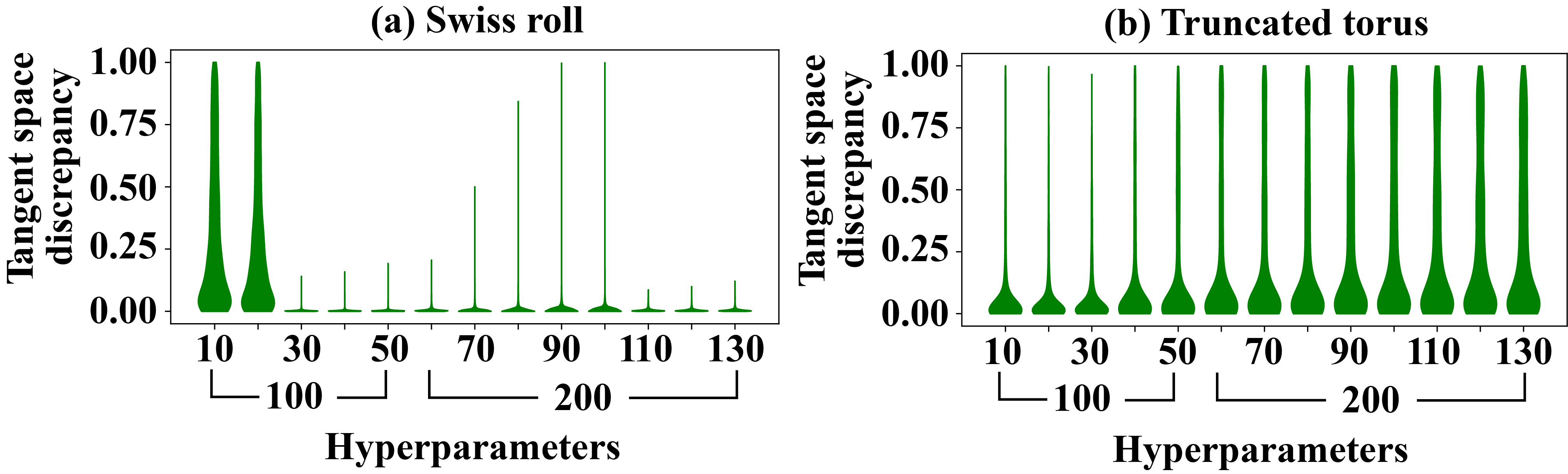}
    \caption{The discrepancy between the true and the estimated tangent spaces on the noisy datasets described in Section~\ref{sec:experiments}, against several different values of the hyperparameters $\numeigforgrad{}$ and $\numeig{}$ in \ouralgoacronym{}, provided at the top and the bottom of the $x$-axis, respectively. The noise level corresponds to the maximum noise in Figure~\ref{fig:noise_ablation}. Note that $\numeigforgrad{}=10$ and $20$ result in high discrepancy for the Swiss roll. This is because of its high aspect ratio which makes the gradients of the first $20$ eigenvectors to be restricted to a one-dimensional subspace.}
    \label{fig:hyperparm_analysis}
\end{figure}

\subsection{Boundary detection}
\label{subsec:boundary_detection}
Building on Berry and Sauer~\cite{berry2017density,VAUGHN2024101593}, we recently developed a robust boundary detection method~\cite{robustboundaryv2} that combines tangent space estimates with a doubly stochastic kernel $\DSKernel{}{}$ derived from a Gaussian kernel via Sinkhorn iterations~\cite{landa2023robust,landa2021doubly}. In our approach, we approximate the normal direction to the boundary $\normalDirn{j}$ at each data point $\datapoint{j}$ by a weighted sum of the projections of the neighbors onto the estimated tangent basis $\OBofTSat{j}$, given by
$\normalDirn{j} \coloneqq \frac{1}{n-1}\sum_{i=1}^{n}\DSKernel{i}{j} \OBofTSat{j}^T(\datapoint{i} - \datapoint{j})$.
Since the vector $\normalDirn{j}$ is nearly zero in the interior, $\datapoint{j}$ is classified as a boundary point if the norm $\|\normalDirn{j}\|_2$ exceeds a fixed percentile threshold (the same percentile is used across all methods for a given dataset).

\subsection{Local intrinsic dimension}
\label{subsec:local_intrinsic_dimension}
To estimate the unknown local intrinsic dimension $\intrinsicdimat{j}$ at $\datapoint{j}$, the approach in \LPCA{} analyzes the eigenvalues $\widetilde{\eigval{}}_{1} \geq \ldots \geq \widetilde{\eigval{}}_{\ambientdim{}}$ of the local covariance matrix $\localCovMat{j} = \sum_{s=1}^{\knn{}}(\datapoint{j_s}-\meanOfNbrhd{j})(\datapoint{j_s}-\meanOfNbrhd{j})^T$ where $\meanOfNbrhd{j} = \frac{1}{\knn{}}\sum_{s=1}^{\knn{}}\datapoint{j_s}$.
The dimension $\intrinsicdimat{j}$ at $\datapoint{j}$ is selected as the smallest index $i$ at which the cumulative explained variance ($\sum_{l=1}^{i}\widetilde{\eigval{}}_{l}/\sum_{k=1}^{\ambientdim{}} \widetilde{\eigval{}}_{k}$) exceeds a user-defined threshold.
Our method adapts this strategy by deriving eigenvalues from the surrogate covariance matrix $\estgradeigvecsat{j}{}\estgradeigvecsat{j}{T}$, constructed from the gradients of low-frequency eigenvectors. 

\section{Proofs from Section~\ref{sec:diff_geom_just}}
\label{sec:tube-eignfns}
\begin{proof}[Proof of Lemma~\ref{lem:pullbackmetric}]
First, we compute the components of the unscaled pullback metric $\metric{}{} = \NBtoRSMap{*}\emetric{\ambientdimlong{}}$ in the local coordinate vector fields defined in Eq.~\ref{eq:coord_vec_fields}, and as derived in~\cite[Lemma~4.1]{haag2015generalised}. Next, we compute the scaled pullback metric $\metric{}{\tubescale{}} = \PBScalingFn{\tubescale{}}\NBtoRSMap{*}\emetric{\ambientdimlong{}}$ by applying the scaling factor of $\tubescale{}$ along the normal coordinate. Subsequently, we introduce a \horizproxy{} lift to block-diagonalize this metric using Gram--Schmidt orthogonalization. Finally, we use this block-diagonal form to decompose the gradient $\RMgrad{}\eigFnLift{}$ into \horizproxy{} and vertical components. For a more detailed exposition of the differential geometry constructs used here, we refer the reader to standard texts (e.g.,~\cite{lee2006riemannian,lee2018introduction}).

The pullback metric $\metric{}{} = \NBtoRSMap{*}\emetric{\ambientdimlong{}}$ is given by
\begin{align*}
    \metric{i,j}{}(\apoint{},\anormalcoord{}) &= \metric{\datamanifold{}}{}(\partial_{\apoint{}^i},\partial_{\apoint{}^j}) - 2\SFF{}_{\anormal{}}(\partial_{\apoint{}^i}, \partial_{\apoint{}^j}) + \metric{\datamanifold{}}{}(\WMap{\anormal{}}(\partial_{\apoint{}^i}),\WMap{\anormal{}}(\partial_{\apoint{}^j})) + \metric{\datamanifold{}}{\perp}(\nabla^{\perp}_{\partial_{\apoint{}^i}}\anormal{}, \nabla^{\perp}_{\partial_{\apoint{}^j}}\anormal{})\\
    \metric{i,\intrinsicdim{}+\alpha}{}(\apoint{},\anormalcoord{}) &= \metric{\datamanifold{}}{\perp}(\nabla^{\perp}_{\partial_{\apoint{}^i}}\anormal{},\StdBasisVec{\alpha})\\
    \metric{\intrinsicdim{}+\alpha,\intrinsicdim{}+\beta}{}(\apoint{},\anormalcoord{}) &= \metric{\datamanifold{}}{\perp}(\StdBasisVec{\alpha},\StdBasisVec{\beta}) = \indicator{\alpha\beta},
\end{align*}
for $i,j \in [1,\intrinsicdim{}]$, $\alpha,\beta \in [1,\extradim{}]$ and where
\begin{enumerate}[leftmargin=*,label=(\roman*)]
    \item $\WMap{\StdBasisVec{\alpha}}: \SmoothVFsOn{\datamanifold{}} \rightarrow \SmoothVFsOn{\datamanifold{}}$ is the Weingarten map that captures the projection of $\nabla^{\mathbb{R}^{\ambientdimlong{}}}_{\partial_{\apoint{}^i}}\StdBasisVec{\alpha}(\apoint{})$ on $\TBof{\datamanifold{}}$ i.e. $\WMap{\StdBasisVec{\alpha}}(\partial_{\apoint{}^i}) = -(\nabla^{\mathbb{R}^{\ambientdimlong{}}}_{\partial_{\apoint{}^i}}\StdBasisVec{\alpha}(\apoint{}))^{\top}$, and
    \item $\nabla^{\perp}:\SmoothVFsOn{\datamanifold{}} \times \SmoothVFsOn{\NBof{\datamanifold{}}} \rightarrow \SmoothSectionOn{\NBof{\datamanifold{}}}$ is the normal connection that captures the projection of $\nabla^{\mathbb{R}^{\ambientdimlong{}}}_{\partial_{\apoint{}^i}}\StdBasisVec{\alpha}(\apoint{})$ on $\NBof{\datamanifold{}}$, and is given by $\nabla^{\perp}_{\partial_{\apoint{}^i}}\StdBasisVec{\alpha}(\apoint{}) = (\nabla^{\mathbb{R}^{\ambientdimlong{}}}_{\partial_{\apoint{}^i}}\StdBasisVec{\alpha}(\apoint{}))^{\perp}$.
\end{enumerate}
Since $\SFF{}_{\anormal{}}(\partial_{\apoint{}^i}, \partial_{\apoint{}^j}) = \metric{\datamanifold{}}{\perp}(\anormal{},\SFF{}(\partial_{\apoint{}^i},\partial_{\apoint{}^j})) =  \metric{\datamanifold{}}{}(\partial_{\apoint{}^i}, \WMap{\anormal{}}(\partial_{\apoint{}^j})) = \metric{\datamanifold{}}{}(\partial_{\apoint{}^j}, \WMap{\anormal{}}(\partial_{\apoint{}^i}))$, therefore, using the definitions of $\CofSSFF{\alpha i}{j}$ and $\CSofNC{i\alpha}{\beta}$ in Eq.~\ref{eq:h_alpha} and~\ref{eq:gamma},
\begin{align*}
    \SFF{}_{\anormal{}}(\partial_{\apoint{}^i}, \partial_{\apoint{}^j})  &= \anormalcoord{}^{\alpha}\CofSSFF{\alpha i}{j} = \anormalcoord{}^{\alpha}\CofSSFF{\alpha j}{i}.\\
    \WMap{\anormal{}}(\partial_{\apoint{}^i}) &= \anormalcoord{}^{\alpha}\metric{\datamanifold{}}{kk'} \CofSSFF{\alpha k'}{i} \partial_{\apoint{}^k}\\
    \nabla^{\perp}_{\partial_{\apoint{}^i}}\anormal{} &= \anormalcoord{}^{\alpha}\CSofNC{i\alpha}{\beta}\StdBasisVec{\beta}.
\end{align*}
Substituting the above equations into the expression for $\metric{i,j}{}(\apoint{},\anormalcoord{})$, we obtain
\begin{align*}
    \metric{i,j}{}(\apoint{},\anormalcoord{}) &= (\metric{\datamanifold{}}{})_{ij} - 2\anormalcoord{}^{\alpha}\CofSSFF{\alpha i}{j} +  \anormalcoord{}^{\alpha}\anormalcoord{}^{\beta}(\metric{\datamanifold{}}{})_{kl}\metric{\datamanifold{}}{kk'}\CofSSFF{\alpha k'}{i}\metric{\datamanifold{}}{ll'}\CofSSFF{\beta l'}{j} + \anormalcoord{}^{\alpha}\anormalcoord{}^{\beta}\CSofNC{i\alpha }{\omega}\CSofNC{j \beta}{\omega'}\indicator{\omega\omega'}\\
    &= (\metric{\datamanifold{}}{})_{ij} - 2\anormalcoord{}^{\alpha}\CofSSFF{\alpha i}{j} +  \anormalcoord{}^{\alpha}\anormalcoord{}^{\beta}\metric{\datamanifold{}}{k'l'}\CofSSFF{\alpha k'}{i}\CofSSFF{\beta l'}{j} + \anormalcoord{}^{\alpha}\anormalcoord{}^{\beta}\CSofNC{i\alpha }{\omega}\CSofNC{j \beta}{\omega'}\indicator{\omega\omega'}\\
    \metric{i,\intrinsicdim{}+\alpha}{}(\apoint{},\anormalcoord{}) &= \anormalcoord{}^{\beta}\CSofNC{i\beta}{\alpha}\\
    \metric{\intrinsicdim{}+\alpha,\intrinsicdim{}+\beta}{}(\apoint{},\anormalcoord{}) &= \indicator{\alpha\beta}.
\end{align*}
Consequently, the scaled pullback metric $\metric{}{\tubescale{}} = \PBScalingFn{\tubescale{}}\NBtoRSMap{*}\emetric{\ambientdimlong{}}$ 
is,
\begin{align}
    \metric{}{\tubescale{}}_{i,j}(\apoint{},\anormalcoord{}) &= (\metric{\datamanifold{}}{})_{ij} - 2\tubescale{} \anormalcoord{}^{\alpha}\CofSSFF{\alpha i}{j} +  \tubescale{}^2\anormalcoord{}^{\alpha}\anormalcoord{}^{\beta}\metric{\datamanifold{}}{k'l'}\CofSSFF{\alpha k'}{i}\CofSSFF{\beta l'}{j} + \tubescale{}^2\anormalcoord{}^{\alpha}\anormalcoord{}^{\beta}\CSofNC{i\alpha }{\omega}\CSofNC{j \beta}{\omega'}\indicator{\omega\omega'}\\
    \metric{}{\tubescale{}}_{i,\intrinsicdim{}+\alpha}(\apoint{},\anormalcoord{}) &= \tubescale{}^2\anormalcoord{}^{\beta}\CSofNC{i\beta}{\alpha}\\
    \metric{}{\tubescale{}}_{\intrinsicdim{}+\alpha,\intrinsicdim{}+\beta}(\apoint{},\anormalcoord{}) &= \tubescale{}^2\indicator{\alpha\beta}
\end{align}
To eliminate the cross-terms and block-diagonalize the metric, we introduce a new basis vector $\partial_{i}^{H}|_{(\apoint{},\anormalcoord{})}$, obtained by projecting $\partial_{i}|_{(\apoint{},\anormalcoord{})}$ orthogonal to the span of $\{\partial_{\intrinsicdim{}+\alpha}|_{(\apoint{},\anormalcoord{})}\}_1^{\extradim{}}$,
\begin{align*}
    \partial_{i}^{H}|_{(\apoint{},\anormalcoord{})} = \partial_{i}|_{(\apoint{},\anormalcoord{})} - \sum_{\alpha=1}^{\extradim{}}\metric{\datamanifold{}}{\perp}(\nabla^{\perp}_{\partial_{\apoint{}^i}}\anormal{},\StdBasisVec{\alpha})\partial_{\intrinsicdim{}+\alpha}|_{(\apoint{},\anormalcoord{})} = \partial_{i}|_{(\apoint{},\anormalcoord{})} - \anormalcoord{}^{\beta}\CSofNC{i\beta}{\alpha}\partial_{\intrinsicdim{}+\alpha}|_{(\apoint{},\anormalcoord{})}.
\end{align*}
In the new local frame $\{\partial_{i}^{H}|_{(\apoint{},\anormalcoord{})}\}_1^d$ and $\{\partial_{\intrinsicdim{}+\alpha}|_{(\apoint{},\anormalcoord{})}\}_1^{\extradim{}}$, the pullback metric $\metric{}{\tubescale{}}$ becomes,
\begin{align*}
    \metric{}{\tubescale{}}_{i,j}(\apoint{},\anormalcoord{}) &= \metric{\datamanifold{}}{}(\partial_{\apoint{}^i},\partial_{\apoint{}^j}) - 2\tubescale{}\SFF{}_{\anormal{}}(\partial_{\apoint{}^i}, \partial_{\apoint{}^j}) + \tubescale{}^2\metric{\datamanifold{}}{}(\WMap{\anormal{}}(\partial_{\apoint{}^i}),\WMap{\anormal{}}(\partial_{\apoint{}^j}))\\
    &= (\metric{\datamanifold{}}{})_{ij} - 2\tubescale{} \anormalcoord{}^{\alpha}\CofSSFF{\alpha i}{j} +  \tubescale{}^2\anormalcoord{}^{\alpha}\anormalcoord{}^{\beta}\metric{\datamanifold{}}{k'l'}\CofSSFF{\alpha k'}{i}\CofSSFF{\beta l'}{j}\\
    \metric{}{\tubescale{}}_{i,\intrinsicdim{}+\alpha}(\apoint{},\anormalcoord{}) &= 0\\
    \metric{}{\tubescale{}}_{\intrinsicdim{}+\alpha,\intrinsicdim{}+\beta}(\apoint{},\anormalcoord{}) &= \tubescale{}^2\metric{\datamanifold{}}{\perp}(\StdBasisVec{\alpha},\StdBasisVec{\beta}) = \tubescale{}^2\indicator{\alpha\beta}.
\end{align*}
Using the definition of $\SSFFMat{\alpha}$ in Eq.~\ref{eq:H_alpha}, we can write $\metric{}{\tubescale{}}$ as a matrix with the basis $\{\partial_{i}^{H}|_{(\apoint{},\anormalcoord{})}\}_1^d$ and $\{\partial_{\intrinsicdim{}+\alpha}|_{(\apoint{},\anormalcoord{})}\}_1^{\extradim{}}$,
\begin{align*}
    \metric{}{\tubescale{}}(\apoint{},\anormalcoord{}) &=
    \begin{bmatrix}
        \metric{\datamanifold{}}{} - 2\tubescale{} \anormalcoord{}^{\alpha}\SSFFMat{\alpha} +  \tubescale{}^2 \anormalcoord{}^{\alpha}\anormalcoord{}^{\beta} \SSFFMat{\alpha} \metric{\datamanifold{}}{-1}\SSFFMat{\beta}& \\
         & \tubescale{}^2 \identity{\extradim{}}
    \end{bmatrix}\\
    &= \begin{bmatrix}
        (\metric{\datamanifold{}}{1/2} - \tubescale{} \anormalcoord{}^{\alpha}\SSFFMat{\alpha}\metric{\datamanifold{}}{-1/2})(\metric{\datamanifold{}}{1/2} - \tubescale{} \metric{\datamanifold{}}{-1/2}\anormalcoord{}^{\beta}\SSFFMat{\beta})& \\
         & \tubescale{}^2 \identity{\extradim{}}
    \end{bmatrix}\\
    &= \begin{bmatrix}
        \metric{\datamanifold{}}{1/2}(\identity{\intrinsicdim{}} - \tubescale{} \anormalcoord{}^{\alpha}\metric{\datamanifold{}}{-1/2}\SSFFMat{\alpha}\metric{\datamanifold{}}{-1/2})^2\metric{\datamanifold{}}{1/2}& \\
         & \tubescale{}^2 \identity{\extradim{}}.
    \end{bmatrix}
\end{align*}
Finally, we decompose $\RMgrad{}\eigFnLift{}$ into a component $(\RMgrad{}\eigFnLift{})^{H}$ on $\projection{}{*}(\TBof{\datamanifold{}})$ and a component $(\RMgrad{}\eigFnLift{})^{V}$ on $\ker(\projection{*}{})$. Specifically, $\RMgrad{}\eigFnLift{} = (\RMgrad{}\eigFnLift{})^{H} + (\RMgrad{}\eigFnLift{})^{V}$ where 
\begin{align*}
    (\RMgrad{}\eigFnLift{})^{H} &= \metric{}{\tubescale{}^{ij}} \partial_{j}^{H}\eigFnLift{}\partial_{i}^{H} = \metric{}{\tubescale{}^{ij}}\left(\frac{\partial \eigFnLift{}}{\partial \apoint{}^j} 
 - \anormalcoord{}^{\beta}\CSofNC{j\beta}{\alpha}\frac{\partial \eigFnLift{}}{\partial \anormalcoord{}^{\alpha}}\right)\partial^{H}_i \text{ and }\\
    (\RMgrad{}\eigFnLift{})^V &=  \metric{}{\tubescale{}^{\intrinsicdim{}+\alpha,\intrinsicdim{}+\beta}}\frac{\partial \eigFnLift{}}{\partial \anormalcoord{}^{\beta}} \partial_{\intrinsicdim{}+\alpha} = \tubescale{}^{-2}\frac{\partial \eigFnLift{}}{\partial \anormalcoord{}^{\alpha}} \partial_{\intrinsicdim{}+\alpha}.
\end{align*}
Using the definition of $\NCMat{\beta}$ in Eq.~\ref{eq:Gamma_beta}, 
\begin{equation}
    \RMgrad{}\eigFnLift{}|_{(\apoint{},\anormalcoord{})} = \begin{bmatrix}
    \metric{\datamanifold{}}{-1/2}(\identity{\intrinsicdim{}} - \tubescale{} \anormalcoord{}^{\alpha}\metric{\datamanifold{}}{-1/2}\SSFFMat{\alpha}\metric{\datamanifold{}}{-1/2})^{-2}\metric{\datamanifold{}}{-1/2}\left(\nabla_{\apoint{}}\eigFnLift{}(\apoint{},\anormalcoord{})-\anormalcoord{}^{\beta}\NCMat{\beta}\nabla_{\anormalcoord{}}\eigFnLift{}(\apoint{},\anormalcoord{})\right)\\
     \tubescale{}^{-2} \nabla_{\anormalcoord{}} \eigFnLift{}(\apoint{},\anormalcoord{})
\end{bmatrix}. \label{eq:gradPsitilde}
\end{equation}
\end{proof}

\begin{proof}[Proof of Lemma~\ref{lem:positivity}]
Using the expression of $\metric{}{\tubescale{}}$ we obtain,

$$\det(\metric{}{\tubescale{}}) = \tubescale{}^{2\extradim{}}\det(\metric{\datamanifold{}}{}) \det\left(\identity{\intrinsicdim{}} - \tubescale{} \anormalcoord{}^{\alpha}\metric{\datamanifold{}}{-1/2}\SSFFMat{\alpha}\metric{\datamanifold{}}{-1/2}\right)^2.$$
Using Cauchy--Schwarz inequality, we obtain

$$\left\|\anormalcoord{}^{\alpha}\metric{\datamanifold{}}{-1/2}\SSFFMat{\alpha}\metric{\datamanifold{}}{-1/2}\right\|_{2} = \sup_{\left\|v\right\|_2=1} |v^T\anormalcoord{}^{\alpha}\metric{\datamanifold{}}{-1/2}\SSFFMat{\alpha}\metric{\datamanifold{}}{-1/2}v| \leq \globalreach{} \absPplCurv{}{}(\apoint{}) \leq \globalreach{} \absPplCurv{}{*}.$$
Since, for each $\apoint{} \in \datamanifold{}$ the maximum value of $\globalreach{} \absPplCurv{}{}(\apoint{})$ can be realized for some $v$ dependent on $\apoint{}$, therefore $\det(\metric{}{\tubescale{}}) > 0$ if and only if $\tubescale{}\globalreach{}\absPplCurv{}{}(x) < 1$. 
Under this constraint, it follows that

$$\tubescale{}^{2\extradim{}}\det(\metric{\datamanifold{}}{})\left(1-\tubescale{} \globalreach{}\absPplCurv{}{*}\right)^{2\intrinsicdim{}} \leq \det(\metric{}{\tubescale{}}) \leq \tubescale{}^{2\extradim{}}\det(\metric{\datamanifold{}}{})\left(1+\tubescale{} \globalreach{}\absPplCurv{}{*}\right)^{2\intrinsicdim{}}.$$
\end{proof}

\begin{proof}[Proof of Theorem~\ref{thm:eval_bounds}]
First note that for $f \in C^{\infty}(\tubeofwidth{\tubescale{} \globalreach{}})$,

$$\int_{\tubeofwidth{\tubescale{} \globalreach{}}} f \volMeasure{\emetric{\ambientdimlong{}}} = \int_{\NBofwidth{\datamanifold{}}{\tubescale{} \globalreach{}}} (\NBtoRSMapLift{}f) \volMeasure{\NBtoRSMap{*}\emetric{\ambientdimlong{}}} = \int_{\NBofwidth{\datamanifold{}}{\globalreach{}}} (\ScalingFnLift{\tubescale{}}{-1}\NBtoRSMapLift{}f) \volMeasure{\PBScalingFn{\tubescale{}}\NBtoRSMap{*}\emetric{\ambientdimlong{}}} = \int_{\NBofwidth{\datamanifold{}}{\globalreach{}}} (\ScalingFnLift{\tubescale{}}{-1}\NBtoRSMapLift{}f) \volMeasure{\metric{}{\tubescale{}}}.$$
Therefore, if $(\eigval{},\eigFn{})$ is an eigenpair of $-\Delta_{\emetric{\ambientdimlong{}}}$ on $\tubeofwidth{\tubescale{} \globalreach{}}$ with Neumann or Dirichlet boundary conditions then  $\eigFnLift{} = \ScalingFnLift{\tubescale{}}{-1}\NBtoRSMapLift{}\eigFn{}$ is an eigenfunction of $-\Delta_{\metric{}{\tubescale{}}}$ with the same eigenvalue. Specifically,

$$\eigval{} = \frac{-\int_{\tubeofwidth{\tubescale{} \globalreach{}}}\eigFn{} \Delta_{\emetric{\ambientdimlong{}}}\eigFn{} \volMeasure{\emetric{\ambientdimlong{}}}}{\int_{\tubeofwidth{\tubescale{} \globalreach{}}}\eigFn{}^2\volMeasure{\emetric{\ambientdimlong{}}}} = \frac{-\int_{\NBofwidth{\datamanifold{}}{\globalreach{}}}\eigFnLift{} \Delta_{\metric{}{\tubescale{}}}\eigFnLift{}\volMeasure{\metric{}{\tubescale{}}}}{\int_{\NBofwidth{\datamanifold{}}{\globalreach{}}}\eigFnLift{}^2\volMeasure{\metric{}{\tubescale{}}}} = \frac{\int_{\NBofwidth{\datamanifold{}}{\globalreach{}}} \langle \RMgrad{}\eigFnLift{},\RMgrad{}\eigFnLift{}\rangle_{\metric{}{\tubescale{}}}\volMeasure{\metric{}{\tubescale{}}}}{\int_{\NBofwidth{\datamanifold{}}{\globalreach{}}}\eigFnLift{}^2\volMeasure{\metric{}{\tubescale{}}}}.$$
Using Lemma~\ref{lem:pullbackmetric}, the definition of $\absPplCurv{}{*}$ in Lemma~\ref{lem:positivity} and Cauchy--Schwarz inequality, we obtain
\begin{align*}
    &\langle \RMgrad{}\eigFnLift{},\RMgrad{}\eigFnLift{}\rangle_{\metric{}{\tubescale{}}} = \RMgrad{}\eigFnLift{}^T\ \metric{}{\tubescale{}} \ \RMgrad{}\eigFnLift{} \geq \frac{\nabla_{\anormalcoord{}} \eigFnLift{}^T\nabla_{\anormalcoord{}} \eigFnLift{}}{\tubescale{}^{2}},
\end{align*}
and the first inequality follows from the definition of $\eigval{}$ and of the normalized vertical energy in Eq.~\ref{eq:normalized_energies}. Next,
\begin{align*}
    &\langle \RMgrad{}\eigFnLift{},\RMgrad{}\eigFnLift{}\rangle_{\metric{}{\tubescale{}}} = \RMgrad{}\eigFnLift{}^T\ \metric{}{\tubescale{}} \ \RMgrad{}\eigFnLift{}\\
    &= { \left(\nabla_{\apoint{}}\eigFnLift{}-\anormalcoord{}^{\beta}\NCMat{\beta}\nabla_{\anormalcoord{}}\eigFnLift{}\right)^T}\metric{\datamanifold{}}{-1/2}(\identity{\intrinsicdim{}} - \tubescale{} \anormalcoord{}^{\alpha}\metric{\datamanifold{}}{-1/2}\SSFFMat{\alpha} \metric{\datamanifold{}}{-1/2})^{-2}\metric{\datamanifold{}}{-1/2}{ \left(\nabla_{\apoint{}}\eigFnLift{}-\anormalcoord{}^{\beta}\NCMat{\beta}\nabla_{\anormalcoord{}}\eigFnLift{}\right)} + { \frac{\nabla_{\anormalcoord{}} \eigFnLift{}^T\nabla_{\anormalcoord{}} \eigFnLift{}}{\tubescale{}^{2}}}\\
    &\geq \frac{1}{(1+ \tubescale{} \globalreach{}\absPplCurv{}{*})^2} \left(\nabla_{\apoint{}}\eigFnLift{}-\anormalcoord{}^{\beta}\NCMat{\beta}\nabla_{\anormalcoord{}}\eigFnLift{}\right)^T\metric{\datamanifold{}}{-1}\left(\nabla_{\apoint{}}\eigFnLift{}-\anormalcoord{}^{\beta}\NCMat{\beta}\nabla_{\anormalcoord{}}\eigFnLift{}\right) + \frac{\nabla_{\anormalcoord{}} \eigFnLift{}^T\nabla_{\anormalcoord{}} \eigFnLift{}}{\tubescale{}^{2}}\\
    &= \frac{1}{(1+ \tubescale{} \globalreach{}\absPplCurv{}{*})^2} \left(\nabla_{\apoint{}}\eigFnLift{}^T\metric{\datamanifold{}}{-1}\nabla_{\apoint{}}\eigFnLift{} + \lengthof{\anormalcoord{}^{\beta}\metric{\datamanifold{}}{-1/2}\NCMat{\beta}\nabla_{\anormalcoord{}}\eigFnLift{}}{\emetric{\intrinsicdim{}}}^2 - 2\anormalcoord{}^{\beta}\nabla_{\apoint{}}\eigFnLift{}^T\metric{\datamanifold{}}{-1}\NCMat{\beta}\nabla_{\anormalcoord{}}\eigFnLift{} \right) + \frac{\nabla_{\anormalcoord{}} \eigFnLift{}^T\nabla_{\anormalcoord{}} \eigFnLift{}}{\tubescale{}^{2}}\\
    &\geq \frac{1}{(1+ \tubescale{} \globalreach{}\absPplCurv{}{*})^2} \left(\nabla_{\apoint{}}\eigFnLift{}^T\metric{\datamanifold{}}{-1}\nabla_{\apoint{}}\eigFnLift{}  - 2\anormalcoord{}^{\beta}\nabla_{\apoint{}}\eigFnLift{}^T\metric{\datamanifold{}}{-1}\NCMat{\beta}\nabla_{\anormalcoord{}}\eigFnLift{} \right) + \frac{\nabla_{\anormalcoord{}} \eigFnLift{}^T\nabla_{\anormalcoord{}} \eigFnLift{}}{\tubescale{}^{2}}\\
    &\geq \frac{1}{(1+ \tubescale{} \globalreach{}\absPplCurv{}{*})^2} \left(\nabla_{\apoint{}}\eigFnLift{}^T\metric{\datamanifold{}}{-1}\nabla_{\apoint{}}\eigFnLift{}  - 2\left(\nabla_{\apoint{}}\eigFnLift{}^T\metric{\datamanifold{}}{-1}\nabla_{\apoint{}}\eigFnLift{}\right)^{1/2}\lengthof{\anormalcoord{}^{\beta}\metric{\datamanifold{}}{-1/2}\NCMat{\beta}\nabla_{\anormalcoord{}}\eigFnLift{}}{\emetric{\intrinsicdim{}}} \right) + \frac{\nabla_{\anormalcoord{}} \eigFnLift{}^T\nabla_{\anormalcoord{}} \eigFnLift{}}{\tubescale{}^{2}}\\
    &\geq \frac{1}{(1+ \tubescale{} \globalreach{}\absPplCurv{}{*})^2} \left(\nabla_{\apoint{}}\eigFnLift{}^T\metric{\datamanifold{}}{-1}\nabla_{\apoint{}}\eigFnLift{}  - 2\left(\nabla_{\apoint{}}\eigFnLift{}^T\metric{\datamanifold{}}{-1}\nabla_{\apoint{}}\eigFnLift{}\right)^{1/2}\left||\anormalcoord{}^{\beta}|\NCCurv{\beta}{}\right|\lengthof{\nabla_{\anormalcoord{}} \eigFnLift{}}{\emetric{\extradim{}}} \right) + \frac{\nabla_{\anormalcoord{}} \eigFnLift{}^T\nabla_{\anormalcoord{}} \eigFnLift{}}{\tubescale{}^{2}}\\
    &\geq \frac{1}{(1+ \tubescale{} \globalreach{}\absPplCurv{}{*})^2} \left(\nabla_{\apoint{}}\eigFnLift{}^T\metric{\datamanifold{}}{-1}\nabla_{\apoint{}}\eigFnLift{}  - 2\globalreach{}\NCCurv{}{*}\left(\nabla_{\apoint{}}\eigFnLift{}^T\metric{\datamanifold{}}{-1}\nabla_{\apoint{}}\eigFnLift{}\right)^{1/2}\lengthof{\nabla_{\anormalcoord{}} \eigFnLift{}}{\emetric{\extradim{}}} \right) + \frac{\nabla_{\anormalcoord{}} \eigFnLift{}^T\nabla_{\anormalcoord{}} \eigFnLift{}}{\tubescale{}^{2}}.
\end{align*}
In the last two equations we used $\lengthof{\anormalcoord{}}{\emetric{\extradim{}}} \leq \globalreach{}$ and the definitions of $\NCCurv{\beta}{}$ and $\NCCurv{}{*}$ in the statement of the theorem. Substituting into the definition of $\eigval{}$ we obtain,
\begin{align*}
    \eigval{} &= \frac{\int_{\NBofwidth{\datamanifold{}}{\globalreach{}}}\langle \RMgrad{}\eigFnLift{},\RMgrad{}\eigFnLift{}\rangle_{\metric{}{\tubescale{}}}\volMeasure{\metric{}{\tubescale{}}}}{\int_{\NBofwidth{\datamanifold{}}{\globalreach{}}} \eigFnLift{}^2\volMeasure{\metric{}{\tubescale{}}}} \geq \frac{\hEnergy{\datamanifold{}}(\eigFn{}) - 2\globalreach{}\NCCurv{}{*}\frac{\int_{\NBofwidth{\datamanifold{}}{\globalreach{}}} \left(\nabla_{\apoint{}}\eigFnLift{}^T\metric{\datamanifold{}}{-1}\nabla_{\apoint{}}\eigFnLift{}\right)^{1/2}\lengthof{\nabla_{\anormalcoord{}} \eigFnLift{}}{\emetric{\extradim{}}}\  \volMeasure{\metric{}{\tubescale{}}}}{\int_{\NBofwidth{\datamanifold{}}{\globalreach{}}}\eigFnLift{}^2 \volMeasure{\metric{}{\tubescale{}}}}}{(1+ \tubescale{} \globalreach{}\absPplCurv{}{*})^2} + \frac{\vEnergy{\datamanifold{}}(\eigFn{})}{\tubescale{}^{2}}\\
    &\geq \frac{\hEnergy{\datamanifold{}}(\eigFn{}) - 2\globalreach{}\NCCurv{}{*}\sqrt{\hEnergy{\datamanifold{}}(\eigFn{})\vEnergy{\datamanifold{}}(\eigFn{})}}{(1+ \tubescale{} \globalreach{}\absPplCurv{}{*})^2} + \frac{\vEnergy{\datamanifold{}}(\eigFn{})}{\tubescale{}^{2}}.
\end{align*}
The result follows from definitions of normalized energies in Eq.~\ref{eq:normalized_energies}. Next,
\begin{align*}
    &\langle \RMgrad{}\eigFnLift{},\RMgrad{}\eigFnLift{}\rangle_{\metric{}{\tubescale{}}} = \RMgrad{}\eigFnLift{}^T\ \metric{}{\tubescale{}} \ \RMgrad{}\eigFnLift{}\\
    &= \left(\nabla_{\apoint{}}\eigFnLift{}-\anormalcoord{}^{\beta}\NCMat{\beta}\nabla_{\anormalcoord{}}\eigFnLift{}\right)^T\metric{\datamanifold{}}{-1/2}(\identity{\intrinsicdim{}} - \tubescale{} \anormalcoord{}^{\alpha}\metric{\datamanifold{}}{-1/2}\SSFFMat{\alpha} \metric{\datamanifold{}}{-1/2})^{-2}\metric{\datamanifold{}}{-1/2}\left(\nabla_{\apoint{}}\eigFnLift{}-\anormalcoord{}^{\beta}\NCMat{\beta}\nabla_{\anormalcoord{}}\eigFnLift{}\right) + \frac{\nabla_{\anormalcoord{}} \eigFnLift{}^T\nabla_{\anormalcoord{}} \eigFnLift{}}{\tubescale{}^{2}}\\
    &\leq \frac{\left(\nabla_{\apoint{}}\eigFnLift{}-\anormalcoord{}^{\beta}\NCMat{\beta}\nabla_{\anormalcoord{}}\eigFnLift{}\right)^T\metric{\datamanifold{}}{-1}\left(\nabla_{\apoint{}}\eigFnLift{}-\anormalcoord{}^{\beta}\NCMat{\beta}\nabla_{\anormalcoord{}}\eigFnLift{}\right)}{(1- \tubescale{} \globalreach{}\absPplCurv{}{*})^2} + \frac{\nabla_{\anormalcoord{}} \eigFnLift{}^T\nabla_{\anormalcoord{}} \eigFnLift{}}{\tubescale{}^{2}}\\
    &= \frac{\left(\metric{\datamanifold{}}{-1/2}\nabla_{\apoint{}}\eigFnLift{}-\anormalcoord{}^{\beta}\metric{\datamanifold{}}{-1/2}\NCMat{\beta}\nabla_{\anormalcoord{}}\eigFnLift{}\right)^T\left(\metric{\datamanifold{}}{-1/2}\nabla_{\apoint{}}\eigFnLift{}-\anormalcoord{}^{\beta}\metric{\datamanifold{}}{-1/2}\NCMat{\beta}\nabla_{\anormalcoord{}}\eigFnLift{}\right)}{(1- \tubescale{} \globalreach{}\absPplCurv{}{*})^2} + \frac{\nabla_{\anormalcoord{}} \eigFnLift{}^T\nabla_{\anormalcoord{}} \eigFnLift{}}{\tubescale{}^{2}}\\
    &\leq \frac{\left(\left(\nabla_{\apoint{}}\eigFnLift{}^T\metric{\datamanifold{}}{-1}\nabla_{\apoint{}}\eigFnLift{}\right)^{1/2} + \lengthof{\anormalcoord{}^{\beta}\metric{\datamanifold{}}{-1/2}\NCMat{\beta}\nabla_{\anormalcoord{}}\eigFnLift{}}{\emetric{\intrinsicdim{}}}\right)^2}{(1- \tubescale{} \globalreach{}\absPplCurv{}{*})^2} + \frac{\nabla_{\anormalcoord{}} \eigFnLift{}^T\nabla_{\anormalcoord{}} \eigFnLift{}}{\tubescale{}^{2}}\\
    &\leq \frac{\left(\left(\nabla_{\apoint{}}\eigFnLift{}^T\metric{\datamanifold{}}{-1}\nabla_{\apoint{}}\eigFnLift{}\right)^{1/2} + ||\anormalcoord{}^{\beta}|\NCCurv{\beta}{}|\left(\nabla_{\anormalcoord{}} \eigFnLift{}^T\nabla_{\anormalcoord{}} \eigFnLift{}\right)^{1/2}\right)^2}{(1- \tubescale{} \globalreach{}\absPplCurv{}{*})^2} + \frac{\nabla_{\anormalcoord{}} \eigFnLift{}^T\nabla_{\anormalcoord{}} \eigFnLift{}}{\tubescale{}^{2}}\\
    &\leq \frac{\left(\left(\nabla_{\apoint{}}\eigFnLift{}^T\metric{\datamanifold{}}{-1}\nabla_{\apoint{}}\eigFnLift{}\right)^{1/2} + (\globalreach{}\NCCurv{}{*})\left(\nabla_{\anormalcoord{}} \eigFnLift{}^T\nabla_{\anormalcoord{}} \eigFnLift{}\right)^{1/2}\right)^2}{(1- \tubescale{} \globalreach{}\absPplCurv{}{*})^2} + \frac{\nabla_{\anormalcoord{}} \eigFnLift{}^T\nabla_{\anormalcoord{}} \eigFnLift{}}{\tubescale{}^{2}}
\end{align*}
Substituting back in the definition of $\eigval{}$, we obtain
\begin{align*}
    \eigval{} &= \frac{\int_{\NBofwidth{\datamanifold{}}{\globalreach{}}}\langle \RMgrad{}\eigFnLift{},\RMgrad{}\eigFnLift{}\rangle_{\metric{}{\tubescale{}}}\volMeasure{\metric{}{\tubescale{}}}}{\int_{\NBofwidth{\datamanifold{}}{\globalreach{}}} \eigFnLift{}^2\volMeasure{\metric{}{\tubescale{}}}} \leq \frac{(\hEnergy{\datamanifold{}}(\eigFn{})^{1/2} + \globalreach{}\NCCurv{}{*}\vEnergy{\datamanifold{}}(\eigFn{})^{1/2})^2}{(1-\tubescale{} \globalreach{}\absPplCurv{}{*})^{2}} + \tubescale{}^{-2}\vEnergy{\datamanifold{}}(\eigFn{}).
\end{align*}
The result follows from the definitions of the normalized energies in Eq.~\ref{eq:normalized_energies}.
\end{proof}

\begin{proof}[Proof of Corollary~\ref{cor:eval_bounds}]
The first inequality follows directly from the previous theorem. Then, using $2\sqrt{ab} \leq \delta a+ \frac{b}{\delta}$ (where $a,b > 0$) for some $\delta > 0$, we obtain
\begin{align*}
    \eigval{} &\geq \frac{\eigval{\datamanifold{}_2}\nhEnergy{\datamanifold{}}(\eigFn{})}{(1+ \tubescale{} \globalreach{}\absPplCurv{}{*})^2} + \frac{\BesselConst{}\nvEnergy{\datamanifold{}}(\eigFn{})}{(\tubescale{}\globalreach{})^{2}} -\frac{2\NCCurv{}{*}\sqrt{\eigval{\datamanifold{}_2}\BesselConst{}\nhEnergy{\datamanifold{}}(\eigFn{})\nvEnergy{\datamanifold{}}(\eigFn{})}}{(1+ \tubescale{} \globalreach{}\absPplCurv{}{*})^2}\\
    &\geq \frac{\eigval{\datamanifold{}_2}(1-\delta)\nhEnergy{\datamanifold{}}(\eigFn{})}{(1+ \tubescale{} \globalreach{}\absPplCurv{}{*})^2} + \left(\frac{1}{(\tubescale{}\globalreach{})^{2}} - \frac{1}{\delta}\left(\frac{\NCCurv{}{*}}{1+ \tubescale{} \globalreach{}\absPplCurv{}{*}}\right)^2\right)\BesselConst{}\nvEnergy{\datamanifold{}}(\eigFn{}).
\end{align*}
Choosing $\delta = \left(\frac{ \tubescale{} \globalreach{}\NCCurv{}{*}}{1+ \tubescale{} \globalreach{}\absPplCurv{}{*}}\right)^2$ which is less than $1$ by assumption, the second inequality follows. Similarly, for the last inequality, 
\begin{align*}
    \eigval{} &\leq \frac{\eigval{\datamanifold{}_2}(1+\delta)\nhEnergy{\datamanifold{}}(\eigFn{})}{(1-\tubescale{} \globalreach{}\absPplCurv{}{*})^{2}} +\left(\left(\frac{\NCCurv{}{*}}{1-\tubescale{} \globalreach{}\absPplCurv{}{*}}\right)^2\left(1+\frac{1}{\delta}\right) + \frac{1}{(\tubescale{}\globalreach{})^{2}}\right)\BesselConst{}\nvEnergy{\datamanifold{}}(\eigFn{})\\
    &\leq \frac{\eigval{\datamanifold{}_2}(1+\delta)\nhEnergy{\datamanifold{}}(\eigFn{})}{(1-\tubescale{} \globalreach{}\absPplCurv{}{*})^{2}} + \left(\left(\frac{\tubescale{}\globalreach{}\NCCurv{}{*}}{1-\tubescale{} \globalreach{}\absPplCurv{}{*}}\right)^2\left(1+\frac{1}{\delta}\right) + 1\right)(1-\eta) \eigval{}
\end{align*}
Therefore,
\begin{align}
    \nhEnergy{\datamanifold{}}(\eigFn{}) \geq \frac{(1-\tubescale{} \globalreach{}\absPplCurv{}{*})^{2}}{1+\delta} \left(1 - \left(\left(\frac{\tubescale{}\globalreach{}\NCCurv{}{*}}{1-\tubescale{} \globalreach{}\absPplCurv{}{*}}\right)^2\left(1+\frac{1}{\delta}\right) + 1\right)(1-\eta)\right) \frac{\eigval{}}{\eigval{\datamanifold{}_2}}.
\end{align}
Now we aim to find the optimal value of $\delta$ that maximizes the lower bound. Note that the non-negative value of $\delta$ that maximizes an expression of the form $f(\delta) = \frac{1}{1+\delta}\left(a + \frac{b}{\delta}\right)$ where $a > 0$ and $b < 0$, is given by $\delta^* = \frac{-b + \sqrt{b(b-a)}}{a}$, and therefore $f(\delta^*) = (\sqrt{a-b} - \sqrt{-b})^2$. Substituting $b = -\left(\frac{\tubescale{}\globalreach{}\NCCurv{}{*}}{1-\tubescale{} \globalreach{}\absPplCurv{}{*}}\right)^2(1-\eta)$ and $a = 1 - \left(\left(\frac{\tubescale{}\globalreach{}\NCCurv{}{*}}{1-\tubescale{} \globalreach{}\absPplCurv{}{*}}\right)^2 + 1\right)(1-\eta) = \eta + b$ (note that $a > 0$ and $b < 0$ by assumption), we obtain $f(\delta^*) = \left(\sqrt{\eta} - \frac{\tubescale{}\globalreach{}\NCCurv{}{*}}{1-\tubescale{} \globalreach{}\absPplCurv{}{*}}\sqrt{1-\eta}\right)^2$. Since $\nhEnergy{\datamanifold{}}(\eigFn{}) \geq (1-\tubescale{} \globalreach{}\absPplCurv{}{*})^{2} f(\delta^*) \frac{\eigval{}}{\eigval{\datamanifold{}_2}}$, the result follows.
\end{proof}

\begin{proof}[Proof of Theorem~\ref{thm:low_eval_exist}]
Using $\nabla_{\anormalcoord{}}\eigFnLift{} = 0$ and Lemma~\ref{lem:positivity}, the proof is similar to that of Theorem~\ref{thm:eval_bounds}. The Rayleigh quotient of the extension satisfies,
\begin{align*}
    \frac{\int_{\tubeofwidth{\tubescale{}\globalreach{}}} |\nabla\eigFn{}|^2 \volMeasure{\emetric{\ambientdimlong{}}}}{\int_{\tubeofwidth{\tubescale{} \globalreach{}}}\eigFn{}^2\volMeasure{\emetric{\ambientdimlong{}}}} = \frac{\int_{\NBofwidth{\datamanifold{}}{\globalreach{}}} \langle \RMgrad{}\eigFnLift{},\RMgrad{}\eigFnLift{}\rangle_{\metric{}{\tubescale{}}}\volMeasure{\metric{}{\tubescale{}}}}{\int_{\NBofwidth{\datamanifold{}}{\globalreach{}}}\eigFnLift{}^2\volMeasure{\metric{}{\tubescale{}}}} = \int_{\NBofwidth{\datamanifold{}}{\globalreach{}}} \langle \RMgrad{}\eigFnLift{},\RMgrad{}\eigFnLift{}\rangle_{\metric{}{\tubescale{}}}\volMeasure{\metric{}{\tubescale{}}}.
\end{align*}
Using $\nabla_{\anormalcoord{}}\eigFnLift{} = 0$ in the lower bound for $\langle \RMgrad{}\eigFnLift{},\RMgrad{}\eigFnLift{}\rangle_{\metric{}{\tubescale{}}}$ derived in the proof of Theorem~\ref{thm:eval_bounds}, followed by Lemma~\ref{lem:positivity}, we obtain
\begin{align*}
    &\int_{\NBofwidth{\datamanifold{}}{\globalreach{}}} \langle \RMgrad{}\eigFnLift{},\RMgrad{}\eigFnLift{}\rangle_{\metric{}{\tubescale{}}}\volMeasure{\metric{}{\tubescale{}}} \geq \frac{\int_{\NBofwidth{\datamanifold{}}{\globalreach{}}} \nabla_{\apoint{}}\eigFnLift{}^T\metric{\datamanifold{}}{-1}\nabla_{\apoint{}}\eigFnLift{}\volMeasure{\metric{}{\tubescale{}}}}{(1 + \tubescale{} \globalreach{}\absPplCurv{}{*})^2} = \frac{1}{(1 + \tubescale{} \globalreach{}\absPplCurv{}{*})^2}\frac{\int_{\NBofwidth{\datamanifold{}}{\globalreach{}}} \nabla_{\apoint{}}\eigFn{}_{\datamanifold{}}^T\metric{\datamanifold{}}{-1}\nabla_{\apoint{}}\eigFn{}_{\datamanifold{}}\volMeasure{\metric{}{\tubescale{}}}}{\int_{\NBofwidth{\datamanifold{}}{\globalreach{}}} (\eigFn{}_{\datamanifold{}} \circ \projection{}{})^2\volMeasure{\metric{}{\tubescale{}}}}\\
    &= \frac{1}{(1 + \tubescale{} \globalreach{}\absPplCurv{}{*})^2}\frac{\int_{\NBofwidth{\datamanifold{}}{\globalreach{}}} \nabla_{\apoint{}}\eigFn{}_{\datamanifold{}}^T\metric{\datamanifold{}}{-1}\nabla_{\apoint{}}\eigFn{}_{\datamanifold{}} \sqrt{\det(\metric{}{\tubescale{}})}d\apoint{}^1\ldots d\apoint{}^{\intrinsicdim{}}d\anormalcoord{}^1\ldots d\anormalcoord{}^{\extradim{}}}{\int_{\NBofwidth{\datamanifold{}}{\globalreach{}}} (\eigFn{}_{\datamanifold{}} \circ \projection{}{})^2 \sqrt{\det(\metric{}{\tubescale{}})}d\apoint{}^1\ldots d\apoint{}^{\intrinsicdim{}}d\anormalcoord{}^1\ldots d\anormalcoord{}^{\extradim{}}}\\
    &\geq \frac{1}{(1 + \tubescale{} \globalreach{}\absPplCurv{}{*})^2}\frac{\int_{\NBofwidth{\datamanifold{}}{\globalreach{}}} \nabla_{\apoint{}}\eigFn{}_{\datamanifold{}}^T\metric{\datamanifold{}}{-1}\nabla_{\apoint{}}\eigFn{}_{\datamanifold{}} \tubescale{}^{\extradim{}}\sqrt{\det(\metric{\datamanifold{}}{})}\left(1-\tubescale{} \globalreach{} \absPplCurv{}{*}\right)^{\intrinsicdim{}} d\apoint{}^1\ldots d\apoint{}^{\intrinsicdim{}}d\anormalcoord{}^1\ldots d\anormalcoord{}^{\extradim{}}}{\int_{\NBofwidth{\datamanifold{}}{\globalreach{}}} (\eigFn{}_{\datamanifold{}} \circ \projection{}{})^2 \tubescale{}^{\extradim{}}\sqrt{\det(\metric{\datamanifold{}}{})}\left(1+\tubescale{} \globalreach{} \absPplCurv{}{*}\right)^{\intrinsicdim{}} d\apoint{}^1\ldots d\apoint{}^{\intrinsicdim{}}d\anormalcoord{}^1\ldots d\anormalcoord{}^{\extradim{}}}\\
    &= \frac{\left(1-\tubescale{} \globalreach{} \absPplCurv{}{*}\right)^{\intrinsicdim{}}}{\left(1+\tubescale{} \globalreach{} \absPplCurv{}{*}\right)^{\intrinsicdim{}+2}}\frac{\int \nabla_{\apoint{}}\eigFn{}_{\datamanifold{}}^T\metric{\datamanifold{}}{-1}\nabla_{\apoint{}}\eigFn{}_{\datamanifold{}} \sqrt{\det(\metric{\datamanifold{}}{})} d\apoint{}^1\ldots d\apoint{}^{\intrinsicdim{}}}{\int \eigFn{}_{\datamanifold{}}^2 \sqrt{\det(\metric{\datamanifold{}}{})} d\apoint{}^1\ldots d\apoint{}^{\intrinsicdim{}}}\\
    &= \frac{\left(1-\tubescale{} \globalreach{} \absPplCurv{}{*}\right)^{\intrinsicdim{}}}{\left(1+\tubescale{} \globalreach{} \absPplCurv{}{*}\right)^{\intrinsicdim{}+2}}\frac{\int \nabla_{\apoint{}}\eigFn{}_{\datamanifold{}}^T\metric{\datamanifold{}}{-1}\nabla_{\apoint{}}\eigFn{}_{\datamanifold{}} \volMeasure{\metric{\datamanifold{}}{}}}{\int \eigFn{}_{\datamanifold{}}^2 \volMeasure{\metric{\datamanifold{}}{}}} = \frac{\left(1-\tubescale{} \globalreach{} \absPplCurv{}{*}\right)^{\intrinsicdim{}}}{\left(1+\tubescale{} \globalreach{} \absPplCurv{}{*}\right)^{\intrinsicdim{}+2}}\mathcal{R}_{\datamanifold{}}(\eigFn{}_{\datamanifold{}})
\end{align*}
Similarly, using $\nabla_{\anormalcoord{}}\eigFnLift{} = 0$ in the upper bound for $\langle \RMgrad{}\eigFnLift{},\RMgrad{}\eigFnLift{}\rangle_{\metric{}{\tubescale{}}}$ derived in the proof of Theorem~\ref{thm:eval_bounds}, and Lemma~\ref{lem:positivity}, the upper bound on $\int_{\NBofwidth{\datamanifold{}}{\globalreach{}}} \langle \RMgrad{}\eigFnLift{},\RMgrad{}\eigFnLift{}\rangle_{\metric{}{\tubescale{}}}\volMeasure{\metric{}{\tubescale{}}}$ follows.
\end{proof}

\begin{proof}[Proof of Corollary~\ref{cor:lowfreqegnfn}]
The min--max principle gives
\[
    \eigval{\ell} = \min_{\substack{W \subset H^1(\tubeofwidth{\tubescale{}\globalreach{}})\\ \dim W = \ell}}\ \max_{\eigFn{} \in W\setminus\{0\}} \frac{\int_{\tubeofwidth{\tubescale{}\globalreach{}}}|\nabla\eigFn{}|^2\volMeasure{\emetric{\ambientdimlong{}}}}{\int_{\tubeofwidth{\tubescale{}\globalreach{}}}\eigFn{}^2\volMeasure{\emetric{\ambientdimlong{}}}}.
\]
Let $\eigFn_{\datamanifold{},1}, \ldots, \eigFn_{\datamanifold{},\ell}$ be orthonormal Neumann eigenfunctions of $-\Delta_{\metric{\datamanifold{}}{}}$ with corresponding eigenvalues $\eigval{\datamanifold{}_1}, \ldots, \eigval{\datamanifold{}_\ell}$. These are smooth by elliptic regularity. By Green's identity and the Neumann condition, $\int_{\datamanifold{}}\langle \RMgrad{}\eigFn_{\datamanifold{},i}, \RMgrad{}\eigFn_{\datamanifold{},k}\rangle_{\metric{\datamanifold{}}{}}\volMeasure{\metric{\datamanifold{}}{}} = \eigval{\datamanifold{}_i}\,\delta_{ik}$, so every nonzero $\eigFn_{\datamanifold{}} = \sum_{i=1}^{\ell} b_i\eigFn_{\datamanifold{},i}$ satisfies
\[
    \mathcal{R}_{\datamanifold{}}(\eigFn_{\datamanifold{}}) = \frac{\sum_{i=1}^{\ell} b_i^2\,\eigval{\datamanifold{}_i}}{\sum_{i=1}^{\ell} b_i^2} \leq \eigval{\datamanifold{}_\ell}.
\]
Let $W \coloneqq \{\NBtoRSMapLift{-1}\ScalingFnLift{\tubescale{}}{}(\eigFn_{\datamanifold{}}\circ\projection{}{}) : \eigFn_{\datamanifold{}} \in \mathrm{span}\{\eigFn_{\datamanifold{},1},\ldots,\eigFn_{\datamanifold{},\ell}\}\}$. Since $\projection{}{}$ is surjective and the lifts are bijections, $\dim W = \ell$. From~\Cref{thm:low_eval_exist}, for every nonzero $\eigFn{} \in W$,
\[
    \frac{\int_{\tubeofwidth{\tubescale{}\globalreach{}}}|\nabla\eigFn{}|^2\volMeasure{\emetric{\ambientdimlong{}}}}{\int_{\tubeofwidth{\tubescale{}\globalreach{}}}\eigFn{}^2\volMeasure{\emetric{\ambientdimlong{}}}} \leq \frac{(1+\tubescale{}\globalreach{}\absPplCurv{}{*})^{\intrinsicdim{}}}{(1-\tubescale{}\globalreach{}\absPplCurv{}{*})^{\intrinsicdim{}+2}}\,\mathcal{R}_{\datamanifold{}}(\eigFn_{\datamanifold{}}) \leq \frac{(1+\tubescale{}\globalreach{}\absPplCurv{}{*})^{\intrinsicdim{}}}{(1-\tubescale{}\globalreach{}\absPplCurv{}{*})^{\intrinsicdim{}+2}}\,\eigval{\datamanifold{}_\ell}.
\]
Using the fact that $\eigval{j} \leq \eigval{\ell}$ whenever $j \leq \ell$, the min--max principle and the above equation, we bound $\lambda_{j}$. Finally, using Eq.~\ref{eq:lambda_lb} of \Cref{thm:eval_bounds}, the result follows.
\end{proof}
\begin{proof}[Proof of~\Cref{prop:vertical_comp_estim_grad}]
For brevity, define $\widetilde{\xi}_{j} = \xi_j^{\circ 2}$ and note that $\left\|\widetilde{\xi}_{j}\right\|_2 = \left\|\xi\right\|_4^2 = \bigO(\sqrt{\knn{}}\knnradius{2})$.
Using Assumption~\ref{assump:eval_bounds},
\begin{align*}
    &\frac{1}{\numpoints{}} \sum_{j=1}^{\numpoints{}}\left\|(\OBofNSat{j})^T\estgradeigvecat{i}{j}\right\|_2^2 = \frac{1}{\numpoints{}} \sum_{j=1}^{\numpoints{}} \left\|(\OBofNSat{j})^T\cdatapoint{j}^{\dagger}\ceigvecat{i}{j}\right\|_2^2\\
    &= \frac{1}{\numpoints{}} \sum_{j=1}^{\numpoints{}} \bigg\{\left\|(\OBofNSat{j})^T\cdatapoint{j}^{\dagger}\cdatapoint{j}\OBofTSat{j}\mathfrak{g}_{ij}  + (\OBofNSat{j})^T\cdatapoint{j}^{\dagger}\cdatapoint{j}\OBofNSat{j}\mathfrak{g}_{ij}^{\perp}\right\|_2^2 +\\
    &\hspace{1.75cm} 2\left((\OBofNSat{j})^T\cdatapoint{j}^{\dagger}\cdatapoint{j}\OBofTSat{j}\mathfrak{g}_{ij}  + (\OBofNSat{j})^T\cdatapoint{j}^{\dagger}\cdatapoint{j}\OBofNSat{j}\mathfrak{g}_{ij}^{\perp}\right)^T\left((\OBofNSat{j})^T\cdatapoint{j}^{\dagger}\widetilde{\xi}_j\right) +  \left\|(\OBofNSat{j})^T\cdatapoint{j}^{\dagger}\widetilde{\xi}_j\right\|_2^2\bigg\}.
\end{align*}
Then, using $\cdatapoint{j}^{\dagger}\cdatapoint{j}\OBofTSat{j}\mathfrak{g}_{ij} = \OBofTSat{j}\mathfrak{g}_{ij}$ (implied by the second assumption) and $(\OBofNSat{j})^T\OBofTSat{j} = 0$, 
\begin{align*}
    (\OBofNSat{j})^T\cdatapoint{j}^{\dagger}\cdatapoint{j}\OBofTSat{j}\mathfrak{g}_{ij}  + (\OBofNSat{j})^T\cdatapoint{j}^{\dagger}\cdatapoint{j}\OBofNSat{j}\mathfrak{g}_{ij}^{\perp} &= (\OBofNSat{j})^T\OBofTSat{j}\mathfrak{g}_{ij} + (\OBofNSat{j})^T\cdatapoint{j}^{\dagger}\cdatapoint{j}\OBofNSat{j}\mathfrak{g}_{ij}^{\perp}\\
    &= (\OBofNSat{j})^T\cdatapoint{j}^{\dagger}\cdatapoint{j}\OBofNSat{j}\mathfrak{g}_{ij}^{\perp}.
\end{align*}
Using the property of orthogonal projectors, we also have $\left\|(\OBofNSat{j})^T\cdatapoint{j}^{\dagger}\cdatapoint{j}\OBofNSat{j}\mathfrak{g}_{ij}^{\perp}\right\|_2 \leq \left\|\mathfrak{g}_{ij}^{\perp}\right\|_2$.
Substituting into the previous equation and using Cauchy--Schwarz inequality,
\begin{align}
    \frac{1}{\numpoints{}} \sum_{j=1}^{\numpoints{}}\left\|(\OBofNSat{j})^T\estgradeigvecat{i}{j}\right\|_2^2 &\leq \frac{1}{\numpoints{}} \sum_{j=1}^{\numpoints{}} \bigg\{\left\|\mathfrak{g}_{ij}^{\perp}\right\|_2^2 + 2\left\|\mathfrak{g}_{ij}^{\perp}\right\|_2\left\|(\OBofNSat{j})^T\cdatapoint{j}^{\dagger}\widetilde{\xi}_j\right\|_2 +  \left\|(\OBofNSat{j})^T\cdatapoint{j}^{\dagger}\widetilde{\xi}_j\right\|_2^2\bigg\}\nonumber\\
    &\leq \left(\left(\frac{1}{n}\sum_{j=1}^{\numpoints{}}\left\|\mathfrak{g}_{ij}^{\perp}\right\|_2^2\right)^{1/2} + \left(\frac{1}{n}\sum_{j=1}^{\numpoints{}}\left\|(\OBofNSat{j})^T\cdatapoint{j}^{\dagger}\widetilde{\xi}_j\right\|_2^2\right)^{1/2}\right)^2 \label{eq:pinv1}
\end{align}
Since $\left\|\OBofNSat{j}\right\|_2 \leq 1$, therefore 
\begin{equation}
    \left\|(\OBofNSat{j})^T\cdatapoint{j}^{\dagger}\widetilde{\xi}_j\right\|_2 \leq \left\|\cdatapoint{j}^{\dagger}\widetilde{\xi}_j\right\|_2 \leq \left\|\cdatapoint{j}^{\dagger}\right\|_2\left\|\widetilde{\xi}_j\right\|_2 = \bigO(\knnradius{2}\min\{\varianceProxy{},\knnradius{}\}^{-1}). \label{eq:pinv_xi_ineq}
\end{equation}
Subsequently, the result follows from the Assumption~\ref{assump:eval_bounds} and the third assumption.

Next,
\begin{align*}
    &\frac{1}{\numpoints{}} \sum_{j=1}^{\numpoints{}}\left\|\OBofTSat{j}^T\estgradeigvecat{i}{j}\right\|_2^2 = \frac{1}{\numpoints{}} \sum_{j=1}^{\numpoints{}} \left\|\OBofTSat{j}^T\cdatapoint{j}^{\dagger}\ceigvecat{i}{j}\right\|_2^2\\
    &= \frac{1}{\numpoints{}} \sum_{j=1}^{\numpoints{}} \bigg\{\left\|\OBofTSat{j}^T\cdatapoint{j}^{\dagger}\cdatapoint{j}\OBofTSat{j}\mathfrak{g}_{ij}  + \OBofTSat{j}^T\cdatapoint{j}^{\dagger}\cdatapoint{j}\OBofNSat{j}\mathfrak{g}_{ij}^{\perp}\right\|_2^2 +\\
    &\hspace{1.75cm} 2\left(\OBofTSat{j}^T\cdatapoint{j}^{\dagger}\cdatapoint{j}\OBofTSat{j}\mathfrak{g}_{ij}  + \OBofTSat{j}^T\cdatapoint{j}^{\dagger}\cdatapoint{j}\OBofNSat{j}\mathfrak{g}_{ij}^{\perp}\right)^T\left(\OBofTSat{j}^T\cdatapoint{j}^{\dagger}\widetilde{\xi}_j\right) + \left\|\OBofTSat{j}^T\cdatapoint{j}^{\dagger}\widetilde{\xi}_j\right\|_2^2\bigg\}.
\end{align*}
Using $\cdatapoint{j}^{\dagger}\cdatapoint{j}\OBofTSat{j} = \OBofTSat{j}$, $(\OBofNSat{j})^T\OBofTSat{j} = 0$ and $\OBofTSat{j}^T\OBofTSat{j} = I_{\intrinsicdim{}}$, we obtain
\begin{align*}
    \OBofTSat{j}^T\cdatapoint{j}^{\dagger}\cdatapoint{j}\OBofTSat{j}\mathfrak{g}_{ij}  + \OBofTSat{j}^T\cdatapoint{j}^{\dagger}\cdatapoint{j}\OBofNSat{j}\mathfrak{g}_{ij}^{\perp} &= \OBofTSat{j}^T\OBofTSat{j}\mathfrak{g}_{ij} + \OBofTSat{j}^T\OBofNSat{j}\mathfrak{g}_{ij}^{\perp} = \mathfrak{g}_{ij}.
\end{align*}
Therefore,
\begin{align*}
    \frac{1}{\numpoints{}} \sum_{j=1}^{\numpoints{}}\left\|\OBofTSat{j}^T\estgradeigvecat{i}{j}\right\|_2^2 &= \frac{1}{\numpoints{}} \sum_{j=1}^{\numpoints{}} \left\|\mathfrak{g}_{ij}\right\|_2^2 + \xi' = \Theta(\eigval{i}) + \xi',
\end{align*}
where using Cauchy--Schwarz inequality,
\begin{align*}
    |\xi'| &\leq  2\left(\frac{1}{n}\sum_{j=1}^{\numpoints{}} \left\|\mathfrak{g}_{ij}\right\|_2^2\right)^{\frac{1}{2}}\left(\frac{1}{n}\sum_{j=1}^{\numpoints{}}\left\|(\OBofTSat{j})^T\cdatapoint{j}^{\dagger}\widetilde{\xi}_j\right\|_2^2\right)^{\frac{1}{2}} +  \frac{1}{n}\sum_{j=1}^{\numpoints{}}\left\|(\OBofTSat{j})^T\cdatapoint{j}^{\dagger}\widetilde{\xi}_j\right\|_2^2
\end{align*}
and the result follows from Assumption~\ref{assump:eval_bounds} and Eq.~\ref{eq:pinv_xi_ineq}.
\end{proof}

\begin{lemma}
\label{lem:tkv_cross_norm}
Let $\delta_{\numpoints{}} = \min\{\varianceProxy{}, \knnradius{}\}$ and $\left\|(\cdatapoint{j}\OBofTSat{j})^T(\cdatapoint{j}\OBofNSat{j})\right\|_2 = o_{\numpoints{}}(\knn{}\knnradius{}\max\{\knnradius{2}, \delta_{\numpoints{}}\})$. Then,
\begin{align*}
    \left\|(\OBofNSat{j})^T(\cdatapoint{j}^T\cdatapoint{j} + \eta_j I_\ambientdim{})^{-1}\OBofTSat{j}\right\|_2 = o_{\numpoints{}}\left(\frac{\max\{\delta_{\numpoints{}},\knnradius{2}\}}{\knn{}\knnradius{}\max\{\delta_{\numpoints{}}^2,  \knnradius{2(1+\beta)}\}}\right)
\end{align*}
and
\begin{align*}
    \left\|(\OBofTSat{j})^T(\cdatapoint{j}^T\cdatapoint{j} + \eta_j I_\ambientdim{})^{-1}A\right\|_2 = \bigO\left(\frac{\left\|\OBofTSat{j}^TA\right\|_2}{\knn{}\knnradius{2}} + \frac{\left\|(\OBofNSat{j})^TA\right\|_2\max\{\delta_{\numpoints{}}, \knnradius{2}\}}{\knn{}\knnradius{}\max\{\delta_{\numpoints{}}^2, \knnradius{2(1+\beta)}\}}\right).
\end{align*}
\end{lemma}
\begin{proof}
For brevity, denote $\cdatapoint{TT} = (\cdatapoint{j}\OBofTSat{j})^T(\cdatapoint{j}\OBofTSat{j}) + \eta_j I_{\intrinsicdim{}}$, $\cdatapoint{NT} = (\cdatapoint{j}\OBofNSat{j})^T(\cdatapoint{j}\OBofTSat{j})$, $\cdatapoint{TN} = \cdatapoint{NT}^T$ and $\cdatapoint{NN} = (\cdatapoint{j}\OBofNSat{j})^T(\cdatapoint{j}\OBofNSat{j}) + \eta_j I_{\extradim{}}$. Then,
\begin{align*}
    &\left\|(\OBofNSat{j})^T(\cdatapoint{j}^T\cdatapoint{j} + \eta_j I_\ambientdim{})^{-1}\OBofTSat{j}\right\|_2 = \left\|(\OBofNSat{j})^T\begin{bmatrix}
        \OBofTSat{j} & \OBofNSat{j} 
    \end{bmatrix} \begin{bmatrix}
        \cdatapoint{TT} & \cdatapoint{TN}\\
        \cdatapoint{NT} & \cdatapoint{NN}
    \end{bmatrix}^{-1} \begin{bmatrix}
        \OBofTSat{j}^T \\ (\OBofNSat{j})^T 
    \end{bmatrix}\OBofTSat{j}\right\|_2\\
    &\hspace{0.25cm}= \left\|\begin{bmatrix}
        0 & I_{\extradim{}} 
    \end{bmatrix} \begin{bmatrix}
        \cdatapoint{TT} & \cdatapoint{TN}\\
        \cdatapoint{NT} & \cdatapoint{NN}
    \end{bmatrix}^{-1} \begin{bmatrix}
        I_{\intrinsicdim{}} \\ 0
    \end{bmatrix}\right\|_2 = \left\|\cdatapoint{NN}^{-1}\cdatapoint{NT}(\cdatapoint{TT} - \cdatapoint{TN}\cdatapoint{NN}^{-1}\cdatapoint{NT})^{-1}\right\|_2
\end{align*}
Note that
\begin{align*}
    \cdatapoint{NT} = (\ccdatapoint{j}\OBofNSat{j})^T(\ccdatapoint{j}\OBofTSat{j}) + (\ccdatapoint{j}\OBofNSat{j})^T(\cnoiseRV{j}\OBofTSat{j}) + (\cnoiseRV{j}\OBofNSat{j})^T(\ccdatapoint{j}\OBofTSat{j}) + (\cnoiseRV{j}\OBofNSat{j})^T(\cnoiseRV{j}\OBofTSat{j}).
\end{align*}
where,
we have $\left\|\ccdatapoint{j}\OBofTSat{j}\right\|_2 = \bigO(\sqrt{\knn{}}\knnradius{})$, $\left\|\ccdatapoint{j}\OBofNSat{j}\right\|_2 = \bigO(\sqrt{\knn{}}\knnradius{2})$, $\left\|\cnoiseRV{j}\OBofNSat{j}\right\|_2 = \bigO\left(\sqrt{\knn{}}\delta_{\numpoints{}}\right)$, $\left\|\cnoiseRV{j}\OBofTSat{j}\right\|_2 = \bigO(\sqrt{\knn{}}\delta_{\numpoints{}}\knnradius{})$, $\sigma_{\min}(\cdatapoint{j}\OBofNSat{j}) = \bigOmega(\sqrt{\knn{}}\delta_{\numpoints{}})$, and $\sigma_{\min}(\cdatapoint{j}\OBofTSat{j}) = \bigOmega(\sqrt{\knn{}}\knnradius{})$. 
These complexity bounds hold up to constants that depend on the geometrical characteristics of the underlying manifold. These are easily derived by analyzing general configuration of local neighborhoods and orthogonal uniform noise~\cite{niyogi2008finding,tyagi2013tangent,singer2012vector}.
Using $\cdatapoint{j} = \ccdatapoint{j} + \cnoiseRV{j}$ and $\eta_j = \Theta(\knn{}\knnradius{2(1+\beta)})$, we obtain
\begin{align}
    \left\|\cdatapoint{NN}^{-1}\right\|_2 &=  \bigO\left(\frac{1}{\knn{}\delta_{\numpoints{}}^2 + \eta_j}\right) = \bigO\left(\frac{1}{\knn{}\max\{\delta_{\numpoints{}}^2,\knnradius{2(1+\beta)}\}}\right)\nonumber\\
    \left\|\cdatapoint{TT}^{-1}\right\|_2 &= \bigO\left(\frac{1}{\knn{}\knnradius{2} + \eta_j}\right) = \bigO\left(\frac{1}{\knn{}\knnradius{2}}\right). \label{eq:tkv5}
\end{align}
Similarly, the deterministic worst-case upper bound of $\left\|\cdatapoint{NT}\right\|_2$ is
\begin{align*}
    \left\|\cdatapoint{NT}\right\|_2    &= \bigO(\knn{}(\knnradius{3} + \delta_{\numpoints{}}\knnradius{3} + \delta_{\numpoints{}}\knnradius{} + \delta_{\numpoints{}}^2\knnradius{})) = \bigO(\knn{}(\knnradius{3} + \delta_{\numpoints{}}\knnradius{})) = \bigO(\knn{}\knnradius{}\max\{\knnradius{2}, \delta_{\numpoints{}}\}),
\end{align*}
which is achieved when the orthogonal noise is highly correlated with the tangential projection of the clean data points. However, the concentration of the zero-mean random orthogonal noise that is independent of the clean data allows us to make a stronger assumption $\left\|\cdatapoint{NT}\right\|_2 = o_{\numpoints{}}(\knn{}\knnradius{}\max\{\knnradius{2}, \delta_{\numpoints{}}\})$, as in the statement of the lemma. Consequently, we have
$$\left\|\cdatapoint{TT}^{-1}\right\|_2\left\|\cdatapoint{NT}\right\|_2^2\left\|\cdatapoint{NN}^{-1}\right\|_2 = \frac{\left\|\cdatapoint{NT}\right\|_2^2}{(\knn{}\knnradius{2})(\knn{}\max\{\delta_{\numpoints{}}^2,\knnradius{2(1+\beta)}\})} = o_{\numpoints{}}(1).$$
Therefore, $\left\|(\cdatapoint{TT} - \cdatapoint{TN}\cdatapoint{NN}^{-1}\cdatapoint{NT})^{-1}\right\|_2 = \bigO\left(\left\|\cdatapoint{TT}^{-1}\right\|_2\right)$ and,
\begin{align*}
    &\left\|(\OBofNSat{j})^T(\cdatapoint{j}^T\cdatapoint{j} + \eta_j I_\ambientdim{})^{-1}\OBofTSat{j}\right\|_2 = \left\|\cdatapoint{NN}^{-1}\cdatapoint{NT}(\cdatapoint{TT} - \cdatapoint{TN}\cdatapoint{NN}^{-1}\cdatapoint{NT})^{-1}\right\|_2\\
    &\leq \left\|\cdatapoint{NN}^{-1}\right\|_2\left\|\cdatapoint{NT}\right\|_2\left\|(\cdatapoint{TT} - \cdatapoint{TN}\cdatapoint{NN}^{-1}\cdatapoint{NT})^{-1}\right\|_2
    = \bigO\left(\left\|\cdatapoint{NN}^{-1}\right\|_2\left\|\cdatapoint{NT}\right\|_2\left\|\cdatapoint{TT}^{-1}\right\|_2\right)\\
    &= o_{\numpoints{}}\left(\frac{1}{\knn{}}\frac{ \knnradius{}\max\{\delta_{\numpoints{}},\knnradius{2}\}}{\knnradius{2}\max\{\delta_{\numpoints{}}^2,  \knnradius{2(1+\beta)}\}}\right) = o_{\numpoints{}}\left(\frac{\max\{\delta_{\numpoints{}},\knnradius{2}\}}{\knn{}\knnradius{}\max\{\delta_{\numpoints{}}^2,  \knnradius{2(1+\beta)}\}}\right).
\end{align*}
The second result follows similarly.
\end{proof}

\begin{proof}[Proof of~\Cref{prop:vertical_comp_reg_grad}]
For brevity, define $\widetilde{\xi}_{j} = \xi_j^{\circ 2}$. Using Assumption~\ref{assump:eval_bounds},
\begin{align}
    &\frac{1}{\numpoints{}} \sum_{j=1}^{\numpoints{}}\left\|(\OBofNSat{j})^T\tkvestgradeigvecat{i}{j}\right\|_2^2 = \frac{1}{\numpoints{}} \sum_{j=1}^{\numpoints{}} \left\|(\OBofNSat{j})^T\cdatapoint{j}^{+}\ceigvecat{i}{j}\right\|_2^2 \nonumber\\
    &= \frac{1}{\numpoints{}} \sum_{j=1}^{\numpoints{}} \bigg\{\left\|(\OBofNSat{j})^T\cdatapoint{j}^{+}\cdatapoint{j}\OBofTSat{j}\mathfrak{g}_{ij}  + (\OBofNSat{j})^T\cdatapoint{j}^{+}\cdatapoint{j}\OBofNSat{j}\mathfrak{g}_{ij}^{\perp}\right\|_2^2 + \label{eq:tkv1}\\
    &\hspace{1.75cm} 2\left((\OBofNSat{j})^T\cdatapoint{j}^{+}\cdatapoint{j}\OBofTSat{j}\mathfrak{g}_{ij}  + (\OBofNSat{j})^T\cdatapoint{j}^{+}\cdatapoint{j}\OBofNSat{j}\mathfrak{g}_{ij}^{\perp}\right)^T\left((\OBofNSat{j})^T\cdatapoint{j}^{+}\widetilde{\xi}_j\right) +  \left\|(\OBofNSat{j})^T\cdatapoint{j}^{+}\widetilde{\xi}_j\right\|_2^2\bigg\}. \nonumber
\end{align}
Using the definition of $\cdatapoint{j}^{+}$, $(\OBofNSat{j})^T\OBofTSat{j} = 0$, and the fact that $\left\|(\OBofNSat{j})^T\cdatapoint{j}^{+}\cdatapoint{j}\OBofNSat{j}\mathfrak{g}_{ij}^{\perp}\right\|_2 \leq \left\|\mathfrak{g}_{ij}^{\perp}\right\|_2$,
\begin{align}
    &\left\|(\OBofNSat{j})^T\cdatapoint{j}^{+}\cdatapoint{j}\OBofTSat{j}\mathfrak{g}_{ij}  + (\OBofNSat{j})^T\cdatapoint{j}^{+}\cdatapoint{j}\OBofNSat{j}\mathfrak{g}_{ij}^{\perp}\right\|_2 \nonumber\\
    &\hspace{1cm}=  \left\|(\OBofNSat{j})^T\OBofTSat{j}\mathfrak{g}_{ij} - \eta_{j}(\OBofNSat{j})^T(\cdatapoint{j}^T\cdatapoint{j} + \eta_j I_\ambientdim{})^{-1}\OBofTSat{j}\mathfrak{g}_{ij}  + (\OBofNSat{j})^T\cdatapoint{j}^{+}\cdatapoint{j}\OBofNSat{j}\mathfrak{g}_{ij}^{\perp}\right\|_2 \nonumber\\
    &\hspace{1cm}=  \left\|- \eta_{j}(\OBofNSat{j})^T(\cdatapoint{j}^T\cdatapoint{j} + \eta_j I_\ambientdim{})^{-1}\OBofTSat{j}\mathfrak{g}_{ij}  + (\OBofNSat{j})^T\cdatapoint{j}^{+}\cdatapoint{j}\OBofNSat{j}\mathfrak{g}_{ij}^{\perp}\right\|_2 \nonumber\\
    &\hspace{1cm}\leq \eta_j\left\|(\OBofNSat{j})^T(\cdatapoint{j}^T\cdatapoint{j} + \eta_j I_\ambientdim{})^{-1}\OBofTSat{j}\mathfrak{g}_{ij}\right\|_2 + \left\|\mathfrak{g}_{ij}^{\perp}\right\|_2\\
    &\hspace{1cm}\leq \eta_j\left\|(\OBofNSat{j})^T(\cdatapoint{j}^T\cdatapoint{j} + \eta_j I_\ambientdim{})^{-1}\OBofTSat{j}\right\|_2\left\|\mathfrak{g}_{ij}\right\|_2 + \left\|\mathfrak{g}_{ij}^{\perp}\right\|_2. \label{eq:tkv2}
\end{align}
Using $\eta_j = \Theta(\knn{}\knnradius{2(1+\beta)})$ and~\Cref{lem:tkv_cross_norm}, 
\begin{align}
    \widetilde{r}_{n} &\coloneqq \eta_j\left\|(\OBofNSat{j})^T(\cdatapoint{j}^T\cdatapoint{j} + \eta_j I_\ambientdim{})^{-1}\OBofTSat{j}\right\|_2 = \bigO\left(\frac{ \knnradius{2(1+\beta)}\max\{\delta_{\numpoints{}}, \knnradius{2}\}}{\knnradius{}\max\{\delta_{\numpoints{}}^2, \knnradius{2(1+\beta)}\}}\right)\nonumber\\
    &= \bigO\left(\frac{ \knnradius{1+2\beta}\max\{\delta_{\numpoints{}}, \knnradius{2}\}}{\max\{\delta_{\numpoints{}}^2, \knnradius{2(1+\beta)}\}}\right) \overset{\varianceProxy{} = \knnradius{a}}{=} \begin{cases}
        \bigO(\knnradius{2\beta}), & a \in (0, 1)\\
        \bigO(\knnradius{1+2\beta-a}), & a \in [1, 1+\beta)\\
        \bigO( \knnradius{a-1}), & a \in [1+\beta,2)\\
        \bigO( \knnradius{}), & a \geq 2.\\
    \end{cases} \label{eq:tkv3}
\end{align}
Substituting back in~\Cref{eq:tkv2}, we obtain
\begin{align*}
    \left\|(\OBofNSat{j})^T\cdatapoint{j}^{+}\cdatapoint{j}\OBofTSat{j}\mathfrak{g}_{ij}  + (\OBofNSat{j})^T\cdatapoint{j}^{+}\cdatapoint{j}\OBofNSat{j}\mathfrak{g}_{ij}^{\perp}\right\|_2 \leq \widetilde{r}_{n}\left\|\mathfrak{g}_{ij}\right\|_2 + \left\|\mathfrak{g}_{ij}^{\perp}\right\|_2.
\end{align*}
Substituting the above equation and $\left\|(\OBofNSat{j})^T\cdatapoint{j}^{+}\widetilde{\xi}_j\right\|_2 \leq \frac{\left\|\widetilde{\xi}_j\right\|_2}{2\sqrt{\eta_j}} = \frac{\left\|\xi_j\right\|_4^2}{2\sqrt{\eta_j}} = \bigO(\knnradius{1-\beta})$ into~\Cref{eq:tkv1}, and applying Cauchy--Schwarz inequality, we obtain the result.

Next, from the definition of $\cdatapoint{j}^{+}$, we obtain $\OBofTSat{j}^T\cdatapoint{j}^{+}\cdatapoint{j}\OBofTSat{j}\mathfrak{g}_{ij} = \mathfrak{g}_{ij} -\eta_j\OBofTSat{j}^T(\cdatapoint{j}^{T}\cdatapoint{j} + \eta_jI_{\ambientdim{}})^{-1}\OBofTSat{j}\mathfrak{g}_{ij}$, and therefore,
\begin{align*}
    &\frac{1}{\numpoints{}} \sum_{j=1}^{\numpoints{}}\left\|\OBofTSat{j}^T\tkvestgradeigvecat{i}{j}\right\|_2^2 = \frac{1}{\numpoints{}} \sum_{j=1}^{\numpoints{}} \left\|\OBofTSat{j}^T\cdatapoint{j}^{+}\ceigvecat{i}{j}\right\|_2^2 \nonumber\\
    &= \frac{1}{\numpoints{}} \sum_{j=1}^{\numpoints{}} \bigg\{\left\|\mathfrak{g}_{ij}\right\|_2^2 + 2\mathfrak{g}_{ij}^T\left(-\eta_j\OBofTSat{j}^T(\cdatapoint{j}^{T}\cdatapoint{j} + \eta_jI_{\ambientdim{}})^{-1}\OBofTSat{j}\mathfrak{g}_{ij} + \OBofTSat{j}^T\cdatapoint{j}^{+}\cdatapoint{j}\OBofNSat{j}\mathfrak{g}_{ij}^{\perp} + \OBofTSat{j}^T\cdatapoint{j}^{+}\widetilde{\xi}_j\right) + \nonumber\\
    &\hspace{2cm} \left\|-\eta_j\OBofTSat{j}^T(\cdatapoint{j}^{T}\cdatapoint{j} + \eta_jI_{\ambientdim{}})^{-1}\OBofTSat{j}\mathfrak{g}_{ij} + \OBofTSat{j}^T\cdatapoint{j}^{+}\cdatapoint{j}\OBofNSat{j}\mathfrak{g}_{ij}^{\perp} + \OBofTSat{j}^T\cdatapoint{j}^{+}\widetilde{\xi}_j\right\|_2^2\bigg\}. \nonumber
\end{align*}
Using~\Cref{eq:tkv3} and $\left\|\OBofTSat{j}^T\cdatapoint{j}^{+}\widetilde{\xi}_j\right\|_2 \leq \frac{\left\|\xi_j\right\|_4^2}{2\sqrt{\eta_j}}$,
\begin{align*}
    &\left\|-\eta_j\OBofTSat{j}^T(\cdatapoint{j}^{T}\cdatapoint{j} + \eta_jI_{\ambientdim{}})^{-1}\OBofTSat{j}\mathfrak{g}_{ij} + \OBofTSat{j}^T\cdatapoint{j}^{+}\cdatapoint{j}\OBofNSat{j}\mathfrak{g}_{ij}^{\perp} + \OBofTSat{j}^T\cdatapoint{j}^{+}\widetilde{\xi}_j\right\|_2\\
    &\hspace{1cm} =\left\|-\eta_j\OBofTSat{j}^T(\cdatapoint{j}^{T}\cdatapoint{j} + \eta_jI_{\ambientdim{}})^{-1}\OBofTSat{j}\mathfrak{g}_{ij} - \eta_j\OBofTSat{j}^T(\cdatapoint{j}^{T}\cdatapoint{j} + \eta_jI_{\ambientdim{}})^{-1}\OBofNSat{j}\mathfrak{g}_{ij}^{\perp} + \OBofTSat{j}^T\cdatapoint{j}^{+}\widetilde{\xi}_j\right\|_2\\
    &\hspace{1cm} \leq \eta_j\left\|\OBofTSat{j}^T(\cdatapoint{j}^{T}\cdatapoint{j} + \eta_jI_{\ambientdim{}})^{-1}\OBofTSat{j}\right\|_2\left\|\mathfrak{g}_{ij}\right\|_2 + \eta_j\left\|\OBofTSat{j}^T(\cdatapoint{j}^{T}\cdatapoint{j} + \eta_jI_{\ambientdim{}})^{-1}\OBofNSat{j}\right\|_2\left\|\mathfrak{g}_{ij}^{\perp}\right\|_2 + \left\|\OBofTSat{j}^T\cdatapoint{j}^{+}\widetilde{\xi}_j\right\|_2\\
    &\hspace{1cm} \leq \eta_j\left\|\OBofTSat{j}^T(\cdatapoint{j}^{T}\cdatapoint{j} + \eta_jI_{\ambientdim{}})^{-1}\OBofTSat{j}\right\|_2\left\|\mathfrak{g}_{ij}\right\|_2 + \widetilde{r}_{\numpoints{}}\left\|\mathfrak{g}_{ij}^{\perp}\right\|_2 + \frac{\left\|\xi_j\right\|_4^2}{2\sqrt{\eta_j}}
\end{align*}
where, using~\Cref{eq:tkv5},
\begin{align*}
    \tilde{r}_{\numpoints{}}' \coloneqq \eta_j\left\|\OBofTSat{j}^T(\cdatapoint{j}^{T}\cdatapoint{j} + \eta_jI_{\ambientdim{}})^{-1}\OBofTSat{j}\right\|_2 = \bigO\left(\frac{\eta_j}{\knn{}\knnradius{2} + \eta_j}\right) = \bigO\left(\knnradius{2\beta}\right).
\end{align*}
Substituting back in the previous equation and using Cauchy--Schwarz inequality,
\begin{align*}
    &\left|\frac{1}{\numpoints{}}\sum_{j=1}^{\numpoints{}}\mathfrak{g}_{ij}^T\left(-\eta_j\OBofTSat{j}^T(\cdatapoint{j}^{T}\cdatapoint{j} + \eta_jI_{\ambientdim{}})^{-1}\OBofTSat{j}\mathfrak{g}_{ij} + \OBofTSat{j}^T\cdatapoint{j}^{+}\cdatapoint{j}\OBofNSat{j}\mathfrak{g}_{ij}^{\perp} + \OBofTSat{j}^T\cdatapoint{j}^{+}\widetilde{\xi}_j\right)\right|\\
    &\hspace{1cm}\leq \frac{1}{\numpoints{}}\sum_{j=1}^{\numpoints{}}\left\|\mathfrak{g}_{ij}\right\|_2\left(\tilde{r}_{\numpoints{}}'\left\|\mathfrak{g}_{ij}\right\|_2 + \widetilde{r}_{\numpoints{}}\left\|\mathfrak{g}_{ij}^{\perp}\right\|_2 + \frac{\left\|\xi_j\right\|_4^2}{2\sqrt{\eta_j}}\right) = \bigO\left(\eigval{i}(\knnradius{2\beta} + \widetilde{r}_{\numpoints{}}\varianceProxy{}^{\noisePowerConst{}}) + \frac{\sqrt{\eigval{i}}\knnradius{1-\beta}}{2}\right)
\end{align*}
and
\begin{align*}
    &\frac{1}{\numpoints{}}\sum_{j=1}^{\numpoints{}}\left\|-\eta_j\OBofTSat{j}^T(\cdatapoint{j}^{T}\cdatapoint{j} + \eta_jI_{\ambientdim{}})^{-1}\OBofTSat{j}\mathfrak{g}_{ij} + \OBofTSat{j}^T\cdatapoint{j}^{+}\cdatapoint{j}\OBofNSat{j}\mathfrak{g}_{ij}^{\perp} + \OBofTSat{j}^T\cdatapoint{j}^{+}\widetilde{\xi}_j\right\|_2^2\\
    &\hspace{1cm}\leq \frac{3}{\numpoints{}}\sum_{j=1}^{\numpoints{}}\left(\left\|\eta_j\OBofTSat{j}^T(\cdatapoint{j}^{T}\cdatapoint{j}+ \eta_jI_{\ambientdim{}})^{-1}\OBofTSat{j}\mathfrak{g}_{ij}\right\|_2^2 +  \left\| \OBofTSat{j}^T\cdatapoint{j}^{+}\cdatapoint{j}\OBofNSat{j}\mathfrak{g}_{ij}^{\perp} \right\|_2^2 + \left\| \OBofTSat{j}^T\cdatapoint{j}^{+}\widetilde{\xi}_j\right\|_2^2\right)\\
    &\hspace{1cm}= \bigO\left(\knnradius{4\beta} \eigval{i} + \widetilde{r}_{\numpoints{}}^2 \eigval{i}\varianceProxy{}^{2\noisePowerConst{}} + \frac{\knnradius{2(1-\beta)}}{4}\right).
\end{align*}
\end{proof}

\section{Proofs from Section~\ref{sec:random-laplacians}}\label{sec:proofs-laplaciannoise}
\begin{proof}[Proof of Lemma~\ref{lem:lipschitz-gaussian}]
Note that $\nabla \GaussianKernel{\bandwidth{}}(z)= -\frac{2}{\bandwidth{}^2}z\GaussianKernel{\bandwidth{}}(z)$. Thus, $\|\nabla\GaussianKernel{\bandwidth{}}(z)\|_2\leq\frac{2}{\bandwidth{}^2}\GaussianKernel{\bandwidth{}}(z) \|z\|_2\leq \frac{\sqrt{2/e}}{\bandwidth{}}$. The claim follows from the properties of Lipschitz functions.
\end{proof}

\begin{lemma}[Specialized version of Theorem 2.1 of \cite{hsu2012tail}]\label{lem:sub-Gaussian-tailbound}
Suppose $\noiseRV{}$ is a centered sub-Gaussian random vector with parameter $\varianceProxy{}\geq 0$. Then for all $t>0$, it holds

$$\pr{\|\noiseRV{}\|_2^2 > \varianceProxy{}^2(\ambientdim{} + 2\sqrt{\ambientdim{}t} + 2t)} \leq e^{-t}.$$
\end{lemma}
\begin{proof}[Proof of Theorem~\ref{th:concentration-of-a-sub-Gaussian}]
We begin by writing

$$\|\adjacency{}{} - \cleanAdjacency{}{}\|_F^2 = \sum_{i,j=1}^{\numpoints{}} (\adjacency{i}{j} - \cleanAdjacency{i}{j})^2 =  \sum_{i,j=1}^{\numpoints{}} |\GaussianKernel{\bandwidth{}}(\datapoint{i} - \datapoint{j}) - \GaussianKernel{\bandwidth{}}(\cleandatapoint{i} - \cleandatapoint{j})|^2.$$
Using Lemma~\ref{lem:lipschitz-gaussian}, we have
$$|\GaussianKernel{\bandwidth{}}(\datapoint{i} - \datapoint{j}) - \GaussianKernel{\bandwidth{}}(\cleandatapoint{i} - \cleandatapoint{j})| \leq \frac{\sqrt{2/e}}{\bandwidth{}} \| (\datapoint{i} - \datapoint{j}) - (\cleandatapoint{i} - \cleandatapoint{j}) \|_2 = \frac{\sqrt{2/e}}{\bandwidth{}}\|\noiseRV{i} - \noiseRV{j}\|_2.$$
Writing $\noiseRV{ij} = \noiseRV{i} - \noiseRV{j}$, we have
\begin{equation}
    \|\adjacency{}{} - \cleanAdjacency{}{}\|_F^2 \leq \frac{2/e}{\bandwidth{}^2 }\sum_{i,j=1}^{\numpoints{}} \|\noiseRV{ij}\|_2^2, \text{ and } \|\adjacency{}{} - \cleanAdjacency{}{}\|_\infty \leq  \frac{\sqrt{2/e}}{\bandwidth{} }\max_{i=1}^{\numpoints{}}\sum_{j=1}^{\numpoints{}} \|\noiseRV{ij}\|_2. \label{eq:conc_eq_1}
\end{equation}
We now bound $\sum_{i,j} \|\noiseRV{ij}\|_2^2$ and $\max_{i}\sum_{j} \|\noiseRV{ij}\|_2$ using the tail bound given in Lemma~\ref{lem:sub-Gaussian-tailbound}. Since $\noiseRV{i}$ and $\noiseRV{j}$ are independent, each $\noiseRV{ij}$ is a centered sub-Gaussian vector with parameter $2\varianceProxy{}$ so that we have, for $t>0$,

$$\pr{\|\noiseRV{ij}\|_2^2 > 4\varianceProxy{}^2(\ambientdim{} + 2\sqrt{\ambientdim{}t} + 2t)} \leq e^{-t}.$$
Therefore, by taking $t = \boundParam{} \log{\numpoints{}}$ in the above inequality and using the union bound, we have
\begin{equation}
    \sum_{i,j=1}^{\numpoints{}} \|\noiseRV{ij}\|_2^2 \leq 4\varianceProxy{}^2 \numpoints{}^2 (\ambientdim{} + 2\sqrt{\boundParam{}\ambientdim{}\log{\numpoints{}}} + 2\boundParam{}\log{\numpoints{}}), \label{eq:conc_eq_2}
\end{equation}
with probability at least $1-\numpoints{}^2e^{-\boundParam{}\log{\numpoints{}}}$, and
\begin{equation}
    \max_{i=1}^{\numpoints{}}\sum_{j=1}^{\numpoints{}} \|\noiseRV{ij}\|_2 \leq 2\varianceProxy{} \numpoints{} (\ambientdim{} + 2\sqrt{\boundParam{}\ambientdim{}\log{\numpoints{}}} + 2\boundParam{}\log{\numpoints{}})^{1/2}, \label{eq:conc_eq_3}
\end{equation}
also with probability at least $1-\numpoints{}^2e^{-\boundParam{}\log{\numpoints{}}}$.
Using $\frac{\varianceProxy{}}{\bandwidth{}}\leq \sqrt{\constantA{}}/\sqrt{\numpoints{}\log{\numpoints{}}}$, we have that for $\numpoints{}$ large enough so that $\boundParam{}\log{\numpoints{}} \geq \max\{\ambientdim{}, 2\sqrt{\boundParam{}\ambientdim{}\log{\numpoints{}}}\}$, it holds

$$\frac{4\varianceProxy{}^2}{\bandwidth{}^2} \numpoints{}^2 (\ambientdim{} + 2\sqrt{\boundParam{}\ambientdim{}\log{\numpoints{}}} + 2\boundParam{}\log{\numpoints{}}) \leq 16 \constantA{} \boundParam{} \numpoints{}.$$
Substituting in Eq~\ref{eq:conc_eq_2} and~\ref{eq:conc_eq_3}, and using Eq~\ref{eq:conc_eq_1}, the result follows.
\end{proof}

\begin{proof}[Proof of Lemma~\ref{lem:degree-bounded-from-zero}]
We proceed with a two-step argument. First, we show that with high probability, $\degree{\min} \geq C \cleanDegree{min}$ for some constant $C\in(0, 1)$, and then we show that $\cleanDegree{min}$ is bounded from below in general. To this end, note that since

$$\|\cleandatapoint{i} - \cleandatapoint{j} + \noiseRV{i} - \noiseRV{j}\|_2^2\leq 2(\|\cleandatapoint{i} - \cleandatapoint{j}\|_2^2 + \|\noiseRV{i} - \noiseRV{j}\|_2^2),$$
we have, for $i$ fixed,
\begin{align*}
    \degree{i} &= \sum_{j=1}^\numpoints{} \GaussianKernel{\bandwidth{}}(\datapoint{i} - \datapoint{j}) = \sum_{j=1}^\numpoints{} \exp(- \|\cleandatapoint{i} - \cleandatapoint{j} + \noiseRV{i} - \noiseRV{j}\|_2^2 / \bandwidth{}^2)\\
    &\geq  \sum_{j=1}^\numpoints{} \exp(- 2\|\cleandatapoint{i} - \cleandatapoint{j}\|_2^2 / \bandwidth{}^2)\exp(- 2\|\noiseRV{i} - \noiseRV{j}\|_2^2 / \bandwidth{}^2)\\
    &\geq \sum_{j=1}^\numpoints{} \exp(- \|\cleandatapoint{i} - \cleandatapoint{j}\|_2^2 / \bandwidth{}^2) \times \min_{1\leq j \leq \numpoints{}}\exp(- \|\cleandatapoint{i} - \cleandatapoint{j}\|_2^2 / \bandwidth{}^2)\times\min_{1\leq j \leq \numpoints{}}\exp(- 2\|\noiseRV{i} - \noiseRV{j}\|_2^2 / \bandwidth{}^2) .
\end{align*}
Taking each term one by one, we first have $\sum_{j=1}^\numpoints{} \exp(- \|\cleandatapoint{i} - \cleandatapoint{j}\|_2^2 / \bandwidth{}^2) = \cleanDegree{i}$. Then, we note that since $\cleandatapoint{i}, \cleandatapoint{j}\in B(0, R)$ where $B(0, R)$ is the ball of radius $R>0$ in $\mathbb{R}^\ambientdim{}$, it holds

$$\exp(- \|\cleandatapoint{i} - \cleandatapoint{j}\|_2^2 / \bandwidth{}^2) \geq \exp(-4 R^2/ \bandwidth{}^2),$$
and therefore
\begin{equation}\label{eq:second-term}
    \min_{1\leq j \leq \numpoints{}}\exp(- \|\cleandatapoint{i} - \cleandatapoint{j}\|_2^2 / \bandwidth{}^2)  \geq \exp(-4 R^2/ \bandwidth{}^2).
\end{equation}
Finally, we focus on the third term. Letting $i, j$ be fixed, note that $\noiseRV{i}-\noiseRV{j}$ is a mean zero centered sub-Gaussian with variance proxy $2\varianceProxy{}$. Therefore, using Lemma~\ref{lem:sub-Gaussian-tailbound}, for $t>0$, 

$$\pr{\|\noiseRV{i} - \noiseRV{j}\|_2^2 > 4\varianceProxy{}^2(\ambientdim{} + 2\sqrt{\ambientdim{}t} + 2t)} \leq e^{-t}.$$
Picking $t=\boundParam{}\log{\numpoints{}}$, we have that by the union bound,

$$\|\noiseRV{i} - \noiseRV{j}\|_2^2\leq 4\varianceProxy{}^2(\ambientdim{} + 2\sqrt{\ambientdim{} \boundParam{}\log{\numpoints{}}} + 2\boundParam{}\log{\numpoints{}})$$
for all $j \in [1,\numpoints{}]$ with probability at least $1-\numpoints{}^{-\boundParam{}+1}$.
Picking $\numpoints{}$ large enough so that $\boundParam{}\log{\numpoints{}}\geq \max\{\ambientdim{}, 2\sqrt{\ambientdim{} \boundParam{}\log{\numpoints{}}}\}$ (as in the proof of Theorem~\ref{th:concentration-of-a-sub-Gaussian}) and $\frac{\varianceProxy{}}{\bandwidth{}}\leq \sqrt{\constantA{}}/\sqrt{\numpoints{}\log{\numpoints{}}}$, it holds

$$4\frac{\varianceProxy{}^2}{\bandwidth{}^2}(\ambientdim{} + 2\sqrt{\ambientdim{} \boundParam{}\log{\numpoints{}}} + 2\boundParam{}\log{\numpoints{}}) \leq \frac{16\constantA{}\boundParam{}}{\numpoints{}}.$$
Thus, with probability at least $1-\numpoints{}^{-\boundParam{}+1}$, and since $\numpoints{}\geq e^{\ambientdim{}/\boundParam{}}$ as before,

$$\min_{1\leq j \leq \numpoints{}}\exp(- 2\|\noiseRV{i} - \noiseRV{j}\|_2^2 / \bandwidth{}^2) \geq \exp\left(-\frac{32 \constantA{}\boundParam{} }{\numpoints{}}\right)\geq \exp\left(-\frac{32 \constantA{}\boundParam{} }{e^{\ambientdim{}/\boundParam{}}}\right).$$
Thus, if we put $c_1 =  \exp\left(-\frac{32 \constantA{}\boundParam{} }{e^{\ambientdim{}/\boundParam{}}}\right)$,
then $0<c_1 <1$ and with probability at least $1-\numpoints{}^{-\boundParam{}+1}$, $\degree{i} \geq c_1 \exp(-4 R^2/ \bandwidth{}^2) \cleanDegree{i}$. Therefore, by the union bound, with probability at least $1-\numpoints{}^{-\boundParam{}+2}$, it holds $\degree{\min} \geq c_1\exp(-4 R^2/ \bandwidth{}^2)\cleanDegree{\min}$. Note that using Eq~\ref{eq:second-term},

$$\cleanDegree{\min} \geq \numpoints{} \min_{1\leq i, j \leq \numpoints{}}\exp(- \|\cleandatapoint{i} - \cleandatapoint{j}\|_2^2 / \bandwidth{}^2) \geq \numpoints{}  \exp(-4 R^2/ \bandwidth{}^2),$$
Therefore, with high probability, $\min\{\degree{\min}, \cleanDegree{\min}\}\geq c_1 \exp(-8 R^2/ \bandwidth{}^2)\numpoints{}$.
\end{proof}

\begin{proof}[Proof of Theorem~\ref{th:concentration-of-k-sub-Gaussian}]
As in the proof of Theorem~\ref{th:concentration-of-a-sub-Gaussian}, assume that $\numpoints{}$ is sufficiently large. Then,

$$\|\nAdjacency{}{} - \nCleanAdjacency{}{}\| = \|\degreeMat{}^{-1}\adjacency{}{}\degreeMat{}^{-1} - \cleanDegreeMat{}^{-1}\cleanAdjacency{}{}\cleanDegreeMat{}^{-1}\| \leq \|\degreeMat{}^{-1}(\adjacency{}{} - \cleanAdjacency{}{})\degreeMat{}^{-1}\| +  \| \degreeMat{}^{-1}\cleanAdjacency{}{}\degreeMat{}^{-1} -\cleanDegreeMat{}^{-1}\cleanAdjacency{}{}\cleanDegreeMat{}^{-1}\|.$$
The first term can be bounded easily using Theorem~\ref{th:concentration-of-a-sub-Gaussian} for both $\|\cdot\|_{F}$ and $\|\cdot\|_{\infty}$. Because $\degreeMat{}^{-1}$ is a diagonal matrix, left- and right-multiplication by $\degreeMat{}^{-1}$ scales the norms of any matrix by at most the square of its maximum diagonal entry, $1/\degree{\min}^2$. Thus, we have

$$\|\degreeMat{}^{-1}(\adjacency{}{} - \cleanAdjacency{}{})\degreeMat{}^{-1}\| \leq  \frac{\|\adjacency{}{} - \cleanAdjacency{}{}\|}{\degree{\min}^2}\leq \frac{\constantB{} \numpoints{}^{1/2}}{\degree{\min}^2}$$
with probability at least $1 - \numpoints{}^{-\boundParam{}+2}$ and where $\constantB{} \equiv \constantB{}(\boundParam{}, \constantA{})>0$.

For the second half, let $\aVector{} = \adjacency{}{} \mathbf{1}_\numpoints{}$, $\aCleanVector{} = \cleanAdjacency{}{}\mathbf{1}_\numpoints{}$ and $\degree{0} = \min\left\{  \degree{\min}, \cleanDegree{\min} \right\}$. Then, in the case of the Frobenius norm,
\begin{align*}
    &\| \degreeMat{}^{-1}\cleanAdjacency{}{}\degreeMat{}^{-1}  -\cleanDegreeMat{}^{-1}\cleanAdjacency{}{}\cleanDegreeMat{}^{-1}\|_F = \sqrt{\sum_{i, j=1}^\numpoints{} |\GaussianKernel{\bandwidth{}}(\cleandatapoint{i}-\cleandatapoint{j})|^2 \left(\frac{1}{\degree{i} \degree{j}} - \frac{1}{\cleanDegree{i} \cleanDegree{j}} \right)^2  } \leq  \sqrt{\sum_{i, j=1}^\numpoints{} \left(\frac{\cleanDegree{i} \cleanDegree{j} - \degree{i} \degree{j}}{\degree{i} \degree{j}\cleanDegree{i}\cleanDegree{j}} \right)^2  } \\
    &\qquad \leq \frac{1}{\degree{0}^4} \sqrt{\sum_{i, j=1}^\numpoints{} \left( {\cleanDegree{i} \cleanDegree{j}} - {\degree{i} \degree{j}} \right)^2  } \leq \frac{1}{\degree{0}^4} \|\aVector{}\aVector{}^T - \aCleanVector{}\aCleanVector{}^T\|_F \leq \frac{1}{\degree{0}^4} \left(\|\aVector{}(\aVector{} - \aCleanVector{})^T \|_F + \|\aCleanVector{}(\aVector{}-\aCleanVector{})^T\|_F\right)\\
    &\qquad \leq  \frac{1}{\degree{0}^4}  \left(\|\aVector{}\|_2\|\aVector{} - \aCleanVector{}\|_2 + \|\aCleanVector{}\|_2\|\aVector{}-\aCleanVector{}\|_2\right)  \leq  \frac{1}{\degree{0}^4}  \left(\|\adjacency{}{}\|_2 + \|\cleanAdjacency{}{}\|_2\right)\|\adjacency{}{} - \cleanAdjacency{}{}\|_2\|\mathbf{1}_\numpoints{}\|_2^2.
\end{align*}
Applying Theorem~\ref{th:concentration-of-a-sub-Gaussian} once again and using $\|\cleanAdjacency{}{}\|_2 \leq \|\cleanAdjacency{}{}\|_F = \sqrt{\sum_{i, j}\cleanAdjacency{i}{j}^2}\leq \numpoints{} \max_{i, j}\cleanAdjacency{i}{j} \leq \numpoints{}$, $\|\adjacency{}{}\|_2 \leq \numpoints{}$ and $\|\adjacency{}{} - \cleanAdjacency{}{}\|_2 \leq \|\adjacency{}{} - \cleanAdjacency{}{}\|_F$,  we have

$$\| \degreeMat{}^{-1}\cleanAdjacency{}{}\degreeMat{}^{-1}  -\cleanDegreeMat{}^{-1}\cleanAdjacency{}{}\cleanDegreeMat{}^{-1}\|_F \leq \frac{\constantB{}\numpoints{}^{5/2}}{ \degree{0}^4}$$
with probability at least $1 - \numpoints{}^{-\boundParam{}+2}$ and where $\constantB{} = \constantB{}(\boundParam{}, \constantA{})>0$. 

Similarly, in the case of infinity norm, using $\degree{i} \leq \numpoints{}$, $\cleanDegree{i} \leq \numpoints{}$ and $|\degree{i}-\cleanDegree{i}| \leq \|\adjacency{}{} - \cleanAdjacency{}{}\|_\infty$, 
\begin{align*}
    &\| \degreeMat{}^{-1}\cleanAdjacency{}{}\degreeMat{}^{-1}  -\cleanDegreeMat{}^{-1}\cleanAdjacency{}{}\cleanDegreeMat{}^{-1}\|_\infty = \max_{i=1}^{\numpoints{}} \sum_{j=1}^\numpoints{} |\GaussianKernel{\bandwidth{}}(\cleandatapoint{i}-\cleandatapoint{j})| \left|\frac{1}{\degree{i} \degree{j}} - \frac{1}{\cleanDegree{i} \cleanDegree{j}} \right| \leq \frac{1}{\degree{0}^4}\sum_{j=1}^{\numpoints{}}|\cleanDegree{i}\cleanDegree{j} - \degree{i}\degree{j}|\\
    &\quad  \leq \frac{1}{\degree{0}^4}\|\cleanDegree{i}\aCleanVector{} - \degree{i}\aVector{}\|_1 \leq \frac{\sqrt{\numpoints{}}}{\degree{0}^4}\|\cleanDegree{i}\aCleanVector{} - \degree{i}\aVector{}\|_2 \leq \frac{\sqrt{\numpoints{}}}{\degree{0}^4}(\degree{i}\|\aCleanVector{} - \aVector{}\|_2 + |\degree{i}-\cleanDegree{i}|\|\aCleanVector{}\|_2 )\\
    &\quad \leq \frac{\numpoints{}^{2}}{\degree{0}^4} (\|\adjacency{}{}-\cleanAdjacency{}{}\|_2 + |\degree{i}-\cleanDegree{i}| ) \leq \frac{\numpoints{}^{2}}{\degree{0}^4} (\|\adjacency{}{}-\cleanAdjacency{}{}\|_2 + \|\adjacency{}{}-\cleanAdjacency{}{}\|_\infty ) \leq \frac{\constantB{}\numpoints{}^{5/2}}{\degree{0}^4}
\end{align*}
with probability at least $1 - \numpoints{}^{-\boundParam{}+2}$ and where $\constantB{} = \constantB{}(\boundParam{}, \constantA{})>0$.

Combining the two estimates, we then have (for both Frobenius and infinity norms)

$$\|\nAdjacency{}{} - \nCleanAdjacency{}{}\| \leq  \frac{\constantB{} \numpoints{}^{1/2}}{\degree{\min}^2} + \frac{\constantB{} \numpoints{}^{5/2}}{ \degree{0}^4} \leq  \frac{\constantB{} \cleanDegree{\min}^2 \numpoints{}^{1/2}}{\cleanDegree{\min}^2d_{\min}^2} +  \frac{\constantB{} \numpoints{}^{5/2}}{ \degree{0}^4} \leq  \frac{\constantB{} \numpoints{}^{5/2}}{ \degree{0}^4}$$
with probability at least $1 - \numpoints{}^{-\boundParam{}+2}$. Finally, using Lemma~\ref{lem:degree-bounded-from-zero}, it follows that

$$\|\nAdjacency{}{} - \nCleanAdjacency{}{}\| \leq \frac{\constantB{} \numpoints{}^{5/2}}{\degree{0}^4} \leq \constantD{} \exp\left(\frac{32\radius{}^2}{\bandwidth{}^2}\right)\numpoints{}^{-3/2}$$
with probability at least $1 - 2\numpoints{}^{-\boundParam{}+2}$ and where $\constantD{} \equiv \constantD{}(\ambientdim{}, \boundParam{}, \constantA{})>0$.
\end{proof}

\begin{proof}[Proof of Theorem~\ref{th:final-concentration-of-l-sub-Gaussian}]
Assume, as in the proof of Theorem~\ref{th:concentration-of-k-sub-Gaussian}, that $\numpoints{}$ is sufficiently large. Begin by writing

$$\|\RWGraphLaplacian{} - \cleanRWGraphLaplacian{}\|_F = \|\nDegreeMat{}^{-1}\nAdjacency{}{} - \nCleanDegreeMat{}^{-1}\nCleanAdjacency{}{}\|_F \leq \|\nDegreeMat{}^{-1}(\nAdjacency{}{} - \nCleanAdjacency{}{})\|_F +  \| \nDegreeMat{}^{-1}\nCleanAdjacency{}{} -\nCleanDegreeMat{}^{-1}\nCleanAdjacency{}{}\|_F.$$
The first term can be bounded easily using Theorem~\ref{th:concentration-of-k-sub-Gaussian}, namely,

$$\|\nDegreeMat{}^{-1}(\nAdjacency{}{} - \nCleanAdjacency{}{})\|_F \leq \|\nDegreeMat{}^{-1}\|_2\|\nAdjacency{}{} - \nCleanAdjacency{}{}\|_F \leq  \frac{\|\nAdjacency{}{} - \nCleanAdjacency{}{}\|_F}{\nDegree{\min}}\leq \frac{\constantD{}\exp\left(\frac{32\radius{}^2}{\bandwidth{}^2}\right) \numpoints{}^{-3/2}}{\nDegree{\min}}$$
with probability at least $1 - 2\numpoints{}^{-\boundParam{}+2}$ and where $\constantD{} \equiv \constantD{}(\ambientdim{}, \boundParam{}, \constantA{})>0$. Note that

$$\nDegree{i} = \sum_{j=1}^{\numpoints{}} \nAdjacency{i}{j} = \frac{1}{\degree{i}}\sum_{j=1}^{\numpoints{}}\frac{\adjacency{i}{j}}{\degree{j}} \geq \frac{1}{\degree{\max}} \frac{1}{\degree{i}}\sum_{j=1}^{\numpoints{}}\adjacency{i}{j} = \frac{1}{\degree{\max}} \geq \frac{1}{\numpoints{}} \implies \frac{1}{\nDegree{\min}} \leq \numpoints{}.$$
Therefore, $\|\nDegreeMat{}^{-1}(\nAdjacency{}{} - \nCleanAdjacency{}{})\|_F \leq  \constantD{} \exp\left(\frac{32\radius{}^2}{\bandwidth{}^2}\right)\numpoints{}^{-1/2}$ with probability at least $1 - 2\numpoints{}^{-\boundParam{}+2}$.

For the second half, let $\aVector{} = \nAdjacency{}{} \mathbf{1}_\numpoints{}$, $\aCleanVector{} = \nCleanAdjacency{}{}\mathbf{1}_\numpoints{}$. Since $\cleanAdjacency{i}{j}\leq 1$, Lemma~\ref{lem:degree-bounded-from-zero} gives $\nCleanAdjacency{i}{j}\leq \cleanDegree{\min}^{-2}\leq \constantC{}^{-2}\exp\left(\frac{16\radius{}^2}{\bandwidth{}^2}\right)\numpoints{}^{-2}$, and, as above, $\nCleanDegree{i}\geq 1/\numpoints{}$. Hence $(\nCleanDegreeMat{}^{-1}\nCleanAdjacency{}{})_{ij}\leq \constantC{}^{-2}\exp\left(\frac{16\radius{}^2}{\bandwidth{}^2}\right)\numpoints{}^{-1}$ and $\|\nCleanDegreeMat{}^{-1}\nCleanAdjacency{}{}\|_F\leq \constantC{}^{-2}\exp\left(\frac{16\radius{}^2}{\bandwidth{}^2}\right)$. Using $\|AB\|_F\leq\|A\|_2\|B\|_F$,
\begin{align*}
    \| \nDegreeMat{}^{-1}\nCleanAdjacency{}{}  -\nCleanDegreeMat{}^{-1}\nCleanAdjacency{}{}\|_F &\leq \|\nDegreeMat{}^{-1}\nCleanDegreeMat{}-\identity{\numpoints{}}\|_2 \|\nCleanDegreeMat{}^{-1}\nCleanAdjacency{}{}\|_F \leq \frac{1}{\nDegree{\min}}\|\nCleanDegreeMat{}-\nDegreeMat{}\|_2\|\nCleanDegreeMat{}^{-1}\nCleanAdjacency{}{}\|_F\\
    &\leq \frac{1}{\nDegree{\min}}\|\aVector{}-\aCleanVector{}\|_\infty\|\nCleanDegreeMat{}^{-1}\nCleanAdjacency{}{}\|_F \leq \frac{1}{\nDegree{\min}}\|\nAdjacency{}{}-\nCleanAdjacency{}{}\|_\infty\|\nCleanDegreeMat{}^{-1}\nCleanAdjacency{}{}\|_F\\
    &\leq \constantC{}^{-2}\constantD{} \exp\left(\frac{48\radius{}^2}{\bandwidth{}^2}\right)\numpoints{}^{-1/2}.
\end{align*}
Combining the two halves gives the claim with constant $(1+\constantC{}^{-2})\constantD{}$.
\end{proof}

\begin{proof}[Proof of Corollary~\ref{cor:eigenvector-stability}]
Since $\nAdjacency{}{}$ and $\nCleanAdjacency{}{}$ are symmetric and nonnegative, $\symGraphLaplacian{}$ and $\cleanSymGraphLaplacian{}$ are symmetric. A direct calculation shows
\begin{align*}
    \RWGraphLaplacian{} &= \identity{\numpoints{}} - \nDegreeMat{}^{-1}\nAdjacency{}{}
    = \nDegreeMat{}^{-1/2}\bigl(\identity{\numpoints{}} - \nDegreeMat{}^{-1/2}\nAdjacency{}{}\nDegreeMat{}^{-1/2}\bigr)\nDegreeMat{}^{1/2} 
    = \nDegreeMat{}^{-1/2}\,\symGraphLaplacian{}\,\nDegreeMat{}^{1/2},\\
    \cleanRWGraphLaplacian{} &= \nCleanDegreeMat{}^{-1/2}\,\cleanSymGraphLaplacian{}\,\nCleanDegreeMat{}^{1/2}.
\end{align*}
Hence $\RWGraphLaplacian{}$ and $\symGraphLaplacian{}$ are similar (and likewise for $\cleanRWGraphLaplacian{}$ and $\cleanSymGraphLaplacian{}$), so they have identical eigenvalues. Moreover, if $\symGraphLaplacian{}\aVector_i = \eigval{i}\aVector_i$ then $\nDegreeMat{}^{-1/2}\aVector_i$ is an eigenvector of $\RWGraphLaplacian{}$ with eigenvalue $\eigval{i}$, and similarly for the clean pair.

On the high-probability event from Theorem~\ref{th:concentration-of-k-sub-Gaussian} and Lemma~\ref{lem:degree-bounded-from-zero}, the degrees satisfy
\begin{align*}
    \degree{\min},\cleanDegree{\min} \geq \constantC{}\exp\left(-\frac{8\radius{}^2}{\bandwidth{}^2}\right)\numpoints{}.
\end{align*}
Since $0 < \adjacency{i}{j} \leq 1$, it also holds that $\degree{i}\leq \numpoints{}$ and $\cleanDegree{i}\leq \numpoints{}$. Therefore, for all $i,j$,
\begin{align*}
    0 < \nAdjacency{i}{j} = \frac{\adjacency{i}{j}}{\degree{i}\degree{j}} \leq \degree{\min}^{-2},\qquad
    0 < \nCleanAdjacency{i}{j} = \frac{\cleanAdjacency{i}{j}}{\cleanDegree{i}\cleanDegree{j}} \leq \cleanDegree{\min}^{-2}.
\end{align*}
Consequently,
\begin{align*}
    \|\nAdjacency{}{}\|_2 \leq \|\nAdjacency{}{}\|_F \leq \numpoints{}\degree{\min}^{-2} \leq C\exp\left(\frac{16\radius{}^2}{\bandwidth{}^2}\right)\numpoints{}^{-1}
\end{align*}
for a constant $C>0$, and the same bound holds for $\|\nCleanAdjacency{}{}\|_2$.
In particular,
\begin{align*}
    \nDegree{i} = \sum_{j=1}^{\numpoints{}} \nAdjacency{i}{j} \leq \numpoints{}\degree{\min}^{-2}
    \leq C\exp\left(\frac{16\radius{}^2}{\bandwidth{}^2}\right)\numpoints{}^{-1}
\end{align*}
and the analogous bound holds for $\nCleanDegree{i}$.

Theorem~\ref{th:concentration-of-k-sub-Gaussian} gives
\begin{align*}
    \|\nAdjacency{}{} - \nCleanAdjacency{}{}\|_2 \leq \|\nAdjacency{}{} - \nCleanAdjacency{}{}\|_F
    \leq \constantD{} \exp\left(\frac{32\radius{}^2}{\bandwidth{}^2}\right)\numpoints{}^{-3/2},
\end{align*}
and the same theorem yields the row-sum bound
\begin{align*}
    \|\nAdjacency{}{} - \nCleanAdjacency{}{}\|_\infty \leq \constantD{} \exp\left(\frac{32\radius{}^2}{\bandwidth{}^2}\right)\numpoints{}^{-3/2},
\end{align*}
where $\|M\|_\infty = \max_i \sum_j |M_{ij}|$. Therefore,
\begin{align*}
    \max_i |\nDegree{i} - \nCleanDegree{i}| \leq \|\nAdjacency{}{} - \nCleanAdjacency{}{}\|_\infty
    \leq \constantD{} \exp\left(\frac{32\radius{}^2}{\bandwidth{}^2}\right)\numpoints{}^{-3/2}.
\end{align*}
Finally, since $0<\adjacency{i}{j}\leq 1$ implies $\degree{j}\leq \numpoints{}$ for all $j$,
\begin{align*}
    \nDegree{i} = \frac{1}{\degree{i}}\sum_{j=1}^{\numpoints{}}\frac{\adjacency{i}{j}}{\degree{j}}
    \geq \frac{1}{\degree{i}}\sum_{j=1}^{\numpoints{}}\frac{\adjacency{i}{j}}{\numpoints{}}
    = \frac{1}{\numpoints{}}.
\end{align*}
Hence
\begin{align*}
    \|\nDegreeMat{}^{-1/2}\|_2 \leq \numpoints{}^{1/2},\qquad
    \|\nCleanDegreeMat{}^{-1/2}\|_2 \leq \numpoints{}^{1/2}.
\end{align*}
Using the mean value theorem for $x\mapsto x^{-1/2}$ and the diagonal structure of $\nDegreeMat{}$, for each $i$ there exists $\zeta_i$ between $\nDegree{i}$ and $\nCleanDegree{i}$ such that
\begin{align*}
    \bigl|\nDegree{i}^{-1/2} - \nCleanDegree{i}^{-1/2}\bigr|
    = \frac{1}{2}\zeta_i^{-3/2}\,|\nDegree{i}-\nCleanDegree{i}|.
\end{align*}
Since $\nDegree{i},\nCleanDegree{i}\geq \numpoints{}^{-1}$, we have $\zeta_i^{-3/2}\leq \numpoints{}^{3/2}$ and therefore
\begin{align*}
    \bigl|\nDegree{i}^{-1/2} - \nCleanDegree{i}^{-1/2}\bigr|
    \leq \frac{1}{2}\numpoints{}^{3/2}|\nDegree{i}-\nCleanDegree{i}|.
\end{align*}
Taking the maximum over $i$ (which equals the operator norm for diagonal matrices) yields
\begin{align*}
    \|\nDegreeMat{}^{-1/2} - \nCleanDegreeMat{}^{-1/2}\|_2
    \leq \frac{1}{2}\numpoints{}^{3/2}\max_i |\nDegree{i} - \nCleanDegree{i}|
    \leq C\exp\left(\frac{32\radius{}^2}{\bandwidth{}^2}\right).
\end{align*}

By the definition of the symmetric normalized Laplacian,
\begin{align*}
    \cleanSymGraphLaplacian{}-\symGraphLaplacian{}
    &= \left(\nDegreeMat{}^{-1/2} - \nCleanDegreeMat{}^{-1/2}\right)\nAdjacency{}{}\,\nDegreeMat{}^{-1/2}
    + \nCleanDegreeMat{}^{-1/2}\left(\nAdjacency{}{} - \nCleanAdjacency{}{}\right)\nDegreeMat{}^{-1/2}
    + \nCleanDegreeMat{}^{-1/2}\,\nCleanAdjacency{}{}\left(\nDegreeMat{}^{-1/2} - \nCleanDegreeMat{}^{-1/2}\right).
\end{align*}
Taking operator norms and using the bounds above gives
\begin{align*}
    \|\symGraphLaplacian{} - \cleanSymGraphLaplacian{}\|_2
    &\leq \|\nDegreeMat{}^{-1/2} - \nCleanDegreeMat{}^{-1/2}\|_2\,\|\nAdjacency{}{}\|_2\,\|\nDegreeMat{}^{-1/2}\|_2 \\
    &\phantom{\leq {}}+ \|\nCleanDegreeMat{}^{-1/2}\|_2\,\|\nAdjacency{}{}-\nCleanAdjacency{}{}\|_2\,\|\nDegreeMat{}^{-1/2}\|_2 \\
    &\phantom{\leq {}}+ \|\nCleanDegreeMat{}^{-1/2}\|_2\,\|\nCleanAdjacency{}{}\|_2\,\|\nDegreeMat{}^{-1/2} - \nCleanDegreeMat{}^{-1/2}\|_2\\
    &\leq C\exp\left(\frac{48\radius{}^2}{\bandwidth{}^2}\right)\numpoints{}^{-1/2}.
\end{align*}
Define the spectral gap
\begin{align*}
    g_i = \begin{cases}
    |\ceigval{2}-\ceigval{1}|, & i=1,\\
    \min\{|\ceigval{i}-\ceigval{i-1}|,\,|\ceigval{i+1}-\ceigval{i}|\}, & 2\leq i\leq \numeigforgrad{}-1.
    \end{cases}
\end{align*}
By assumption, $g_i = \omega(\numpoints{}^{-1/2})$. Davis--Kahan (e.g., \cite{yuUsefulVariantDavis2015}) applied to the symmetric matrices $\symGraphLaplacian{}$ and $\cleanSymGraphLaplacian{}$ yields a sign $\tau_i\in\{\pm1\}$ such that
\begin{align*}
    \|\aVector_i - \tau_i\aCleanVector_i\|_2
    \leq \frac{2^{3/2}\|\symGraphLaplacian{} - \cleanSymGraphLaplacian{}\|_2}{g_i}
    \leq C\exp\left(\frac{48\radius{}^2}{\bandwidth{}^2}\right)\frac{\numpoints{}^{-1/2}}{g_i}
    = \exp\left(\frac{48\radius{}^2}{\bandwidth{}^2}\right)o_n(1).
\end{align*}

By construction,
\begin{align*}
    \eigvec{i} = \frac{\nDegreeMat{}^{-1/2}\aVector_i}{\|\nDegreeMat{}^{-1/2}\aVector_i\|_2},\qquad
    \ceigvec{i} = \frac{\nCleanDegreeMat{}^{-1/2}\aCleanVector_i}{\|\nCleanDegreeMat{}^{-1/2}\aCleanVector_i\|_2}.
\end{align*}
Since $\nDegree{i}\in[\numpoints{}^{-1}, C\exp\left(\frac{16\radius{}^2}{\bandwidth{}^2}\right)\numpoints{}^{-1}]$, we have
\begin{align*}
    c\exp\left(-\frac{8\radius{}^2}{\bandwidth{}^2}\right)\numpoints{}^{1/2} \leq \|\nDegreeMat{}^{-1/2}\aVector_i\|_2,\,\|\nCleanDegreeMat{}^{-1/2}\aCleanVector_i\|_2 \leq C\numpoints{}^{1/2}
\end{align*}
for constants $0<c<C$. Then
\begin{align*}
    \|\eigvec{i} - \tau_i\ceigvec{i}\|_2
    &\leq \frac{\|\nDegreeMat{}^{-1/2}(\aVector_i - \tau_i\aCleanVector_i)\|_2}{\|\nDegreeMat{}^{-1/2}\aVector_i\|_2}
    + \frac{\|(\nDegreeMat{}^{-1/2}-\nCleanDegreeMat{}^{-1/2})\aCleanVector_i\|_2}{\|\nDegreeMat{}^{-1/2}\aVector_i\|_2}\\
    &\phantom{\leq {}}+ \left|\frac{1}{\|\nDegreeMat{}^{-1/2}\aVector_i\|_2} - \frac{1}{\|\nCleanDegreeMat{}^{-1/2}\aCleanVector_i\|_2}\right|\,\|\nCleanDegreeMat{}^{-1/2}\aCleanVector_i\|_2.
\end{align*}
The first term is bounded by $C\exp\left(\frac{8\radius{}^2}{\bandwidth{}^2}\right)\|\aVector_i - \tau_i\aCleanVector_i\|_2$. The second term satisfies
\begin{align*}
    \frac{\|(\nDegreeMat{}^{-1/2}-\nCleanDegreeMat{}^{-1/2})\aCleanVector_i\|_2}{\|\nDegreeMat{}^{-1/2}\aVector_i\|_2}
    \leq C\exp\left(\frac{40\radius{}^2}{\bandwidth{}^2}\right)\numpoints{}^{-1/2}.
\end{align*}
For the last term, note that
\begin{align*}
    \bigl|\|\nDegreeMat{}^{-1/2}\aVector_i\|_2 - \|\nCleanDegreeMat{}^{-1/2}\aCleanVector_i\|_2\bigr|
    &\leq \|(\nDegreeMat{}^{-1/2}-\nCleanDegreeMat{}^{-1/2})\aCleanVector_i\|_2 + \|\nDegreeMat{}^{-1/2}(\aVector_i - \tau_i\aCleanVector_i)\|_2\\
    &\leq C\exp\left(\frac{32\radius{}^2}{\bandwidth{}^2}\right) + C\numpoints{}^{1/2}\|\aVector_i - \tau_i\aCleanVector_i\|_2,
\end{align*}
so
\begin{align*}
    \left|\frac{1}{\|\nDegreeMat{}^{-1/2}\aVector_i\|_2} - \frac{1}{\|\nCleanDegreeMat{}^{-1/2}\aCleanVector_i\|_2}\right|\,\|\nCleanDegreeMat{}^{-1/2}\aCleanVector_i\|_2
    &\leq \frac{\bigl|\|\nDegreeMat{}^{-1/2}\aVector_i\|_2 - \|\nCleanDegreeMat{}^{-1/2}\aCleanVector_i\|_2\bigr|}{\|\nDegreeMat{}^{-1/2}\aVector_i\|_2\,\|\nCleanDegreeMat{}^{-1/2}\aCleanVector_i\|_2}\,\|\nCleanDegreeMat{}^{-1/2}\aCleanVector_i\|_2\\
    &\leq C\exp\left(\frac{40\radius{}^2}{\bandwidth{}^2}\right)\numpoints{}^{-1/2} + C\exp\left(\frac{8\radius{}^2}{\bandwidth{}^2}\right)\|\aVector_i - \tau_i\aCleanVector_i\|_2.
\end{align*}
Combining the three bounds yields
\begin{align*}
    \|\eigvec{i} - \tau_i\ceigvec{i}\|_2
    &\leq C\exp\left(\frac{8\radius{}^2}{\bandwidth{}^2}\right)\|\aVector_i - \tau_i\aCleanVector_i\|_2
    + C\exp\left(\frac{40\radius{}^2}{\bandwidth{}^2}\right)\numpoints{}^{-1/2}\\
    &\leq C\exp\left(\frac{56\radius{}^2}{\bandwidth{}^2}\right)\left(\frac{\numpoints{}^{-1/2}}{g_i} + \numpoints{}^{-1/2}\right)\\
    &= \exp\left(\frac{56\radius{}^2}{\bandwidth{}^2}\right)o_n(1).
\end{align*}
This proves the corollary on the same high-probability event, hence with probability at least $1-2\numpoints{}^{-\boundParam{}+2}$.
\end{proof}

\begin{proof}[Proof of~\Cref{prop:rmt_eig_grad_low_noise}]
For brevity, define $\widetilde{\xi}_j = \xi_{j}^{\circ 2}$ and note that $\|\widetilde{\xi}_j\|_2 = \|\xi_j\|_4^2 = \bigO(\sqrt{\knn{}}\knnradius{2})$.
Then, $\|\cceigvecat{i}{j}\|_2 \leq \|\ccdatapoint{j}\OBofTSat{j}\|_2\|\mathfrak{g}_{ij}\|_2 + \|\widetilde{\xi}_{j}\|_2$ which implies $\|\cceigvecat{i}{j}\|_2 = \bigO(\sqrt{\knn{}} \knnradius{})$.
Also, using the definition of $(\ccdatapoint{j})^{+}$,
\begin{align*}
    (\ccdatapoint{j})^{+}\cceigvecat{i}{j} &= (\ccdatapoint{j})^{+}\ccdatapoint{j}\OBofTSat{j}\mathfrak{g}_{ij} + (\ccdatapoint{j})^{+}\widetilde{\xi}_j\\
    &= \OBofTSat{j}\mathfrak{g}_{ij} - \overline{\eta}_j((\ccdatapoint{j})^{T}\ccdatapoint{j} + \overline{\eta}_{j} I_{\ambientdim{}})^{-1}\OBofTSat{j}\mathfrak{g}_{ij} + (\ccdatapoint{j})^{+}\widetilde{\xi}_j
\end{align*}
where, using $\overline{\eta}_j = \Theta(\knn{} \knnradius{3})$ and~\Cref{lem:tkv_cross_norm} (adapted for clean data i.e., with $\varianceProxy{} = 0$),
\begin{align*}
    \|\overline{\eta}_j((\ccdatapoint{j})^{T}\ccdatapoint{j} + \overline{\eta}_{j} I_{\ambientdim{}})^{-1}\OBofTSat{j}\|_2 &= \bigO\left(\frac{\knn{}\knnradius{3}}{\knn{}\knnradius{2} + \knn{}\knnradius{3}}\right) = \bigO(\knnradius{})\\
    \|(\ccdatapoint{j})^{+}\widetilde{\xi}_j\|_2 \leq \frac{\|\widetilde{\xi}_j\|_2}{2\sqrt{\overline{\eta}_j}} &= \frac{\|\xi_j\|_4^2}{2\sqrt{\overline{\eta}_j}} = \bigO\left(\frac{\sqrt{\knn{}}\knnradius{2}}{\sqrt{\knn{}}\knnradius{3/2}}\right) = \bigO(\knnradius{1/2}).
\end{align*}
Combining the above equations, we conclude that $\|(\ccdatapoint{j})^{+}\cceigvecat{i}{j}\|_2 = \bigO(1)$. Then,
\begin{align*}
    \|\OBofTSat{j}^T\tkvestgradeigvecat{i}{j} - \OBofTSat{j}^T\tkvestgradceigvecat{i}{j}\|_2 &\leq \|\OBofTSat{j}^T(\cdatapoint{j}^{+}(\ceigvecat{i}{j} - \cceigvecat{i}{j}))\|_2 + \|\OBofTSat{j}^T(\cdatapoint{j}^{+} - (\ccdatapoint{j})^{+})\cceigvecat{i}{j}\|_2 \\
    &\leq \|\OBofTSat{j}^T\cdatapoint{j}^{+}\|_2\|\ceigvecat{i}{j} - \cceigvecat{i}{j}\|_2 + \|\OBofTSat{j}^T(\cdatapoint{j}^{+} - (\ccdatapoint{j})^{+})\cceigvecat{i}{j}\|_2.
\end{align*}
Since $\cdatapoint{j} = \ccdatapoint{j} + \cnoiseRV{j}$,
\begin{align*}
    &\cdatapoint{j}^{+} - (\ccdatapoint{j})^{+} = (\cdatapoint{j}^T\cdatapoint{j} + \eta_j I_{\ambientdim{}})^{-1}\cdatapoint{j}^T - (\ccdatapoint{j}^T\ccdatapoint{j} + \overline{\eta}_{j}I_{\ambientdim{}})^{-1}\ccdatapoint{j}^T\\
    &= (\cdatapoint{j}^T\cdatapoint{j} + \eta_j I_{\ambientdim{}})^{-1}\cnoiseRV{j}^T + ((\cdatapoint{j}^T\cdatapoint{j} + \eta_j I_{\ambientdim{}})^{-1} - (\ccdatapoint{j}^T\ccdatapoint{j} + \overline{\eta}_{j}I_{\ambientdim{}})^{-1})\ccdatapoint{j}^T\\
    &= (\cdatapoint{j}^T\cdatapoint{j} + \eta_j I_{\ambientdim{}})^{-1}\cnoiseRV{j}^T - (\cdatapoint{j}^T\cdatapoint{j} + \eta_j I_{\ambientdim{}})^{-1}(\cdatapoint{j}^T\cdatapoint{j} + \eta_j I_{\ambientdim{}} - \ccdatapoint{j}^T\ccdatapoint{j} - \overline{\eta}_{j}I_{\ambientdim{}})(\ccdatapoint{j})^{+}\\
    &= (\cdatapoint{j}^T\cdatapoint{j} + \eta_j I_{\ambientdim{}})^{-1}\cnoiseRV{j}^T - (\cdatapoint{j}^T\cdatapoint{j} + \eta_j I_{\ambientdim{}})^{-1}(\ccdatapoint{j}^T\cnoiseRV{j} + \cnoiseRV{j}^T\ccdatapoint{j} + \cnoiseRV{j}^T\cnoiseRV{j} + (\eta_j-\overline{\eta}_{j})I_{\ambientdim{}})(\ccdatapoint{j})^{+}.
\end{align*}
Analyzing each term separately using~\Cref{lem:tkv_cross_norm},
\begin{align*}
    \|\OBofTSat{j}^T(\cdatapoint{j}^T\cdatapoint{j} + \eta_j I_{\ambientdim{}})^{-1}\cnoiseRV{j}^T\cceigvecat{i}{j}\|_2 &\leq \|\OBofTSat{j}^T(\cdatapoint{j}^T\cdatapoint{j} + \eta_j I_{\ambientdim{}})^{-1}\cnoiseRV{j}^T\|_2\|\cceigvecat{i}{j}\|_2\\
    &=\bigO\left(\left(\frac{\|\OBofTSat{j}^T\cnoiseRV{j}^T\|_2}{\knn{}\knnradius{2}} + \frac{\|(\OBofNSat{j})^T\cnoiseRV{j}^T\|_2\knnradius{3/2+\gamma}}{\knn{}\knnradius{4}}\right)\|\cceigvecat{i}{j}\|_2\right)\\
    &=\bigO\left(\left(\frac{\sqrt{\knn{}}\varianceProxy{}_{\numpoints{}}\knnradius{}}{\knn{}\knnradius{2}} + \frac{(\sqrt{\knn{}}\varianceProxy{}_{\numpoints{}})\knnradius{3/2+\gamma}}{\knn{}\knnradius{4}}\right)\sqrt{\knn{}}\knnradius{}\right)\\
    &=\bigO\left(\varianceProxy{}_{\numpoints{}} + \varianceProxy{}_{\numpoints{}}\frac{\knnradius{3/2+\gamma}}{\knnradius{3}}\right) = \bigO(\knnradius{2\gamma}),
\end{align*}
\begin{align*}
    &\|\OBofTSat{j}^T(\cdatapoint{j}^T\cdatapoint{j} + \eta_j I_{\ambientdim{}})^{-1}\ccdatapoint{j}^T\cnoiseRV{j}(\ccdatapoint{j})^{+}\cceigvecat{i}{j}\|_2\\
    &\hspace{0.5cm}\leq \|\OBofTSat{j}^T(\cdatapoint{j}^T\cdatapoint{j} + \eta_j I_{\ambientdim{}})^{-1}\ccdatapoint{j}^T\|_2\|\cnoiseRV{j}\|_2\|(\ccdatapoint{j})^{+}\cceigvecat{i}{j}\|_2\\
    &\hspace{0.5cm}=\bigO\left(\left(\frac{\|\OBofTSat{j}^T\ccdatapoint{j}^T\|_2}{\knn{}\knnradius{2}} + \frac{\|(\OBofNSat{j})^T\ccdatapoint{j}^T\|_2\knnradius{3/2+\gamma}}{\knn{}\knnradius{4}}\right)\|\cnoiseRV{j}\|_2\|(\ccdatapoint{j})^{+}\cceigvecat{i}{j}\|_2\right)\\
    &\hspace{0.5cm}=\bigO\left(\left(\frac{\sqrt{\knn{}}\knnradius{}}{\knn{}\knnradius{2}} + \frac{(\sqrt{\knn{}}\knnradius{2})\knnradius{3/2+\gamma}}{\knn{}\knnradius{4}}\right)\sqrt{\knn{}}\varianceProxy{}_{\numpoints{}}\right) =\bigO\left(\frac{\varianceProxy{}_{\numpoints{}}}{\knnradius{}} + \knnradius{}\varianceProxy{}_{\numpoints{}}\frac{\knnradius{3/2+\gamma}}{\knnradius{3}}\right) = \bigO(\knnradius{1/2 + \gamma}),
\end{align*}
\begin{align*}
    &\|\OBofTSat{j}^T(\cdatapoint{j}^T\cdatapoint{j} + \eta_j I_{\ambientdim{}})^{-1}\cnoiseRV{j}^T\ccdatapoint{j}(\ccdatapoint{j})^{+}\cceigvecat{i}{j}\|_2\\
    &\hspace{0.5cm}\leq \|\OBofTSat{j}^T(\cdatapoint{j}^T\cdatapoint{j} + \eta_j I_{\ambientdim{}})^{-1}\cnoiseRV{j}^T\|_2\|\ccdatapoint{j}\|_2\|(\ccdatapoint{j})^{+}\cceigvecat{i}{j}\|_2\\
    &\hspace{0.5cm}=\bigO\left(\left(\frac{\|\OBofTSat{j}^T\cnoiseRV{j}^T\|_2}{\knn{}\knnradius{2}} + \frac{\|(\OBofNSat{j})^T\cnoiseRV{j}^T\|_2\knnradius{3/2+\gamma}}{\knn{}\knnradius{4}}\right)\|\ccdatapoint{j}\|_2\|(\ccdatapoint{j})^{+}\cceigvecat{i}{j}\|_2\right)\\
    &\hspace{0.5cm}=\bigO\left(\left(\frac{\sqrt{\knn{}}\varianceProxy{}_{\numpoints{}}\knnradius{}}{\knn{}\knnradius{2}} + \frac{(\sqrt{\knn{}}\varianceProxy{}_{\numpoints{}})\knnradius{3/2+\gamma}}{\knn{}\knnradius{4}}\right)\sqrt{\knn{}}\knnradius{}\right) =\bigO\left(\varianceProxy{}_{\numpoints{}} + \varianceProxy{}_{\numpoints{}}\frac{\knnradius{3/2+\gamma}}{\knnradius{3}}\right) = \bigO(\knnradius{2\gamma}),
\end{align*}
\begin{align*}
    \|\OBofTSat{j}^T(\cdatapoint{j}^T\cdatapoint{j} + \eta_j I_{\ambientdim{}})^{-1}\cnoiseRV{j}^T\cnoiseRV{j}(\ccdatapoint{j})^{+}\cceigvecat{i}{j}\|_2 = \bigO\left(\frac{\varianceProxy{}_{\numpoints{}}^2}{\knnradius{}}\left(1 + \frac{\knnradius{3/2+\gamma}}{\knnradius{3}}\right)\right)= \bigO(\knnradius{1/2+3\gamma}),
\end{align*}
and, using $A = I$ in~\Cref{lem:tkv_cross_norm} with $\eta_{j} = \sum_{s=1}^{\knn{}} \left\|\datapoint{j_s}-\datapoint{j}\right\|_2^{3}, \overline{\eta}_{j}  = \sum_{s=1}^{\knn{}} \left\|\cleandatapoint{j_s}-\cleandatapoint{j}\right\|_2^{3} \implies \eta_j - \overline{\eta}_{j} = \bigO(\knn{}\knnradius{2}\varianceProxy{}_{\numpoints{}})$,
\begin{align*}
    \OBofTSat{j}^T(\cdatapoint{j}^T\cdatapoint{j} + \eta_j I_{\ambientdim{}})^{-1}(\eta_j - \overline{\eta}_{j})(\ccdatapoint{j})^{+}\cceigvecat{i}{j} = \bigO(\varianceProxy{}_{\numpoints{}}\knnradius{\gamma-1/2}) = \bigO(\knnradius{1+2\gamma})
\end{align*}

Similarly, using~\Cref{lem:tkv_cross_norm},
\begin{align*}
    &\|\OBofTSat{j}^T\cdatapoint{j}^{+}\|_2 \|\ceigvecat{i}{j} - \cceigvecat{i}{j}\|_2\\
    &\hspace{0.5cm}=\|\OBofTSat{j}^T(\cdatapoint{j}^T\cdatapoint{j} + \eta_j I_{\ambientdim{}})^{-1}\cdatapoint{j}^T\|_2 \|\ceigvecat{i}{j} - \cceigvecat{i}{j}\|_2\\
    &\hspace{0.5cm}=\bigO\left(\left(\frac{\|\OBofTSat{j}^T\cdatapoint{j}^T\|_2}{\knn{}\knnradius{2}} + \frac{\|(\OBofNSat{j})^T\cdatapoint{j}^T\|_2\knnradius{3/2+\gamma}}{\knn{}\knnradius{4}}\right) \|\ceigvecat{i}{j} - \cceigvecat{i}{j}\|_2\right)\\
    &\hspace{0.5cm}=\bigO\left(\left(\frac{\sqrt{\knn{}}(\knnradius{} + \varianceProxy{}_{\numpoints{}}\knnradius{})}{\knn{}\knnradius{2}} + \frac{(\sqrt{\knn{}}(\knnradius{2} + \varianceProxy{}_{\numpoints{}}))\knnradius{3/2+\gamma}}{\knn{}\knnradius{4}}\right) \|\ceigvecat{i}{j} - \cceigvecat{i}{j}\|_2\right)\\
    &\hspace{0.5cm}=\bigO\left(\left(\frac{1}{\sqrt{\knn{}}\knnradius{}} + \frac{\knnradius{3+2\gamma}}{\sqrt{\knn{}}\knnradius{4}}\right) \|\ceigvecat{i}{j} - \cceigvecat{i}{j}\|_2\right)\\
    &\hspace{0.5cm}=\bigO\left(\left(\frac{1}{\sqrt{\knn{}}\knnradius{}} + \frac{\knnradius{2\gamma}}{\sqrt{\knn{}}\knnradius{}}\right) \|\ceigvecat{i}{j} - \cceigvecat{i}{j}\|_2\right)\\
    &\hspace{0.5cm}=\bigO\left(\left(\frac{1}{\sqrt{\knn{}}\knnradius{}}\right) \|\ceigvecat{i}{j} - \cceigvecat{i}{j}\|_2\right) = \bigO\left(\frac{\varianceProxy{}_{\numpoints{}}}{\knnradius{}}\right) = \bigO(\knnradius{1/2+\gamma}).
\end{align*}

Combining the inequalities, the result follows.
\end{proof}

\section*{Acknowledgments}
DK and SJR were partially supported by the Halicio\u{g}lu Data Science Institute PhD Fellowship. DK and GM acknowledge support from NSF EFRI 2223822. AC and GM acknowledge support from NSF CCF 2403452.

\bibliographystyle{plain} 
\bibliography{common/main}

\end{document}